\documentclass{article}

\PassOptionsToPackage{numbers, compress}{natbib}

\usepackage[main, final]{neurips_2026}

\usepackage[utf8]{inputenc} \usepackage[T1]{fontenc}    \usepackage{hyperref}       \usepackage{url}            \usepackage{booktabs}       \usepackage{amsfonts}       \usepackage{nicefrac}       \usepackage{microtype}      \usepackage{xcolor}

\usepackage{multirow}
\usepackage{hwemoji}
\usepackage{subcaption}
\usepackage[utf8]{inputenc}
\usepackage{amsmath}
\usepackage[capitalise,noabbrev]{cleveref}

\usepackage{soul}
\usepackage{tcolorbox}
\usepackage[dvipsnames,table]{xcolor}
\newcommand{\highlight}[2]{\setlength{\fboxsep}{1pt}\colorbox{SkyBlue!#1}{\strut{#2}}}

\ExplSyntaxOn
\NewDocumentCommand{\rdim}{m}
{
  \mathbb{R}^{\custom_rdim:n {#1}}
}
\seq_new:N \l_custom_rdim_seq
\cs_new:Npn \custom_rdim:n #1
{
  \seq_set_from_clist:Nn \l_custom_rdim_seq {#1}
  \seq_set_map:NNn \l_custom_rdim_seq \l_custom_rdim_seq { d\sb{\text{##1}} }
  \seq_use:Nn \l_custom_rdim_seq { \times }
}
\ExplSyntaxOff

\usepackage{amsmath,amsfonts,bm}

\def\eqref#1{equation~\ref{#1}}

\def\1{\bm{1}}

\def\va{{\bm{a}}}
\def\vb{{\bm{b}}}

\def\vk{{\bm{k}}}
\def\vl{{\bm{l}}}

\def\vq{{\bm{q}}}

\def\vs{{\bm{s}}}

\def\vv{{\bm{v}}}

\def\vx{{\bm{x}}}

\def\vz{{\bm{z}}}

\def\mA{{\bm{A}}}

\def\mS{{\bm{S}}}

\def\mW{{\bm{W}}}

\DeclareMathAlphabet{\mathsfit}{\encodingdefault}{\sfdefault}{m}{sl}
\SetMathAlphabet{\mathsfit}{bold}{\encodingdefault}{\sfdefault}{bx}{n}

\def\sR{{\mathbb{R}}}

\title{What Does a Sparse Autoencoder Feature Do? \\A Weight-Based Account}

\author{Yiting Liu \qquad Zhi-Hong Deng\thanks{Corresponding author.} \\
  State Key Laboratory of General Artificial Intelligence\\
  School of Intelligence Science and Technology\\
  Peking University\\
  \texttt{\{arkcia,zhdeng\}@pku.edu.cn} \\
}

\begin{document}

\maketitle

\begin{abstract}
  Sparse autoencoders (SAEs) have emerged as a powerful technique for decomposing language model representations into interpretable features.
  Current interpretation pipelines infer feature semantics from activation patterns, implicitly assuming every feature has a semantic explanation while overlooking the computational roles features inherit from the SAE training objective.
  We introduce a weight-based interpretation framework that requires no activation data and grounds claims causally through targeted feature-ablation interventions.
  Across Gemma-2 and Llama-3.1, three depth-dependent signatures show that SAE features inherit the base model's geometry:
  semantic features form a U-shape under tied embeddings but bifurcate when untied, attention participation peaks mid-layer, and the two roles couple oppositely for input- vs. output-oriented features.
  This weight-based view supplies the causal half missing from activation-based interpretability.
\end{abstract}

\section{Introduction}

A language model represents more features than it has directions, forcing it to encode features in superposition, which likely explains why many neurons are polysemantic and difficult to explain \citep{elhage2022ToyModelsSuperposition,bills2023language}.
Sparse autoencoders (SAEs) have emerged as a promising approach for decomposing such representations into more interpretable features \citep{huben2023SparseAutoencodersFind,bricken2023MonosemanticityDecomposingLanguage,gao2024ScalingEvaluatingSparse,templeton2024ScalingMonosemanticityExtracting}.
By training higher-dimensional reconstructions of model activations, SAEs aim to disentangle the superposed representations into sparse and semantically meaningful components.

\begin{figure}[t]
  \centering
  \begin{subfigure}{0.48\linewidth}
    \centering
    \resizebox{\linewidth}{!}{
      \input{figures/11433}
    }
    \caption{\small Activation-based explanation by GPT-4o-mini.}
    \label{fig:11433}
  \end{subfigure}
  \begin{subfigure}{0.48\linewidth}
    \centering
    \includegraphics[width=\linewidth,trim={.0cm .0cm .0cm .0cm}]{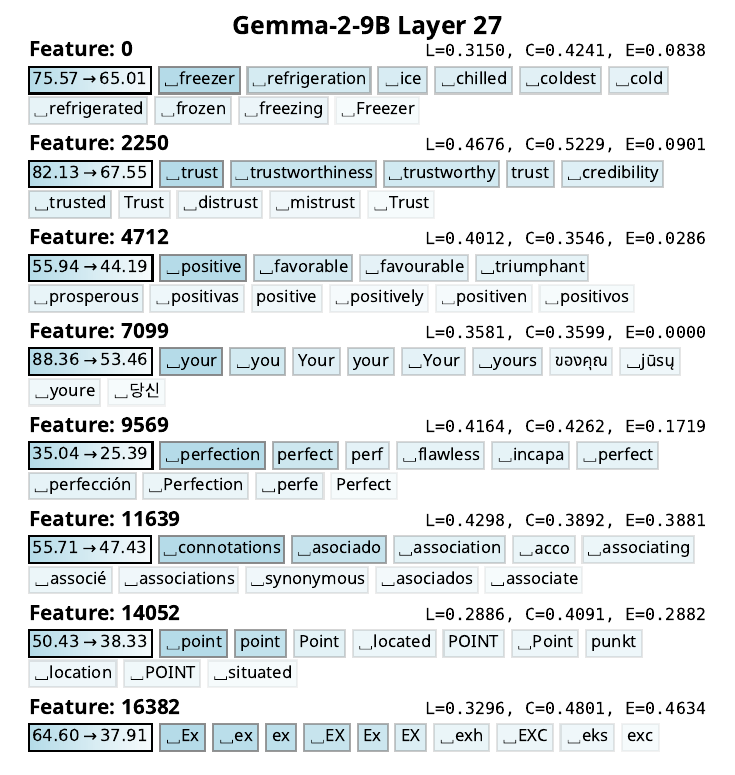}
    \caption{\small Sample of weight-flagged semantic features with their top-10 predicted tokens.}
    \label{fig:feature-token-sample-gemma9b-layer27}
  \end{subfigure}
  \caption{Two views of SAE features. (\subref{fig:11433}) An activation-based explanation read off the feature's top-activating contexts: tokens where the feature fires are highlighted and the highest activations are boxed, and GPT-4o-mini summarizes the co-occurring text. The summary does not specify what changes in the model's computation when the feature is suppressed. (\subref{fig:feature-token-sample-gemma9b-layer27}) A uniform sample of features that pass all three weight-based metrics in \Cref{sec:exp1} on Gemma-2-9B at layer 27, shown with their top-10 predicted tokens. ``L'' is Levenshtein similarity, ``C'' is cosine similarity, and ``E'' is top-100 entropy; boxed scores correspond to the range of displayed tokens.}
\end{figure}

Because SAE features emerge as unlabeled latent variables, their meanings must be inferred post-hoc.
The dominant paradigm asks ``\textit{when does a feature activate?}'' and infers meaning from the input contexts that drive the feature high; at scale this is automated by prompting an LLM with top-activating examples \citep{bricken2023MonosemanticityDecomposingLanguage,gao2024ScalingEvaluatingSparse,paulo2025AutomaticallyInterpretingMillions}.
The resulting descriptions are often broad summaries of co-occurring text and have been shown to be superficial and oversimplified \citep{ma2025RevisingFalsifyingSparse};
even when descriptively accurate, they cannot specify what changes in the model's computation when the feature is activated or suppressed.
This is a structural limitation: SAEs are trained to reconstruct activations, so features inherit the computational function of those activations, yet activation-based interpretation only reads back the contexts in which the feature appears.
Activation-based interpretation is, in this sense, correlational rather than mechanistic, leaving the computational roles of most features unexamined.

We complement this paradigm with the question that titles the paper: \textit{what does a feature do?}
We propose a weight-based methodology that analyzes features through their direct interactions with model components, requiring no datasets or external models, and uses targeted ablation to verify the structurally predicted effect.
Because we have full access to the SAE and the surrounding model, every feature can be projected onto the components they read from and write to, allowing causal claims to be grounded in counterfactual interventions on the feature itself rather than in correlations across activating prompts.

Our contributions are threefold.
\textbf{1)} We propose a weight-based SAE interpretation framework that requires neither LLM inference nor a corpus, and grounds claims through counterfactual interventions rather than activation correlations.
\textbf{2)} We investigate both the semantic and the previously overlooked computational roles of SAE features, challenging the implicit assumption underlying activation-based explanations that contexts suffice to characterize what a feature does.
\textbf{3)} We surface 3 depth-dependent signatures that reveal how SAEs inherit the base model's architectural geometry, with implications for cross-architecture comparison and depth-aggregated feature analysis.

The code to reproduce the experiments is available at \url{https://github.com/ArkciaTheDragon/what-sae-features-do}.

\section{Related Work}
As discussed, the predominant approach to SAE interpretability is activation-based: textual examples that activate a feature are presented to an LLM to generate semantic explanations \citep{bricken2023MonosemanticityDecomposingLanguage, gao2024ScalingEvaluatingSparse, paulo2025AutomaticallyInterpretingMillions}.
This paradigm has spurred a rich ecosystem, including online tools for interactive inspection \citep{neuronpedia} and downstream applications such as classification \citep{gallifant2025SparseAutoencoderFeatures} and specialized training for rare concepts \citep{muhamed2025DecodingDarkMatter}.
The quality of these automated explanations remains a known challenge:
benchmarking efforts have revealed difficulties in effective steering \citep{wu2025AxBenchSteeringLLMs}, and providing more context can adversely lead to vaguer explanations \citep{juang2024OpenSourceAutomated}.
More broadly, faithful interpretability calls for non-opaque means rather than relying on one LLM to interpret another \citep{huang2023RigorouslyAssessingNatural}.

Computational roles have so far been touched on only in passing.
Features have been treated as computational intermediates through input-specific attribution and ablation \citep{templeton2024ScalingMonosemanticityExtracting}, but that analysis remains in-context.
The closest prior work to ours is the ``output-centric'' analysis of \citep{gur-arieh2025EnhancingAutomatedInterpretability}, which projects the decoder onto the unembedding ($\mW_{\text{dec}}\mW_{\text{U}}$) to enrich an otherwise LLM-driven explanation pipeline; the authors discard the embedding pairing because it scored worse in their pipeline even on tied-weight models, and do not consider attention circuits.
We instead make the weight-based projections the primary unit of analysis: we retain the encoder--embedding pairing, where its divergence from the decoder--unembedding projection exposes a tied-versus-untied bifurcation between architectures (\Cref{sec:exp1}).
Intervention-based scoring has also been proposed as a complementary signal \citep{paulo2025AutomaticallyInterpretingMillions}, but it still serves to refine correlational explanations rather than to define them.

Two earlier observations underpin the present approach.
Next-token prediction was applied to vanilla LLM neurons in \cite{bills2023language}, where it generally failed to meet the baseline; the gap is consistent with polysemanticity, and motivates carrying the same idea to the disentangled SAE basis.
Logit-lens analysis of SAE features in \cite{bloom2024understandingfeatureslogitlens} confirmed that many features can be read as promoting or suppressing specific tokens, and the present paper picks up that thread by quantifying when a feature behaves this way versus when its decoder writes do not concentrate on any token at all.

\section{Framework}
\label{sec:framework}

Before the experiments, we consolidate the components shared across them: how we identify a feature, the projections we use to interrogate it, and the intervention primitive that turns a structural prediction into a causal test.

Given an activation $\vx \in \sR^{d_{\text{model}}}$ at one layer of the model, an SAE encodes and then decodes it as
\begin{equation}
  \vz = \sigma\left(\vx \mW_{\text{enc}} + \vb_{\text{enc}}\right), \qquad \hat{\vx} = \vz\mW_{\text{dec}} + \vb_{\text{dec}},
\end{equation}
where $\vz \in \sR^{d_{\text{sae}}}$ is the sparse feature vector and $\sigma$ is an activation function chosen to induce sparsity, so that most elements of $\vz$ are zero on any given input.
We index features by $i \in \{1, \dots, d_{\text{sae}}\}$. Prior work usually identifies a feature with its decoder row $\mW_{\text{dec}}^{(i)} \in \sR^{d_{\text{model}}}$ alone \citep{galichin2025HaveCoveredAll}. We instead treat a feature as the pair $(\mW_{\text{enc}}^{(\cdot,i)}, \mW_{\text{dec}}^{(i)})$ of an encoder column and decoder row, because the two play complementary roles in the forward pass: the encoder column \emph{reads from} the residual stream and gates whether the feature activates, and the decoder row \emph{writes into} the residual stream when it does. Reading and writing are distinct operations, and a feature may carry interpretable signal on either side independently.

\paragraph{Input and output token alignment.} The first pair of projections asks what tokens a feature is tied to, on the input side and on the output side:
\begin{align}
  \vl^i_{\text{E}} &= \mW_{\text{enc}}^{(\cdot,i)} \mW_{\text{E}}^{\top}     && \text{(input-token recognition)} \label{eq:enc-emb}\\
  \vl^i_{\text{U}} &= \operatorname{FinalNorm}(\mW_{\text{dec}}^{(i)})\, \mW_{\text{U}} && \text{(output-token prediction)} \label{eq:dec-unemb}
\end{align}
Here $\mW_{\text{E}} \in \sR^{|V| \times d_{\text{model}}}$ is the model's token embedding, $\mW_{\text{U}} \in \sR^{d_{\text{model}} \times |V|}$ is the unembedding, $|V|$ is the vocabulary size, and $\operatorname{FinalNorm}$ is the layer normalization the model applies before the unembedding reads. Both $\vl^i_{\text{E}}$ and $\vl^i_{\text{U}}$ are vectors of size $|V|$, one score per token.

Intuitively, $\vl^i_{\text{E}}$ asks which input tokens switch the feature \emph{on}: if a token's embedding sits along the direction the encoder reads, that token in the residual stream tends to drive feature $i$ to activate. Symmetrically, $\vl^i_{\text{U}}$ asks which output tokens the feature pushes the model toward: when the feature writes its decoder row into the residual stream, the unembedding turns that write into a logit increase for the tokens whose unembedding columns align with it. Together these are the bottom-up analogue of activation-based interpretation: rather than collecting prompts and asking what they share, we read the model's own embedding and unembedding to recover what the feature recognises and what it predicts.

\paragraph{Attention participation in the next layer.} The second pair of projections asks how a feature's decoder write is consumed by the very next attention layer:
\begin{align}
  \vq_i^{(h)} &= \text{AttnNorm}_{L+1}(\mW_{\text{dec},L}^{(i)}) \mW_{Q,L+1}^{(h)} && \text{(query participation in head $h$)} \label{eq:dec-q}\\
  \vk_i^{(h)} &= \text{AttnNorm}_{L+1}(\mW_{\text{dec},L}^{(i)}) \mW_{K,L+1}^{(h)} && \text{(key participation in head $h$)} \label{eq:dec-k}
\end{align}
Here $L$ indexes the SAE's host layer and $h$ indexes one head in the attention layer at $L+1$; $\mW_{Q,L+1}^{(h)}, \mW_{K,L+1}^{(h)} \in \sR^{d_{\text{model}} \times d_{\text{head}}}$ are that head's query and key projections; and $\text{AttnNorm}_{L+1}$ is the layer normalization applied before the head reads. The resulting $\vq_i^{(h)}$ and $\vk_i^{(h)}$ live in the head's $d_{\text{head}}$-dimensional QK space.

Intuitively, when the feature is active at a token position, its decoder row is written into the residual stream there; the next layer's head $h$ then reads that position through its query and key projections. $\vq_i^{(h)}$ measures how that write tilts the token's outgoing query, and $\vk_i^{(h)}$ measures how it tilts the token's incoming key. The first asks ``does this feature, when active at a position, change what that position is searching for?'' and the second asks ``does it change how attractive this position is as a target of attention from later positions?'' Token-level interpretation cannot see either of these effects: a feature can have a precise computational role in routing attention without ever lining up with a recognisable token on the input or output side.

\paragraph{Intervention primitive.} To turn a structural prediction made by any of the four projections into a causal claim, we use a single counterfactual operation \citep{pearl2009causality}. Given a feature $i$ at SAE layer $L$ on a prompt where it activates with value $z_i$, we set
\begin{equation}
  \mathrm{do}(z_i = 0)
  \label{eq:do-intervention}
\end{equation}
inside the SAE and re-decode, holding the rest of the forward pass fixed. Equivalently, this subtracts $z_i \mW_{\text{dec},L}^{(i)}$ once from the residual stream at layer $L$. Because every other component is untouched, any downstream change is attributable to the feature's own contribution. \Cref{sec:exp1,sec:exp2} pair the four projections with this primitive to test whether each structurally predicted effect is borne out causally.

\paragraph{Three points worth noting.}
The projections above depend on three modeling choices that attentive readers may reasonably question: (i) the per-feature application of $\operatorname{FinalNorm}$ and $\operatorname{AttnNorm}$ rather than the in-context normalization the model actually uses; (ii) whether the encoder--embedding and decoder--unembedding projections carry distinct signal, given that for tied-embedding models they would coincide if the encoder column and decoder row of a feature were themselves aligned; and (iii) the asymmetry that the do-intervention in \cref{eq:do-intervention} validates claims routed through the decoder write but is not the right test for the read-side encoder--embedding projection in \cref{eq:enc-emb}. \Cref{sec:framework-clarification} addresses each in turn, with the supporting empirical evidence (rank-correlation between normalized and unnormalized scores, encoder--decoder cosine statistics, and the scope of the do-operation).

\section{Experiment 1: Semantic Features}
\label{sec:exp1}
Activation-based interpretability assumes that each feature carries semantic meaning and derives a description from its top activating contexts.
Here we quantify token-level semantic alignment on the weight side, complementing context-based accounts with an analysis that requires no prompts or LLM annotators.
A feature qualifies as semantic under this operational definition when its decoder--unembedding projection $\vl^i_{\text{U}}$ from \Cref{eq:dec-unemb} or its encoder--embedding projection $\vl^i_{\text{E}}$ from \Cref{eq:enc-emb} concentrates on a coherent token cluster.
Quantifying this population is the prerequisite for examining how semantic alignment relates to computational participation in \Cref{sec:exp3}.

\subsection{Method}
We designed 23 candidate metrics (\Cref{sec:metrics}) and analyzed the pairwise correlations among them (\cref{fig:metrics-corr-9b}) to select a comprehensive set.
Ultimately, we chose the following 3 metrics, which demonstrate only moderate correlation
(mean absolute Spearman across layers $0.53 < \overline{|\rho|} < 0.69$ for Gemma-2-9B; see \cref{fig:metrics-scatter-9b} for an example of $\rho$ in layer 27),
thereby providing complementary perspectives on a feature's properties.
Throughout the experiments, semantic and non-semantic label features that pass and fail the joint criterion defined by the following 3 metrics:

\textbf{Levenshtein similarity.} Semantic features should have a low average Levenshtein distance \citep{levenshtein1965BinaryCC} among the top-10 predicted tokens, indicating that these tokens are mostly lexical variants of the same underlying concept (e.g., ``http'', ``HTTP'', ``https'', with or without a leading space).

\textbf{Cosine similarity.} Semantic features should have a high cosine similarity among the embeddings of the top-10 predicted tokens, indicating that the model represents them as closely related.
While the Levenshtein similarity reflects what humans expect, cosine similarity captures the model's perspective and complements the former.

\textbf{Top-100 Entropy.} Semantic features should have low entropy over logits, resulting in a ``spiky'' distribution where the feature's influence is decisive, often affecting only a single token.

With the three metrics in hand, the next question is how to use them to delimit a semantic population.
Each metric is naturally continuous with no apparent cutoff (see \cref{fig:metrics-dist-9b}), but setting a threshold per metric allows us to count, compare, and intervene on populations across layers and models at no additional inference cost.
This contrasts with the opaque alternative of using an LLM judge to directly label all features in all layers, which represents the activation-based approach we compare against.
The resulting classification depends on the chosen thresholds, so we assess its sensitivity alongside the cross-layer trends.

We calibrate the thresholds by manually inspecting different layers at various values, eventually setting each cutoff at the 50th percentile of the layer at 2/3 depth for the decoder--unembedding pairing (\cref{tab:exp1-thresholds}).
We propose this choice as a rule of thumb for future interpretation needs, as it provides sufficient depth to capture well-formed predictive features while avoiding the excessive specialization that occurs at later layers.
\Cref{fig:feature-token-sample-gemma9b-layer27} samples the resulting calibration-layer population.
For the encoder--embedding pairing, we calibrate an early layer of the encoder where the population is most informative.
We then apply these values at every other layer of the model to surface cross-layers trends.
We assess calibration sensitivity through a joint percentile sweep, keeping each calibration layer fixed (full setup in \Cref{sec:exp1-validation-thresholds}).
Finally, we employ an LLM judge to verify that our calibration aligns with expectations.

\subsection{Results}

\begin{figure}[t]
  \centering
  \begin{subfigure}{0.48\linewidth}
    \centering
    \includegraphics[width=\linewidth,trim={.0cm .0cm .0cm .0cm}]{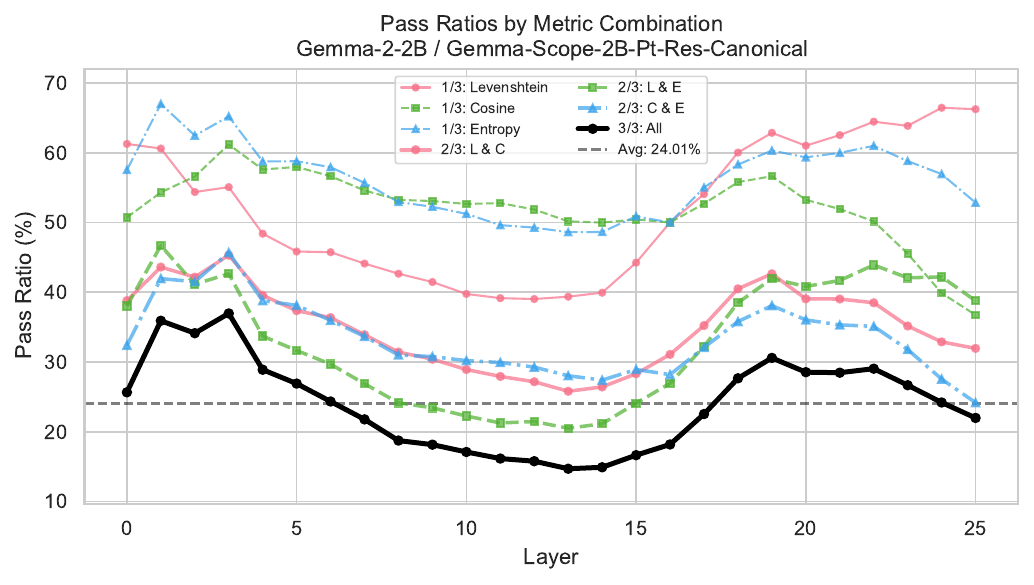}
  \end{subfigure}
  \begin{subfigure}{0.48\linewidth}
    \centering
    \includegraphics[width=\linewidth,trim={.0cm .5cm .0cm .0cm}]{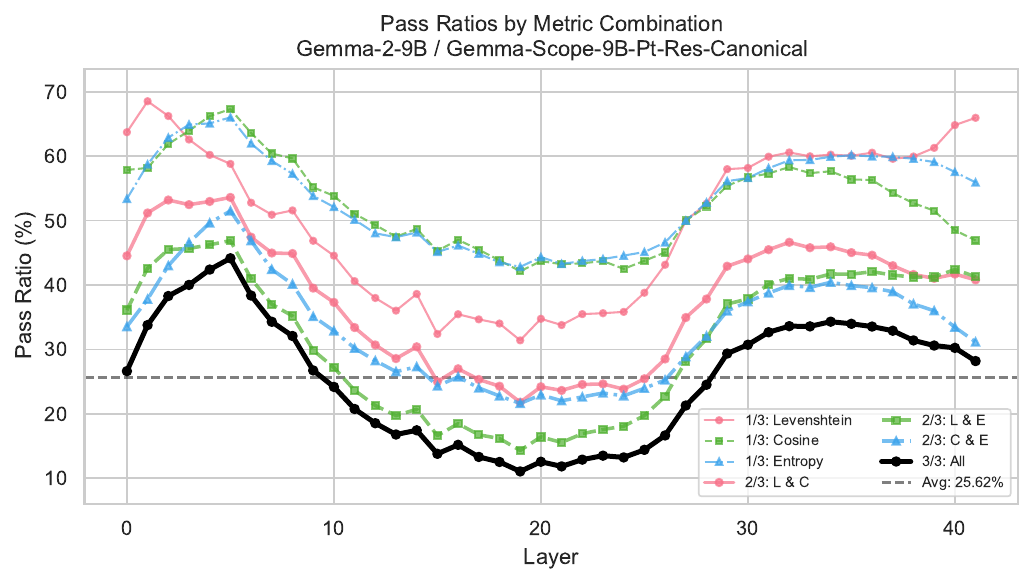}
  \end{subfigure}
  \caption{Joint and per-metric pass rates for the decoder--unembedding classifier on Gemma-2-2B and 9B. Both sizes show a U-shaped distribution; ablating any single metric uniformly increases the pass rate, confirming the three are complementary.}
  \label{fig:pass-rate-subsets-gemma}
\end{figure}

\begin{figure}[t]
  \centering
  \begin{subfigure}{0.48\linewidth}
    \centering
    \includegraphics[width=\linewidth,trim={.0cm .0cm .0cm .0cm}]{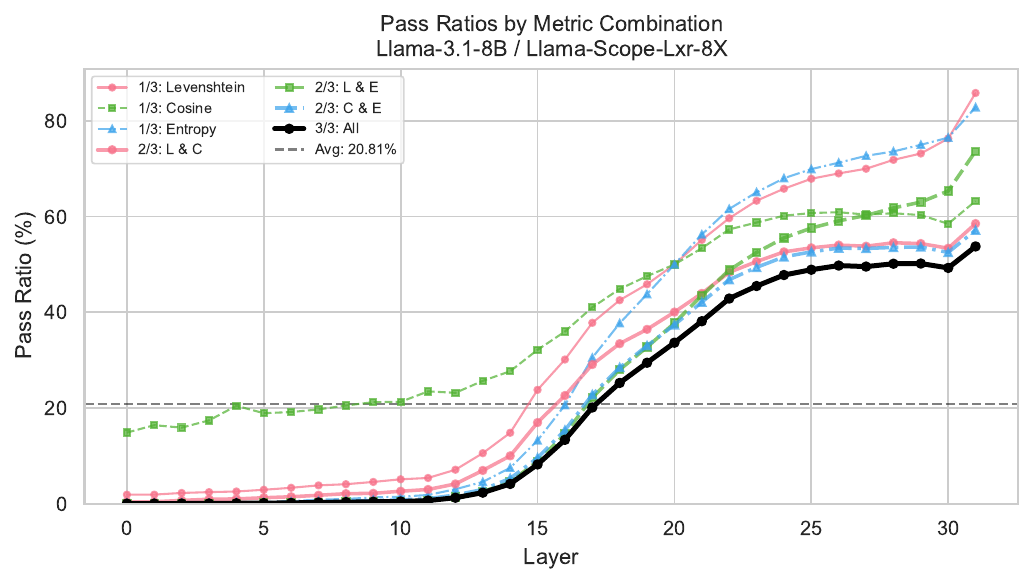}
  \end{subfigure}
  \begin{subfigure}{0.48\linewidth}
    \centering
    \includegraphics[width=\linewidth,trim={.0cm .5cm .0cm .0cm}]{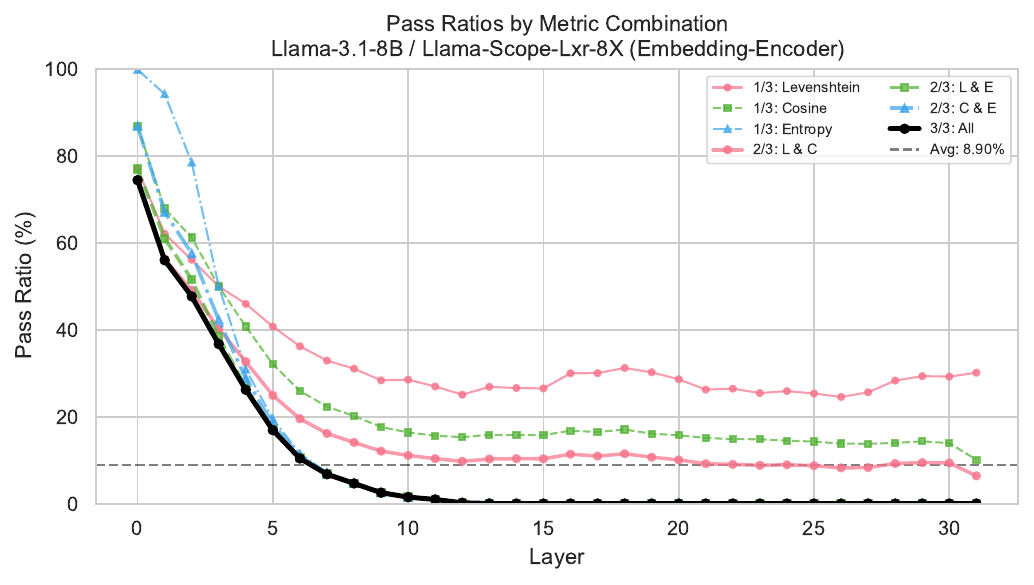}
  \end{subfigure}
  \caption{Pass rates for Llama-3.1-8B under the two projections of the dual analysis. The decoder--unembedding pairing (left) is near-zero in early layers and rises with depth; the encoder--embedding pairing (right) is high in early layers and falls. The two trajectories partition the network into an input-recognition regime and an output-prediction regime, each captured by a different projection.}
  \label{fig:pass-rate-subsets-llama}
\end{figure}

We apply both projections of the dual analysis ($\vl^i_{\text{U}}$ and $\vl^i_{\text{E}}$) to all three models using the 50th-percentile calibration.
For Gemma-2-2B and Gemma-2-9B (\cref{fig:pass-rate-subsets-gemma}), the decoder--unembedding projection identifies a substantial population of semantic features with mean joint pass rate of 24.01\% and 25.62\%, respectively.
The pass rate is high in the initial layers, decreases through the middle, and rises again toward the final layers, forming a U-shape across depth.
The encoder--embedding projection produces a quantitatively similar U-shape across both Gemma sizes (\cref{fig:pass-rate-subsets-gemma-enc}), as expected from the tied embedding $\mW_{\text{E}} = \mW_{\text{U}}^\top$ and the high alignment between the encoder and decoder vectors of Gemma 2 SAEs (\cref{fig:enc-dec-cosine}).

For Llama-3.1-8B (\cref{fig:pass-rate-subsets-llama}) the two projections diverge.
Under decoder--unembedding, the pass rate starts near zero in the initial layers and climbs steadily to over 50\% at the final layer.
Under encoder--embedding the pass rate is exceptionally high in the initial layers and declines rapidly, becoming negligible after the first third of the model.
The two projections, asked of the same SAE features, describe two different stages.
Early-layer features specialize in recognizing input tokens, visible to the encoder--embedding projection and invisible to the ``output-centric'' decoder--unembedding one, while late-layer features specialize in predicting output tokens.
Untied embeddings in the model shapes the SAE to allocate capacity to whichever side of the residual stream a feature serves.

The percentile sweep supports these architecture-dependent depth patterns: across all six model--projection combinations, the minimum Spearman correlation with the $p=50$ baseline is approximately $0.878$ over $p \in [40,60]$ (\Cref{tab:exp1-threshold-stability}).
The decoder--unembedding profiles remain strongly correlated over the wider range $p \in [25,75]$, with a minimum correlation of approximately $0.953$ across the three models.

\begin{table}[t]
  \centering
  \small
  \caption{Stability of the per-layer joint pass-rate profiles. Each $\rho_{\min}$ is the minimum Spearman correlation with the same cell's $p=50$ profile over the tested percentiles in the specified band. Calibration layers remain fixed throughout the sweep.}
  \label{tab:exp1-threshold-stability}
  \begin{tabular}{llccc}
  \toprule
  Projection & Model & Calibration layer & $\rho_{\min}$, $p \in [25,75]$ & $\rho_{\min}$, $p \in [40,60]$ \\
  \midrule
  Decoder--unembedding & Gemma-2-2B & 16 & 0.971 & 0.992 \\
   & Gemma-2-9B & 27 & 0.953 & 0.993 \\
   & Llama-3.1-8B & 20 & 0.971 & 0.997 \\
  \midrule
  Encoder--embedding & Gemma-2-2B & 3 & 0.720 & 0.878 \\
   & Gemma-2-9B & 6 & 0.905 & 0.978 \\
   & Llama-3.1-8B & 3 & 0.785 & 0.964 \\
  \bottomrule
  \end{tabular}
\end{table}

To validate the classifier's token-level semantic judgments, we compare its labels with those of Llama-3-70B-Instruct \cite{dubey2024Llama3Herd} on 1000 semantic and 1000 non-semantic features per layer, across 8 layers per (model, projection) pair, with the judge blinded to label (full setup in \Cref{sec:exp1-validation-judge}).
The judge's per-feature decisions agree with the classifier at mean Cohen's $\kappa = 0.83$ across all six cells (range 0.80--0.87; \Cref{tab:exp1-llm-judge}), which is substantial-to-near-perfect under the Landis--Koch bands, and the agreement holds uniformly across model sizes, the tied-vs-untied divide, and both projections despite the calibration layer and absolute thresholds varying.
This agreement supports using the inference-free classifier for large-scale analysis of token-level semantic alignment.

The projection-based criterion is a structural prediction.
To verify that semantic features influence the model on real inference passes as we predict, we apply $\mathrm{do}(z_i = 0)$ from \Cref{eq:do-intervention} to 100 semantic features per layer in Gemma-2-9B and Llama-3.1-8B, on each feature's max-activating prompt, and read the change in the next-token logit at the feature's top-1 associated token.
Each sampled feature is paired with a matched-norm non-semantic feature drawn from the same layer that fires at the same position, isolating coherent-token alignment as the only difference between the two interventions. Across 16 sweeps, removing the flagged feature produces a strictly larger $|\Delta\text{logit}|$ than its matched-norm baseline in 88--100\% of pairs, with median $|\Delta\text{logit}|$ of 0.8--7.1 versus 0.02--0.13 for the baseline.
\Cref{tab:exp1-intervention} reports the per-layer numbers, and the full setup is detailed in \Cref{sec:exp1-validation-intervention}.
The structural projection thus carries a causal signal on the model's output.

The judge validates token-cluster coherence, and the interventions confirm effects on the associated output tokens.
Taken together, our results identify a token-coherent subpopulation, challenging the assumption that semantic descriptions alone characterize what every feature does and motivating a complementary analysis of computational roles.

\section{Experiment 2: Computational Features}
\label{sec:exp2}
Experiment~1 identifies a token-coherent subpopulation under our weight-based criterion.
To characterize feature function more fully, we also examine how decoder directions participate in the model's internal computation.
Attention provides a concrete setting for this analysis: decoder directions project into the next layer's query and key circuits, which govern information routing between token positions.
We ask whether individual features exhibit specialized attention roles, concentrating on a small set of keys or attracting strong attention from multiple queries.
We assess this property across the full feature population, independently of the semantic classification, and trace its variation with depth and architecture.

\subsection{Method}

Given that SAEs are trained to reconstruct activations that participate in the forward pass, it is natural to expect that their features would inherit these computational roles.
The primary mechanism for information routing and composition within a transformer layer is the attention head \citep{vaswani2017AttentionAllYou}.
We therefore hypothesize that many of these features are specialized for participating in attention mechanisms.
In this experiment, we test this by measuring the alignment of SAE features from a given layer with the query and key (QK) circuits of the subsequent layer's attention heads.

The query and key participations $\vq_i^{(h)}$, $\vk_i^{(h)}$ defined in \cref{eq:dec-q,eq:dec-k} let us read off the pre-softmax attention score between feature $i$ as a query and feature $j$ as a key:
\begin{equation}
  s^{(h)}_{i,j} = \frac{\vq_i^{(h)} \cdot \vk_j^{(h)}}{\sqrt{d_{\text{head}}}} \quad\in \sR.
  \label{eq:sij}
\end{equation}
The matrix $\mS^{(h)} = (s_{i,j}^{(h)}) \in\rdim{sae,sae}$ records the raw QK alignment between every pair of features at head $h$ (derivation in \cref{sec:feature-attn}).

Pre-softmax scores expose raw participation strength but cannot say whether a feature wins a given query's competition: attention is row-wise normalized, so what matters downstream is the relative ordering of scores across keys. We therefore apply softmax row-wise to $\mS^{(h)}$, obtaining the simulated post-softmax attention matrix $\mA^{(h)} \in \rdim{sae,sae}$ whose entry $a_{i,j}^{(h)}$ is the fraction of attention feature $i$ (as a query) would allocate to feature $j$ (as a key) in head $h$.

We now define two complementary post-softmax criteria that consider both the query and key aspects to identify computationally specialized features.
Let $\mA^{(h)}_{i,\cdot}, \mA^{(h)}_{\cdot,j} \in \rdim{sae}$ denote the $i$-th row and the $j$-th column, respectively, of the post-softmax attention matrix $\mA^{(h)}$.

\textbf{Query-based specialization.}
A feature $i$ is classified as a \textit{query specialist} if, for at least one head $h$, the sum of its top-$k$ attention allocations exceeds threshold $\tau_Q$:
\begin{equation}
  \textstyle \sum_{j \in \mathcal{I}_k(\mA^{(h)}_{i,\cdot})} a_{i,j}^{(h)} > \tau_Q,
\end{equation}
where $\mathcal{I}_k(\vv) = \{j_1, \ldots, j_k\}$ denotes the index set of the $k$ largest components of vector $\vv$.
This identifies features that concentrate their attention on a small set of targets, acting as specialized ``lookers'' for certain keys.

\textbf{Key-based specialization.}
A feature $j$ is classified as a \textit{key hub} if, for at least one head $h$, the mean attention it receives from its top-$k$ attending queries exceeds threshold $\tau_K$.
Then:
\begin{equation}
  \textstyle \frac{1}{k}\sum_{i \in \mathcal{I}_k(\mA^{(h)}_{\cdot,j})} a_{i,j}^{(h)} > \tau_K.
\end{equation}
This identifies features that are emphasized by multiple queries, serving as information hubs.

\subsection{Results}

\begin{figure}[t]
  \centering
  \begin{subfigure}{0.48\linewidth}
    \centering
    \includegraphics[width=\linewidth,trim={.0cm .0cm .0cm .0cm}]{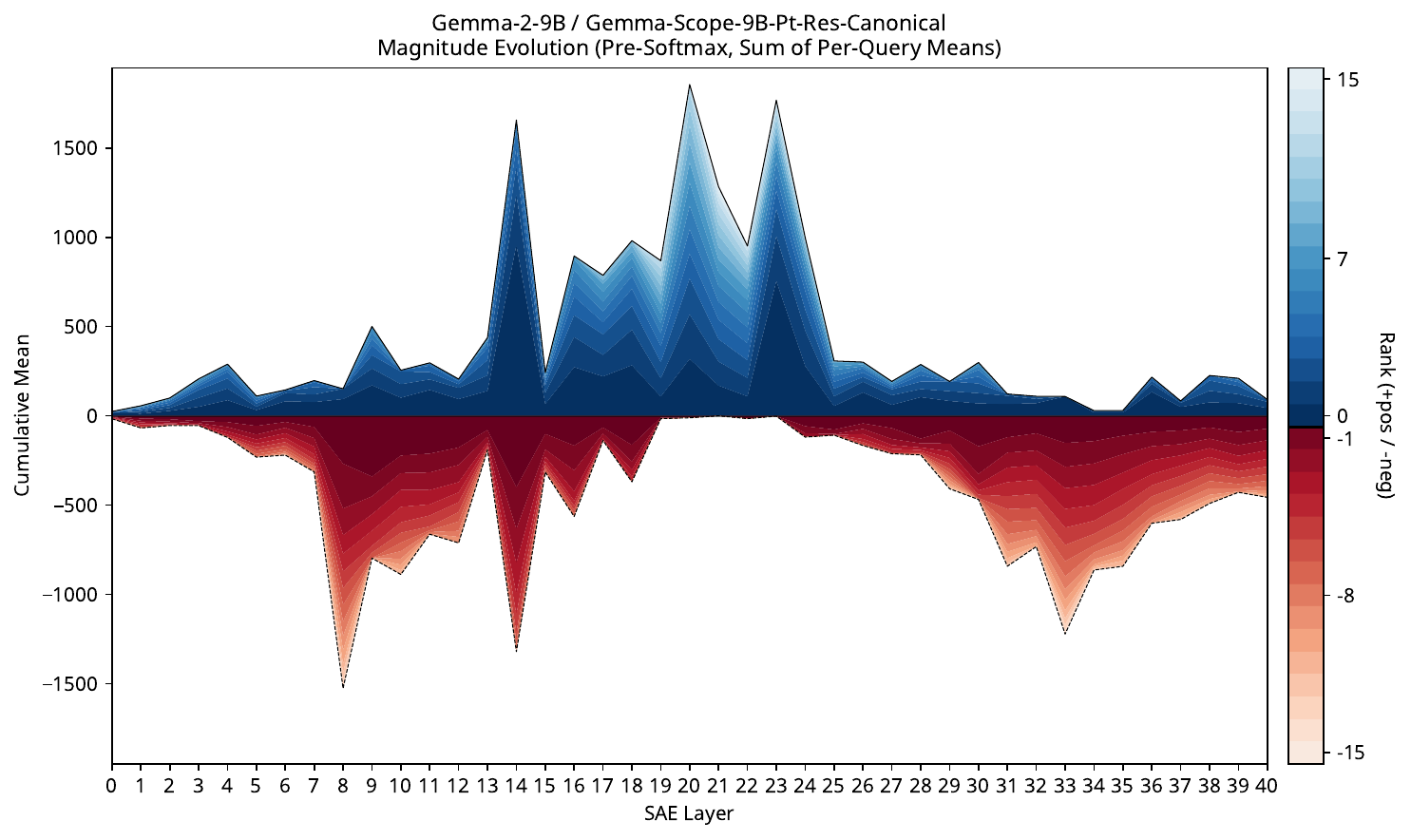}
  \end{subfigure}
  \begin{subfigure}{0.48\linewidth}
    \centering
    \includegraphics[width=\linewidth,trim={.0cm .0cm .0cm .0cm}]{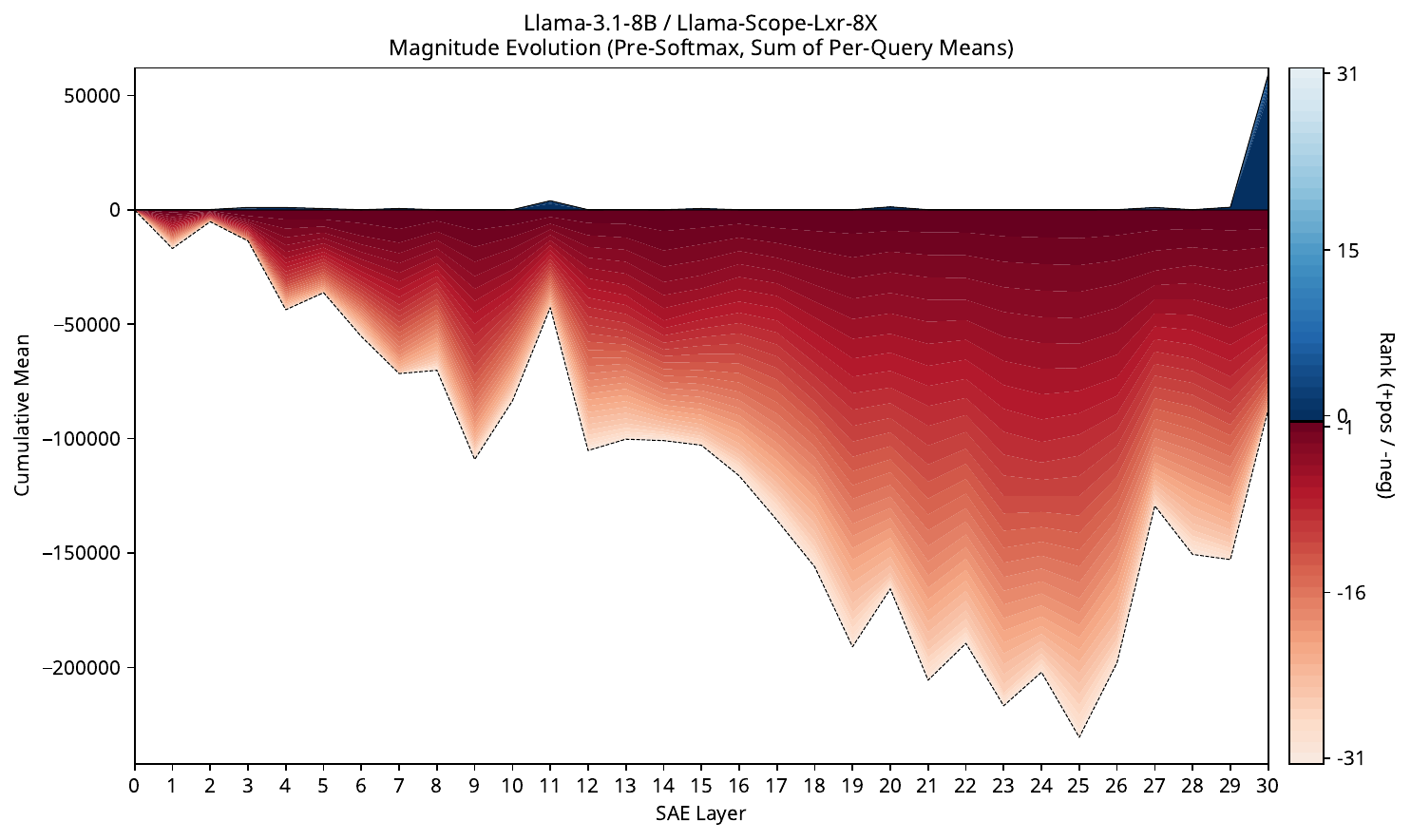}
  \end{subfigure}
  \caption{Main results for Experiment 2 on Gemma-2-9B and Llama-3.1-8B:
  Magnitude of attention participation across layers, with the total height at each layer representing the aggregate pre-softmax score.
  These scores are calculated as the mean across all keys for each query feature, then summed for each head.
  The stacking direction of the head values is determined based on whether they are positive or negative.
  Each color represents a rank of magnitude, rather than a specific head index, across all layers.}
  \label{fig:mag-evo-9b-8b}
\end{figure}

\textbf{Pre-softmax.}
To visualize the raw computational potential across layers, we first analyze the evolution of pre-softmax attention scores.
Positive pre-softmax values signify a tendency for constructive alignment within a head's QK circuit, while negative values suggest anti-alignment.

In Gemma-2-9B (\cref{fig:mag-evo-9b-8b} left) and Gemma-2-2B (\cref{fig:mag-evo-2b}), we observe a distinct low--high--low pattern in the aggregate attention potential.
The scores are negative in early and late layers and become strongly positive in the middle layers, mirroring the inverse of the semantic U-shape from Experiment 1.
This suggests that middle-layer features are indeed configured for constructive, high-potential interactions within QK circuits.

At first glance, Llama-3.1-8B (\cref{fig:mag-evo-9b-8b} right) appears to operate ``wrongly''.
Its aggregate pre-softmax scores are almost uniformly negative across all layers.
However, this is not necessarily a contradiction.
Consider that the softmax function is shift-invariant; only the relative differences between scores matter for the final attention distribution.
In line with this, the absolute magnitude of Llama's scores follows a similar low--high--medium pattern, peaking in the middle layers where computational activity is expected to be highest.
The pre-softmax magnitude evolution thus places both architectures on the same depth-dependent computational rhythm, even though Gemma achieves it via signed alignment and Llama via signed disalignment.

\textbf{Post-softmax.}
We set $k=10$, $\tau_Q = 0.95$, and $\tau_K = 0.25$.
These thresholds are chosen to surface clearly focused queries and clearly hub-like keys; qualitative depth profiles are stable to substantial loosening (e.g., $\tau_Q$ down to $0.25$, $\tau_K$ down to $0.05$; full rationale in \cref{sec:attn-rationale}), confirming that the shapes are architectural signatures.
Across all three models, both criteria identify non-trivial specialist populations at every depth (\cref{fig:feature-count-9b,fig:feature-count-2b,fig:feature-count-8b}); per-head mean counts remain low while the any-head union is large, indicating that each feature plays a specialist role only in a few specific head contexts rather than uniformly across heads.

A more architecture-revealing pattern emerges in how those populations are paced through depth.
In both Gemma-2 SAEs, an oscillatory pattern emerges with a 2-layer period (\cref{fig:feature-count-9b,fig:feature-count-2b}).
This period matches Gemma-2's documented alternation between local-window and full attention across consecutive layers \citep{team2024Gemma2Improving}.
The oscillation is absent in Llama-3.1-8B, confirming that an SAE's computational properties track the model architecture from which they originated.

\textbf{Intervention validation.}
We test the causal effect of the weight-identified specialists by applying \cref{eq:do-intervention} to flagged features on their max-activating prompt: removing a query specialist should perturb the attention row at the firing position in the flagged head, and removing a key hub should influence the incoming attention from queries later in the sequence.
Each flagged feature is paired with a non-specialist baseline matched on decoder norm and firing on the same prompt at the same position, isolating the structural property under inspection.
Across a total of 35 sweeps, each comprising up to 100 feature pairs, the flagged feature's effect on its target attention quantity exceeds its matched baseline by a median per-pair ratio of $43\!\times\!$--$2161\!\times\!$ (queries) and $9\!\times\!$--$37\!\times\!$ (keys), with win-rates of $0.93$--$1.00$ and $0.88$--$0.98$.
Our framework therefore identifies features that causally steer the next layer's attention during inference. Full setup and per-cell numbers are presented in \Cref{sec:exp2-validation}.

Together with \Cref{sec:exp1}, these results expose two populations of features at the resolution of single layers: a semantic population that writes into the embedding/unembedding path, and a computational population that writes into the next layer's QK circuits.

\section{Experiment 3: Semantic vs. Computational Specialization}
\label{sec:exp3}
Experiments~1 and~2 characterized semantic and computational features as independent properties. We now ask how the two populations relate: do semantic features show systematically different attention participation than non-semantic ones, and does that relationship vary with depth and architecture?
If the two roles are inversely related, semantic features should engage QK circuits less; if orthogonal, both populations should display similar attention distributions; if positively coupled in some layers, the same features would simultaneously serve interpretive and computational functions.

\subsection{Method}

For each feature $i$ at layer $L$, we summarize its participation in the attention of layer $L{+}1$ from two sides. As a query, we sum its top-$k$ pre-softmax scores over the keys it most strongly attends to, across all heads:
\begin{equation}
  w_i^Q = \sum_{h=1}^{n_{\text{heads}}} \sum_{j \in \mathcal{I}_k(\vs_i^{(h)})} s_{i,j}^{(h)},
\end{equation}
where $s_{i,j}^{(h)}$ is from \cref{eq:sij} and $\mathcal{I}_k(\cdot)$ selects the indices of the $k$ largest components. As a key, we sum the top-$k$ scores over the queries that most strongly attend to it:
\begin{equation}
  w_j^K = \sum_{h=1}^{n_{\text{heads}}} \sum_{i \in \mathcal{I}_k(\vs_{\cdot,j}^{(h)})} s_{i,j}^{(h)}.
\end{equation}
$w_i^Q$ thus measures how concentrated a feature's outgoing attention is when it acts as a query, and $w_j^K$ how concentrated the incoming attention is when it acts as a key. We use $k=10$ as in Experiment~2.

We then compare semantic features (passing all three metric thresholds) against non-semantic features per-layer via density estimates and boxplots, and across layers via mean, median, and the share of features exceeding the global $P_{75}$ threshold pooled across all layers.

\subsection{Results}

\begin{figure}[t]
  \centering
  \begin{subfigure}{0.48\linewidth}
    \centering
    \includegraphics[width=\linewidth,trim={.0cm .0cm .0cm .0cm}]{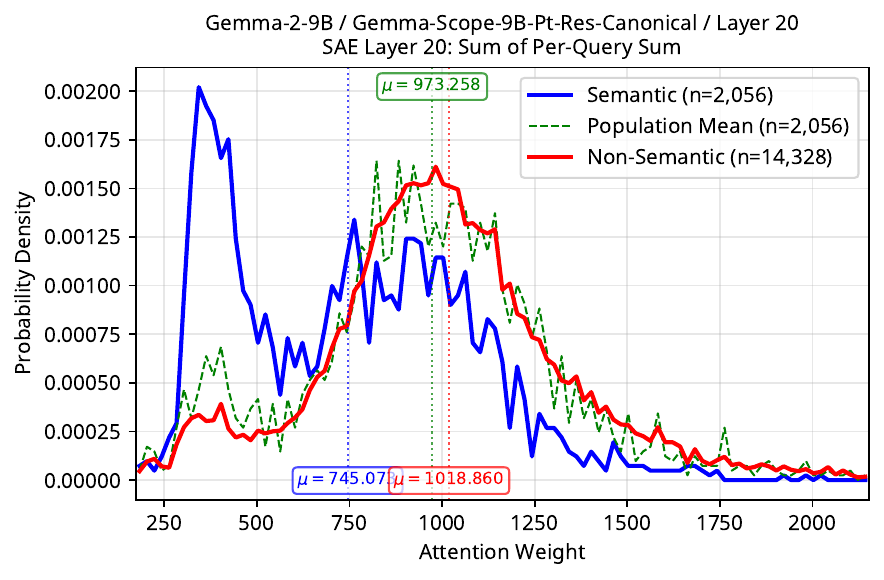}
  \end{subfigure}
  \begin{subfigure}{0.48\linewidth}
    \centering
    \includegraphics[width=\linewidth,trim={.0cm .3cm .0cm .2cm}]{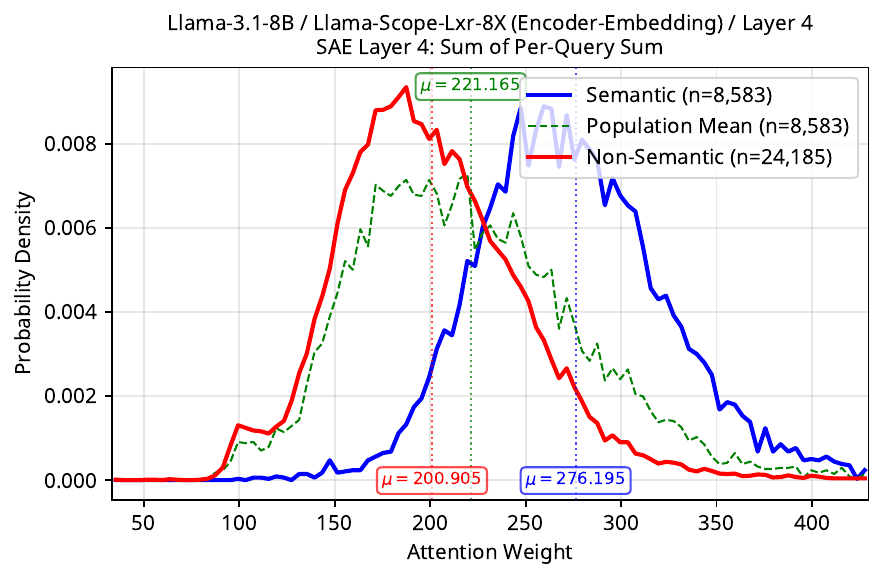}
  \end{subfigure}
  \caption{Per-layer query-side attention-weight densities for semantic vs.\ non-semantic features. Left: Gemma-2-9B layer 20, where the semantic distribution sits below the non-semantic one. Right: Llama-3.1-8B layer 4, in the input-semantic phase, where the relation inverts and semantic features carry the higher attention weights.}
  \label{fig:per-layer-density-9b-8b-small}
\end{figure}

\textbf{Per-layer distributions.}
The semantic and non-semantic populations overlap substantially, but the semantic distribution is shifted toward lower values (\cref{fig:per-layer-density-9b-8b-small}, left).
The non-semantic side carries a significantly heavier right tail, which the box-plot confirms (\cref{fig:per-layer-boxplot-9b}).

\textbf{Cross-layer trends.}
The picture sharpens when we move from individual layers to depth-wide aggregates.
In both Gemma-2-2B (\cref{fig:cross-layer-mean-2b,fig:cross-layer-median-2b}) and Gemma-2-9B (\cref{fig:cross-layer-mean-9b,fig:cross-layer-median-9b}), non-semantic features hold a mean/median lead across most layers, and the separation widens at the upper tail: the share of features exceeding the global $P_{75}$ threshold is markedly lower for semantic features (\cref{fig:cross-layer-p75-9b,fig:cross-layer-p75-2b}).
Non-semantic features therefore dominate the most computationally influential regime, and the effect reproduces across two Gemma sizes, indicating it is not an artifact.

\textbf{Llama's dual-phase architecture.}
The untied Llama-3.1-8B admits two semantically aligned subpopulations that occupy disjoint depth ranges.
We split it at layer~11, where the two pass-rate trajectories cross below 1\% (\cref{fig:pass-rate-subsets-llama}), and analyse each phase against their respective non-semantic baseline.
The output-semantic phase mirrors Gemma (\cref{fig:per-layer-density-8b-dec}):
semantic features sit below non-semantic features in mean (\cref{fig:cross-layer-mean-8b-dec}), median (\cref{fig:cross-layer-median-8b-dec}), and tail (\cref{fig:cross-layer-p75-8b-dec}).
The input-semantic phase, however, inverts the pattern: embedding-aligned semantic features can attain \emph{higher} attention weights than non-semantic features (\cref{fig:per-layer-density-9b-8b-small}, right).
A natural reading is that recognizing and encoding incoming tokens for downstream consumption is itself an attention-heavy job at the beginning of the layers.

These results refine the picture from Experiments~1 and~2.
The relationship between the two roles is not universally inverse.
It is shaped by depth, by the embedding--unembedding structure, and by whether a feature serves input encoding or output prediction.

\section{Conclusion}
We reframed SAE interpretation around the question \textit{what does a feature do?}, replacing prompts and LLM annotators with four weight-space projections and a single ablation primitive that turns each structural prediction into a causal test.
Reading the encoder and decoder against the model's own embedding, unembedding, and next-layer QK circuits surfaces three depth-dependent signatures that all trace back to the host architecture.
First, semantic depth profiles form a U-shape in tied-embedding Gemma and separate into input-recognition and output-prediction regimes in untied Llama.
Second, a population of features specializes in steering the next layer's attention, with a depth profile that tracks the host's computational schedule.
Third, whether semantics and computation compete or coincide depends on the purpose a feature serves and the trade-off involved in how features allocate their representational capacity to specialize.

Each of the three signatures is an architectural fingerprint apparent only from the weight side. Activation-based interpretation reads features one prompt at a time and cannot see this structure.
The two views answer different halves of the same question, and we call on future interpretability research to incorporate the weight-based aspects that have been neglected.

\appendix

\section{Interpretability Metrics}
\label{sec:metrics}

We introduce a comprehensive suite of interpretability metrics designed to quantify different aspects of SAE feature behavior based on their logit attribution patterns.

\subsection{Semantic Coherence Metrics}

\paragraph{Cosine Similarity} measures the mean pairwise cosine similarity between embeddings of the top-$k$ tokens. Let $\mathcal{T}_k = \{t_1, \ldots, t_k\}$ denote the top-$k$ tokens by attribution score, and $e_i \in \mathbb{R}^{d}$ their corresponding embeddings. We compute:
\begin{equation}
\text{CosineSim}_k = \frac{2}{k(k-1)} \sum_{i=1}^{k-1} \sum_{j=i+1}^{k} \frac{e_i \cdot e_j}{\|e_i\| \|e_j\|}
\end{equation}
Higher values indicate that top tokens occupy similar regions in embedding space, suggesting semantic coherence.

\paragraph{Levenshtein Similarity} quantifies string-level similarity between tokens using normalized edit distance. For each pair of tokens $(t_i, t_j)$ in the top-$k$ set:
\begin{equation}
\text{LevenSim}_k = \frac{2}{k(k-1)} \sum_{i=1}^{k-1} \sum_{j=i+1}^{k} \left(1 - \frac{d_{\text{Lev}}(t_i, t_j)}{\max(|t_i|, |t_j|)}\right)
\end{equation}
where $d_{\text{Lev}}$ is the Levenshtein distance and $|t|$ denotes string length. This metric captures syntactic patterns such as shared morphological features.

\paragraph{String Overlap} measures the fraction of unique normalized tokens in the top-$k$ set. After removing non-word characters and lowercasing, we compute:
\begin{equation}
\text{Overlap}_k = 1 - \frac{|\{\text{normalize}(t) : t \in \mathcal{T}_k\}|}{k}
\end{equation}
Higher values indicate repeated tokens (possibly with different capitalization or punctuation).

\subsection{Distribution Concentration Metrics}

\paragraph{Entropy} quantifies the concentration of attribution mass across the vocabulary. Given attribution scores $s = (s_1, \ldots, s_V)$ for vocabulary size $V$, we compute the Shannon entropy of the softmax distribution:
\begin{equation}
H_k(s) = -\sum_{i \in \text{top-}k} p_i \log p_i, \quad p_i = \frac{\exp(s_i/\tau)}{\sum_{j \in \text{top-}k} \exp(s_j/\tau)}
\end{equation}
where $\tau$ is temperature (default 1.0). Lower entropy indicates more concentrated attribution, suggesting clearer feature semantics. We compute this at multiple scales: top-10, top-25, top-50, top-100, and full vocabulary.

\paragraph{Mass Concentration} measures the fraction of total positive attribution mass contained in the top-$k$ tokens:
\begin{equation}
\text{Mass}_k = \frac{\sum_{i=1}^{k} s_i}{\sum_{j : s_j > 0} s_j}
\end{equation}
Higher values indicate that a small number of tokens account for most of the feature's impact.

\paragraph{Gini Coefficient} quantifies inequality in the score distribution. For top-$k$ scores sorted in ascending order $(s_{(1)}, \ldots, s_{(k)})$:
\begin{equation}
G_k = \frac{2\sum_{i=1}^{k} i \cdot s_{(i)}}{k \sum_{i=1}^{k} s_{(i)}} - \frac{k+1}{k}
\end{equation}
Values range from 0 (perfect equality) to 1 (maximal inequality). Higher values indicate sparser, more concentrated attribution.

\begin{figure}[t]
  \centering
  \includegraphics[width=\linewidth,trim={.0cm .0cm .0cm .0cm}]{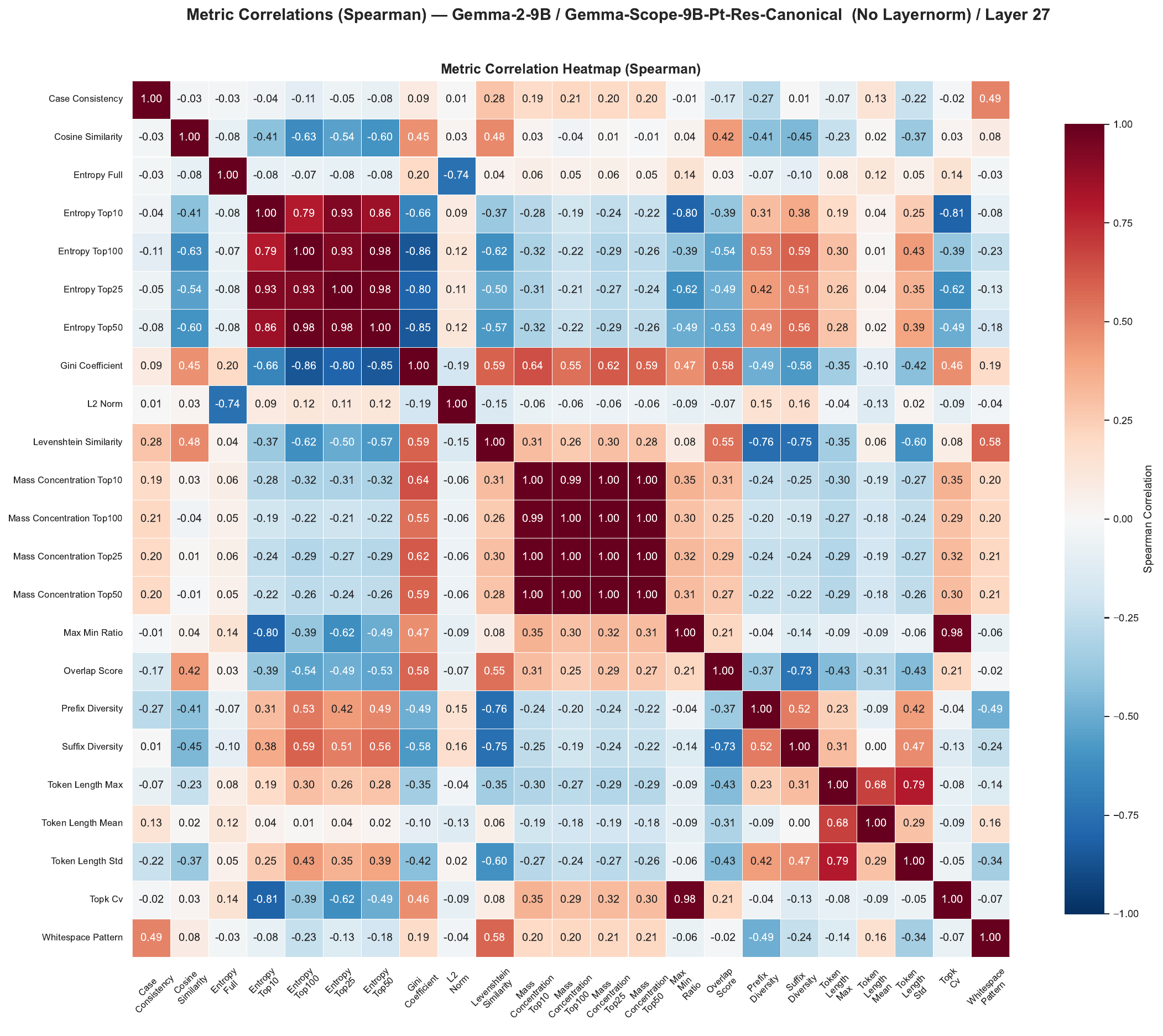}
  \caption{Spearman rank correlation ($\rho$) matrix for 23 candidate semantic metrics, calculated on all features from the Gemma-2-9B layer 27 SAE.}
  \label{fig:metrics-corr-9b}
\end{figure}

\paragraph{Coefficient of Variation (CV)} measures relative dispersion of the top-$k$ scores:
\begin{equation}
\text{CV}_k = \frac{\sigma(\{s_i|i\in \text{top-}k\})}{\mu(\{s_i|i\in \text{top-}k\})}
\end{equation}
where $\sigma$ and $\mu$ denote standard deviation and mean. Lower values indicate more uniform score magnitudes among top tokens.

\paragraph{Max/Min Ratio} captures the dominance of the top-scoring token:
\begin{equation}
\text{Ratio}_k = \frac{s_1}{s_k}
\end{equation}
where $s_1$ is the highest score and $s_k$ is the $k$-th highest. Higher values indicate a single dominant token.

\begin{figure}[ht]
  \centering
  \includegraphics[width=\linewidth,trim={.0cm .0cm .0cm .0cm}]{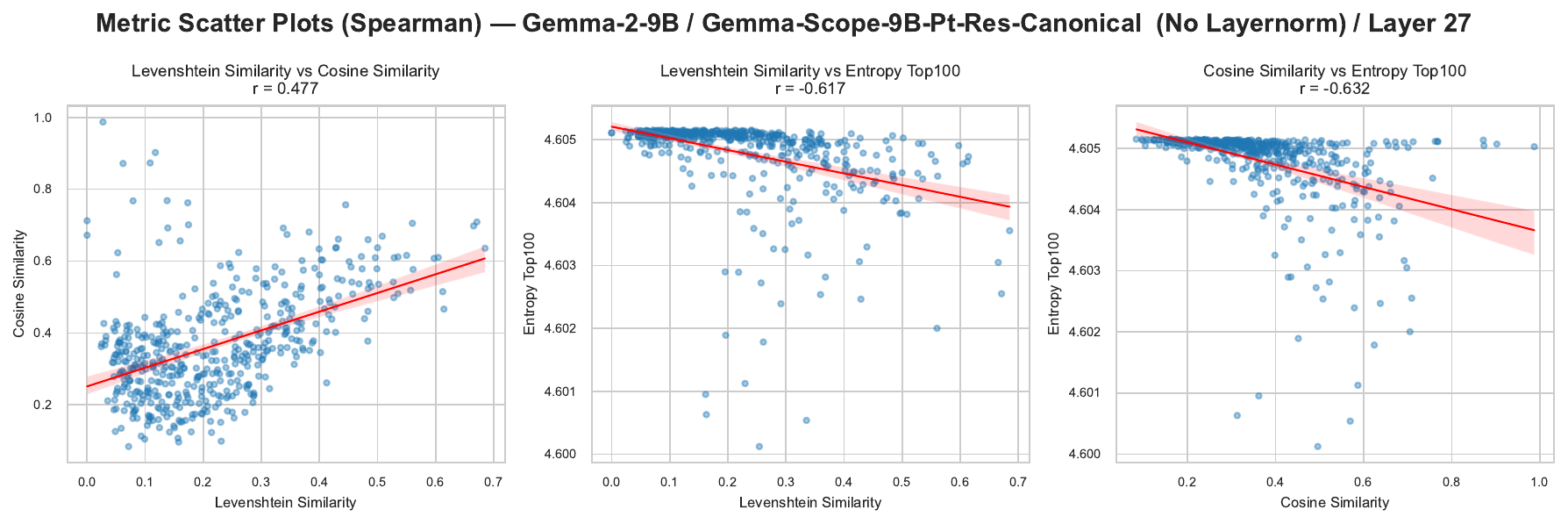}
  \caption{
  Pairwise scatter samples illustrating the relationships between the three selected semantic metrics for the Gemma-2-9B layer 27 SAE.
  The plots demonstrate that while the metrics are not entirely independent, their correlations are moderate.
  This supports their selection as a complementary set, where each metric provides a distinct perspective for evaluating the semantic coherence of a feature.
  }
  \label{fig:metrics-scatter-9b}
\end{figure}

\begin{figure}[ht]
  \centering
  \includegraphics[width=\linewidth,trim={.0cm .0cm .0cm .0cm}]{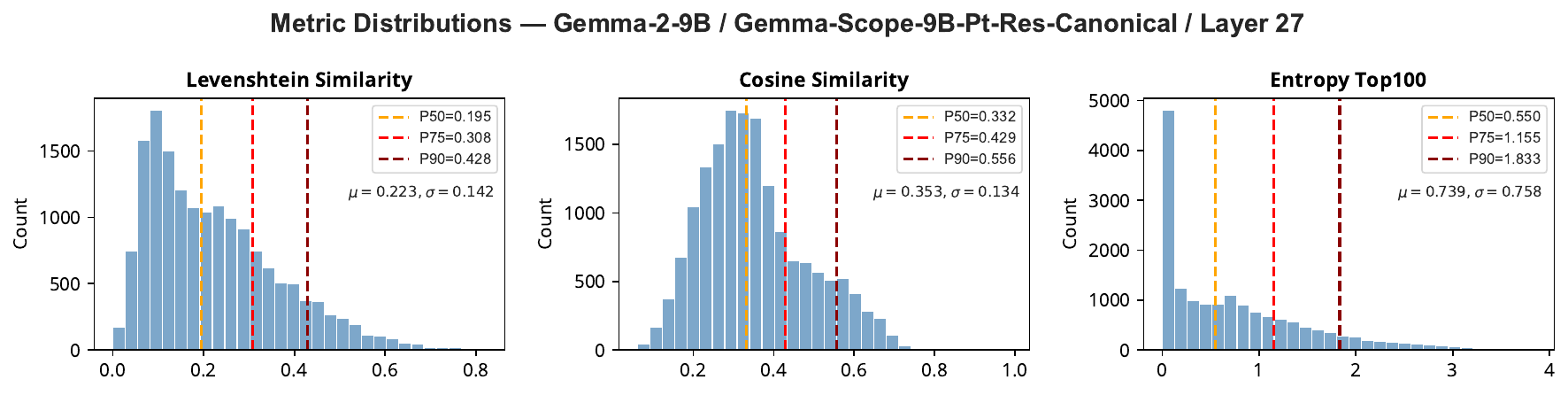}
  \caption{
  Distributions of the three primary semantic metrics---Levenshtein similarity (higher is better), cosine similarity (higher is better), and top-100 entropy (lower is better)---for all features in the Gemma-2-9B layer 27 SAE.
  The distributions are continuous and generally unimodal, lacking a distinct bimodal structure or natural cutoff point.
  This characteristic motivates our use of a percentile-based thresholding approach in Experiment 1, as it provides a consistent relative standard for what constitutes a ``semantic'' feature, rather than relying on an arbitrary absolute value.
  The distribution plot applies normalization (RMSNorm) because, without it, the entropy is highly concentrated near the maximum value with extreme outliers, making visualization difficult.
  }
  \label{fig:metrics-dist-9b}
\end{figure}

\paragraph{L2 Norm} measures the overall magnitude of feature impact, normalized by vocabulary size:
\begin{equation}
\|s\|_{\text{norm}} = \frac{\|s\|_2}{\sqrt{V}}
\end{equation}
This provides scale-independent comparison of attribution strength across features.

\subsection{Morphological Pattern Metrics}

\paragraph{Prefix/Suffix Diversity} quantify the variety of token beginnings and endings. For prefix length $\ell$ (default 3):
\begin{equation}
\text{PrefixDiv}_k = \frac{|\{t[:\ell] : t \in \mathcal{T}_k\}|}{k}
\end{equation}
and analogously for suffixes using $t[-\ell:]$. Lower diversity suggests morphological consistency (e.g., shared verb conjugations).

\paragraph{Token Length Statistics} include mean, standard deviation, and maximum character length across top-$k$ tokens. These capture whether features specialize to tokens of particular lengths.

\paragraph{Case Consistency} measures the fraction of top-$k$ tokens sharing the dominant case pattern (lowercase, uppercase, title case, or mixed):
\begin{equation}
\text{CaseConsist}_k = \frac{\max_{c \in \{\text{lower, upper, title, mixed}\}} |\{t \in \mathcal{T}_k : \text{case}(t) = c\}|}{k}
\end{equation}

\paragraph{Whitespace Pattern} computes the fraction of top-$k$ tokens beginning with whitespace, capturing formatting conventions:
\begin{equation}
\text{Whitespace}_k = \frac{|\{t \in \mathcal{T}_k : t[0] \in \text{whitespace}\}|}{k}
\end{equation}

To avoid redundancy and select a minimal set of metrics that provide complementary perspectives, we performed a correlation analysis.
We computed the pairwise Spearman rank correlation coefficient ($\rho$) among all 23 metrics for the features in a representative SAE (Gemma-2-9B, layer 27).
As illustrated in \cref{fig:metrics-corr-9b}, this analysis revealed distinct clusters of highly correlated metrics, indicating that many captured similar underlying properties.

\section{Validating the Semantic Population}
\label{sec:exp1-validation}

This appendix evaluates the three-criteria semantic classifier introduced in \Cref{sec:exp1} through three complementary checks: threshold sensitivity, LLM-judge agreement, and decoder-write interventions.
The threshold sweep tests how the population size and depth profile depend on calibration.
The judge reads each feature's top logit-lens tokens, blinded to label, to assess token-cluster coherence.
The $\mathrm{do}(z_i = 0)$ intervention tests whether the projection-based prediction corresponds to a downstream behavioral effect.

\subsection{Threshold sensitivity}
\label{sec:exp1-validation-thresholds}

\paragraph{Setup.}
We introduce a joint strictness percentile $p \in \{25,30,\ldots,75\}$ and keep the calibration layer fixed for each (model, projection) cell, using the same layers as in \Cref{tab:exp1-thresholds}.
The Levenshtein-similarity and cosine-similarity cutoffs are set at percentile $P_p$ of their respective calibration-layer distributions, and the top-100 entropy cutoff is set at $P_{100-p}$.
Thus, increasing $p$ makes all three criteria stricter, and $p=50$ reproduces the paper's calibration.
At every sweep point we apply the resulting cutoffs to every layer and count features passing all three criteria.
\Cref{tab:exp1-threshold-sweep} reports the arithmetic mean of these per-layer pass rates.
For each cell, we also compute the Spearman correlation between the per-layer pass-rate profile at each $p$ and its $p=50$ baseline, summarizing the minimum within each tested percentile band in \Cref{tab:exp1-threshold-stability}.

\paragraph{Results.}
The threshold sweep supports the architecture-dependent depth patterns identified in Experiment 1.
Around the chosen calibration, Gemma retains its mid-depth decoder--unembedding troughs, and Llama retains the contrasting depth trends of its encoder--embedding and decoder--unembedding profiles.
Across all six model--projection combinations, the minimum Spearman correlation with the $p=50$ baseline is approximately $0.878$ over $p \in [40,60]$.
The decoder--unembedding profiles remain highly correlated over the wider range $p \in [25,75]$, with a minimum correlation of approximately $0.953$ across the three models.

\begin{table}[ht]
  \centering
  \small
  \caption{Threshold sensitivity of the three-criteria classifier. Entries are mean joint pass rates (\%) across all layers. The joint strictness percentile $p$ raises both similarity cutoffs and lowers the entropy cutoff; the bold row reproduces the paper's calibration.}
  \label{tab:exp1-threshold-sweep}
  \begin{tabular}{rcccccc}
  \toprule
  & \multicolumn{3}{c}{Decoder--unembedding} & \multicolumn{3}{c}{Encoder--embedding} \\
  \cmidrule(lr){2-4}\cmidrule(lr){5-7}
  $p$ & Gemma-2-2B & Gemma-2-9B & Llama-3.1-8B & Gemma-2-2B & Gemma-2-9B & Llama-3.1-8B \\
  \midrule
  25 & 50.4 & 50.5 & 33.4 & 48.8 & 44.9 & 12.5 \\
  30 & 43.5 & 43.8 & 30.8 & 42.1 & 39.0 & 11.5 \\
  35 & 37.4 & 38.2 & 28.3 & 36.8 & 34.5 & 10.7 \\
  40 & 32.3 & 33.7 & 25.8 & 32.2 & 30.8 & 10.1 \\
  45 & 28.0 & 29.4 & 23.4 & 28.3 & 27.8 & 9.5 \\
  \textbf{50} & \textbf{24.0} & \textbf{25.6} & \textbf{20.8} & \textbf{24.7} & \textbf{25.2} & \textbf{8.9} \\
  55 & 20.8 & 22.1 & 18.1 & 20.3 & 22.2 & 8.2 \\
  60 & 17.9 & 18.9 & 15.4 & 13.3 & 17.3 & 7.2 \\
  65 & 15.2 & 15.9 & 12.4 & 7.6 & 11.1 & 5.7 \\
  70 & 12.6 & 12.9 & 9.6 & 4.5 & 7.1 & 4.3 \\
  75 & 9.9 & 10.0 & 6.9 & 2.7 & 4.5 & 3.2 \\
  \bottomrule
  \end{tabular}
\end{table}

\subsection{LLM-judge agreement at scale}
\label{sec:exp1-validation-judge}

The three-criteria classifier in \Cref{sec:exp1} is calibrated by taking each metric's 50th percentile over the feature population at one representative layer (the model's 2/3-depth for the decoder--unembedding pairing; see \Cref{tab:exp1-thresholds}) and reusing those three numbers at every other layer. The thresholds are then validated by manual inspection of sampled features. To check that this inference-free decision rule scales beyond our manual inspection, we ask an external reader to classify the same features under the same definition.

\paragraph{Setup.} For each (model, projection, layer) cell we draw 1000 strictly-passing features (passing all three thresholds) and 1000 strictly-failing features (failing all three). We render each feature's top-5 logit-lens tokens with leading-space markers preserved, present them to Llama-3-70B-Instruct \cite{dubey2024Llama3Herd}, and ask whether the five tokens collectively express one coherent meaning. The judge sees the token list only. Outputs are restricted to \{yes, no\}, with six fixed few-shot exemplars and temperature 0. We aggregate the per-feature verdicts to per-layer sensitivity, specificity, agreement, and Cohen's $\kappa$, and average across the eight sampled layers within each cell.

\begin{figure}[t]
  \centering
  \input{tables/exp1-judge-prompt}
  \caption{LLM-judge prompt for the semanticity check. The system message defines what counts as a coherent token cluster, six few-shot exemplars (three positive, three negative) anchor the decision boundary, and the final user turn presents the feature's top-5 logit-lens tokens for verdict. The judge is restricted to \{yes, no\} at temperature 0. Tokens are rendered with their leading-space markers preserved so the judge sees the same surface form as the model's vocabulary.}
  \label{fig:judge-prompt}
\end{figure}

\paragraph{Result.} \Cref{tab:exp1-llm-judge} reports the six (model, projection) cells. Mean $\kappa$ ranges from 0.80 to 0.87 across cells, with both Gemma sizes and both Llama projections falling inside that band. The overall aggregate sits at $\kappa = 0.83$. By the Landis--Koch bands, this is substantial-to-near-perfect agreement.

Two key takeaways emerge from these findings. First, the percentile-calibrated classifier proves to be effective: a capable LLM that reads only the same logit-lens tokens reproduce its decisions reliably at scale. Second, the agreement remains consistent across different model sizes and both projection methods, despite variations in the calibration layer and absolute thresholds across cells (see \Cref{tab:exp1-thresholds}). This consistency suggests that the classifier can be extended to resource-constrained settings where running an LLM judge would be prohibitively expensive, such as when conducting comprehensive feature sweeps across every layer.

\begin{table}[ht]
  \centering
  \caption{LLM-judge agreement with the three-criteria semantic classifier. For each (model, projection) cell we sample 1000 strict-pass and 1000 strict-fail features at every listed layer (top-5 logit-lens tokens shown to the judge, blinded to label). Sensitivity, specificity, agreement, and Cohen's $\kappa$ are computed per layer and averaged within the cell. Judge: Llama-3-70B-Instruct.}
  \label{tab:exp1-llm-judge}
  \resizebox{\linewidth}{!}{
    \begin{tabular}{l|lccccc}
    \toprule
    Projection & Model & Layers & Sensitivity & Specificity & Agreement & Cohen's $\kappa$ \\
    \midrule
  \multirow{3}{*}{\begin{tabular}[c]{@{}l@{}}Decoder--\\Unembedding\end{tabular}}
      & Gemma-2-2B & 2, 5, 8, 11, 14, 17, 20, 23 & 0.916 & 0.908 & 0.912 & 0.824 \\
      & Gemma-2-9B & 5, 10, 15, 20, 25, 30, 35, 40 & 0.954 & 0.862 & 0.908 & 0.816 \\
      & Llama-3.1-8B & 16, 18, 20, 22, 24, 26, 28, 30 & 0.889 & 0.925 & 0.907 & 0.814 \\
    \midrule
  \multirow{3}{*}{\begin{tabular}[c]{@{}l@{}}Encoder--\\Embedding\end{tabular}}
      & Gemma-2-2B & 2, 5, 8, 11, 14, 17, 20, 23 & 0.916 & 0.952 & 0.934 & 0.868 \\
      & Gemma-2-9B & 5, 10, 15, 20, 25, 30, 35, 40 & 0.950 & 0.924 & 0.937 & 0.874 \\
      & Llama-3.1-8B & 1, 2, 3, 4, 5, 6, 8, 10 & 0.801 & 0.995 & 0.898 & 0.796 \\
    \bottomrule
    \end{tabular}
  }
\end{table}

\subsection{Causal effect of the decoder write}
\label{sec:exp1-validation-intervention}

The judge confirms that semantic features carry coherent token-level meaning at the projection level. We now investigate whether removing such a feature influences the model's decision on those tokens, using the intervention primitive defined in \Cref{eq:do-intervention}.

\paragraph{Setup.}
For each sampled SAE layer, we draw 100 features that pass all three semantic criteria and have a decoder norm above the per-layer 5th percentile, ensuring the comparison is not dominated by weak features. Each flagged feature $f$ is paired with a \emph{matched-norm baseline} $f_b$ drawn from the same layer: $f_b$ fails all three criteria (strict non-semantic), has decoder norm within $\pm 5\%$ of $f$'s, and fires on the same prompt at the same position.
The baseline therefore removes a non-semantic write of similar size to $f$'s, on the same forward pass, at the same residual-stream position. Each flagged feature is tested on its top max-activating prompt from Neuronpedia \cite{neuronpedia}, truncated at the position where the feature peaks, so the last sequence position is exactly where the feature fires. We then read the next-token logits at the feature's top-1 logit-lens token under a)~the clean forward pass and b)~the intervened forward pass with $\mathrm{do}(z = 0)$ applied to that one feature inside the SAE. The reported metrics are $|\Delta\mathrm{logit}|$ at the top-1 target and the full-vocabulary KL divergence between the clean and intervened next-token distributions.

\paragraph{Result.} \Cref{tab:exp1-intervention} shows the 16 sweeps. The flagged feature's removal moves the target token's logit by 0.8--7.1 (median, log-space units), against 0.02--0.12 for its matched-norm non-semantic neighbour at the same layer on the same prompt. The win-rate is the fraction of pairs in which the flagged feature produces a strictly larger $|\Delta\mathrm{logit}|$ than its baseline. It ranges from 0.88 to 1.00 across the 16 sweeps. The KL effect tracks $|\Delta\mathrm{logit}|$ but at much larger ratios: median full-vocabulary KL is 0.02--0.38 for flagged removals and $1\!\!\times\!\!10^{-4}$--$5\!\!\times\!\!10^{-3}$ for matched removals.
The stark comparison clearly demonstrates the causal downstream effects of the features identified by our method on their semantic tokens.

\begin{table}[ht]
  \centering
  \caption{Intervention validation for Experiment 1. For each sampled SAE layer we draw 100 flagged features and pair each with a matched-norm baseline: a strict-non-semantic feature whose decoder norm is within $\pm 5\%$ of the flagged feature's and that fires on the same prompt. We apply $\mathrm{do}(z=0)$ to each and read the change in the next-token logit at the flagged feature's top-1 logit-lens token. KL is over the full vocabulary. $n_{\mathrm{m}}$ is the number of pairs for which a matched baseline was available.}
  \label{tab:exp1-intervention}
  \begin{tabular}{l|cc|cc|cc|c}
  \toprule
  \multirow{2}{*}{Model} & \multirow{2}{*}{Layer} & \multirow{2}{*}{$n_{\mathrm{m}}$} & \multicolumn{2}{c|}{Median $|\Delta\text{logit}|$} & \multicolumn{2}{c|}{Median KL} & Win-rate \\
   &  &  & semantic & matched & semantic & matched & $|\Delta|$ \\
  \midrule
  Gemma-2-9B & 5 & 89 & 1.359 & 0.0391 & 0.0359 & 0.0001 & 0.92 \\
   & 10 & 99 & 0.988 & 0.0156 & 0.0156 & 0.0001 & 0.96 \\
   & 15 & 99 & 1.187 & 0.0312 & 0.0310 & 0.0001 & 0.92 \\
   & 20 & 93 & 0.818 & 0.0391 & 0.0230 & 0.0002 & 0.97 \\
   & 25 & 97 & 1.227 & 0.0469 & 0.0518 & 0.0003 & 0.97 \\
   & 30 & 96 & 2.133 & 0.0469 & 0.1012 & 0.0003 & 0.95 \\
   & 35 & 93 & 3.461 & 0.0469 & 0.2306 & 0.0003 & 1.00 \\
   & 40 & 83 & 7.129 & 0.0312 & 0.3782 & 0.0005 & 1.00 \\
  \midrule
  Llama-3.1-8B & 16 & 92 & 1.212 & 0.1094 & 0.0677 & 0.0035 & 0.92 \\
   & 18 & 80 & 1.879 & 0.1055 & 0.1302 & 0.0023 & 0.96 \\
   & 20 & 79 & 2.253 & 0.0781 & 0.1833 & 0.0034 & 0.97 \\
   & 22 & 65 & 2.926 & 0.0938 & 0.3066 & 0.0032 & 0.97 \\
   & 24 & 62 & 3.744 & 0.1035 & 0.1918 & 0.0048 & 0.95 \\
   & 26 & 57 & 3.596 & 0.1250 & 0.2571 & 0.0023 & 0.96 \\
   & 28 & 54 & 4.023 & 0.1035 & 0.2206 & 0.0019 & 0.94 \\
   & 30 & 51 & 4.643 & 0.0938 & 0.1227 & 0.0017 & 0.88 \\
  \bottomrule
  \end{tabular}
\end{table}

\section{Feature-Level Attention Analysis}
\label{sec:feature-attn}

We now formalize the relationship between our weight-based feature-feature attention scores and the actual attention computations in the forward pass.

In our experiments, we apply normalization to individual feature contributions, following the convention used in prior work employing the logit lens.
In contrast, during the model’s actual forward pass, normalization operates on the full combination of features prior to attention computation.
Applying normalization at the level of individual features is therefore an approximation, since normalization is a non-linear operation.
For analytical clarity, the derivation below omits normalization.

\subsection{SAE Reconstruction of the Residual Stream}

In the actual forward pass, the residual stream at the end of layer $L$ is:
\begin{equation}
  \vx_{\text{true}} = \vx_{\text{SAE}} + \vx_{\text{error}}
\end{equation}
where $\vx_{\text{SAE}}$ is the SAE reconstruction and $\vx_{\text{error}}$ is the reconstruction error.
Our analysis focuses on $\vx_{\text{SAE}}$, which for well-trained SAEs captures the dominant systematic structure.

When features activate with strengths $z_1, z_2, \ldots, z_{d_{\text{sae}}}$, the SAE reconstruction is:
\begin{equation}
  \vx_{\text{SAE}} = \sum_{i=1}^{d_{\text{sae}}} z_i \mW_{\text{dec}, L}^{(i)} + \vb_{\text{dec}}
\end{equation}

\subsection{Attention Computation in Layer \texorpdfstring{$L+1$}{L+1}}

For attention head $h$ in layer $L+1$, under our approximation where normalization effects are set aside, the query and key projections are:
\begin{align}
  \vq^{(h)} &= \vx_{\text{SAE}} \mW_{Q, L+1}^{(h)} = \sum_{i=1}^{d_{\text{sae}}} z_i \mW_{\text{dec}, L}^{(i)} \mW_{Q, L+1}^{(h)} + \vb_{\text{dec}} \mW_{Q, L+1}^{(h)} \\
  \vk^{(h)} &= \vx_{\text{SAE}} \mW_{K, L+1}^{(h)} = \sum_{i=1}^{d_{\text{sae}}} z_i \mW_{\text{dec}, L}^{(i)} \mW_{K, L+1}^{(h)} + \vb_{\text{dec}} \mW_{K, L+1}^{(h)}
\end{align}

Defining the per-feature query and key vectors:
\begin{equation}
  \vq_i^{(h)} = \mW_{\text{dec}, L}^{(i)}\mW_{Q, L+1}^{(h)}, \quad \vk_i^{(h)} = \mW_{\text{dec}, L}^{(i)}\mW_{K, L+1}^{(h)}
\end{equation}
and the bias contributions:
\begin{equation}
  \vq_b^{(h)} = \vb_{\text{dec}} \mW_{Q, L+1}^{(h)}, \quad \vk_b^{(h)} = \vb_{\text{dec}} \mW_{K, L+1}^{(h)}
\end{equation}
we can rewrite the projections as:
\begin{equation}
  \vq^{(h)} = \sum_{i=1}^{d_{\text{sae}}} z_i \vq_i^{(h)} + \vq_b^{(h)}, \quad \vk^{(h)} = \sum_{j=1}^{d_{\text{sae}}} z_j \vk_j^{(h)} + \vk_b^{(h)}
\end{equation}

\subsection{Decomposition of Attention Scores}

The pre-softmax attention score in head $h$ is the scaled dot product:
\begin{align}
  s^{(h)} &= \frac{\vq^{(h)} \cdot \vk^{(h)}}{\sqrt{d_{\text{head}}}} \\
  &= \frac{1}{\sqrt{d_{\text{head}}}} \Biggl( \sum_{i=1}^{d_{\text{sae}}} \sum_{j=1}^{d_{\text{sae}}} z_i z_j (\vq_i^{(h)} \cdot \vk_j^{(h)}) + \sum_{i=1}^{d_{\text{sae}}} z_i (\vq_i^{(h)} \cdot \vk_b^{(h)}) \notag \\
  &\qquad\qquad {}+ \sum_{j=1}^{d_{\text{sae}}} z_j (\vq_b^{(h)} \cdot \vk_j^{(h)}) + \vq_b^{(h)} \cdot \vk_b^{(h)} \Biggr)
\end{align}

The most interesting term is the double sum over feature pairs.
Defining the per-feature pairwise attention potential:
\begin{equation}
  s^{(h)}_{i,j} = \frac{\vq_i^{(h)} \cdot \vk_j^{(h)}}{\sqrt{d_{\text{head}}}}
\end{equation}
the forward pass attention score decomposes as:
\begin{equation}
  s^{(h)} = \sum_{i=1}^{d_{\text{sae}}} \sum_{j=1}^{d_{\text{sae}}} z_i z_j \, s^{(h)}_{i,j} + \text{(bias terms)}
\end{equation}

This decomposition shows that the pre-softmax matrix $\mS^{(h)} = [s^{(h)}_{i,j}]$ captures the weight-based structural potentials that, when weighted by actual feature activations $z_i z_j$, determine the attention scores in the forward pass (up to the normalization approximation discussed above).

\section{Framework clarifications}
\label{sec:framework-clarification}

This appendix expands on the three modeling choices flagged at the end of \Cref{sec:framework}: per-feature normalization, the encoder--decoder asymmetry that motivates our paired projections, and the decoder-side scope of the do-intervention.

\subsection{Per-feature normalization preserves rank ordering}
\label{sec:framework-clarification-norm}

The projections in \cref{eq:dec-unemb,eq:dec-q,eq:dec-k} apply $\operatorname{FinalNorm}$ or $\operatorname{AttnNorm}$ to a single decoder row in isolation, whereas in the model's forward pass these layer normalizations act on the full residual-stream vector at a token position. Because the in-context norm depends on every other component present in the stream, per-feature normalization is necessarily an approximation. Quantifying this approximation against the in-context norm directly would require running inference across a corpus and tracking feature co-activation, which moves outside the weight-based premise of the framework. We instead address the concern that motivates the question: whether the approximation distorts the \emph{ordering} of features, since rank ordering is what every claim in the paper relies on.

\begin{figure}[t]
  \centering
  \includegraphics[width=0.85\linewidth]{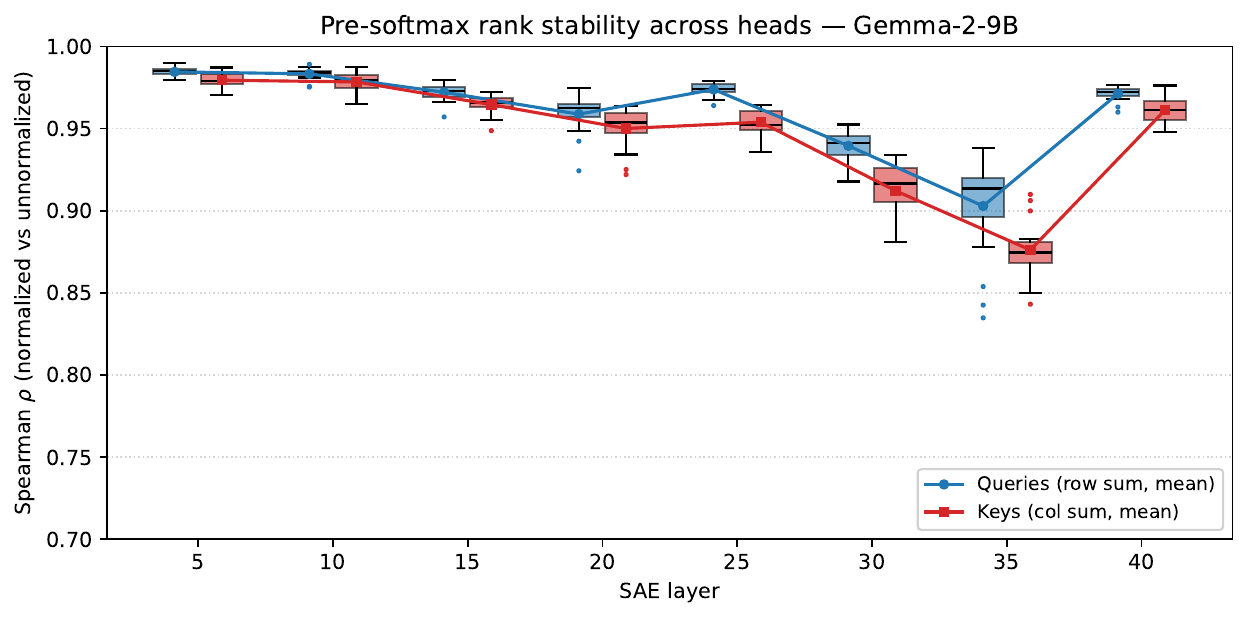}
  \caption{Per-layer Spearman rank correlation between per-feature-normalized and unnormalized pre-softmax aggregate scores, across all attention heads in 8 sampled layers of Gemma-2-9B. Boxes show the per-head distribution; lines join the per-layer means. Both query (row-sum) and key (column-sum) aggregates remain above $\bar\rho \approx 0.88$ at every depth, confirming that the per-feature normalization preserves the rank ordering on which our claims depend.}
  \label{fig:norm-rank-stability}
\end{figure}

We compare the per-feature normalized attention scores against an unnormalized baseline across 128 heads spanning 8 evenly sampled layers (layers 5--40) of Gemma-2-9B. The mean Spearman rank correlation between the two is $0.902 \le \bar\rho \le 0.985$ for queries and $0.876 \le \bar\rho \le 0.980$ for keys (\Cref{fig:norm-rank-stability}); the per-head distributions sit tightly above $0.85$ at every layer for both query (row-sum) and key (column-sum) aggregates. All architectural findings reported in \Cref{sec:exp2,sec:exp3} are qualitatively identical under both normalization schemes. The per-feature approximation therefore preserves the essential rank structure that our analyses exploit.

\subsection{Encoder--decoder asymmetry justifies the paired projections}
\label{sec:framework-clarification-encdec}

A natural concern about \cref{eq:enc-emb,eq:dec-unemb} is that, in a model with tied embeddings, $\mW_{\text{enc}}^{(\cdot,i)} \mW_{\text{E}}^{\top}$ and $\mW_{\text{dec}}^{(i)} \mW_{\text{U}}$ would carry the same information whenever the encoder column and decoder row of feature $i$ are themselves aligned, making the paired analysis redundant. Whether this redundancy holds is therefore an empirical question about the SAE itself, not about the host model's weight tying.

\begin{figure}[t]
  \centering
  \includegraphics[width=\linewidth]{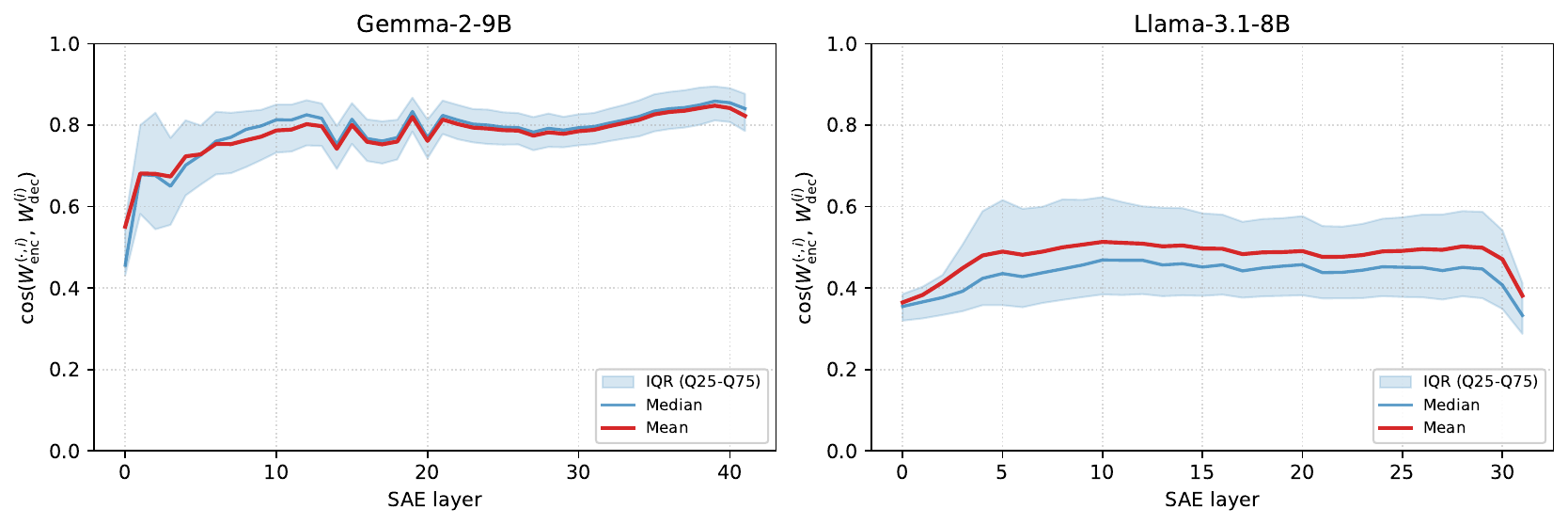}
  \caption{Per-layer cosine similarity between the encoder column and decoder row of the same SAE feature, for the Gemma Scope SAEs on Gemma-2-9B (left) and the Llama Scope SAEs on Llama-3.1-8B (right). Solid lines are the per-layer mean (red) and median (blue) across features; the shaded band is the inter-quartile range. Gemma sits around $0.8$ on average with a dip in early layers; Llama sits substantially lower throughout, making the paired encoder--embedding and decoder--unembedding analyses non-redundant by an even wider margin.}
  \label{fig:enc-dec-cosine}
\end{figure}

We measure the per-feature cosine similarity between $\mW_{\text{enc}}^{(\cdot,i)}$ and $\mW_{\text{dec}}^{(i)}$ across layers of the Gemma Scope SAEs on Gemma-2-9B and the Llama Scope SAEs on Llama-3.1-8B (\Cref{fig:enc-dec-cosine}). On Gemma-2-9B the layer-wise mean is approximately $0.8$, with early layers dropping to $\approx 0.55$. Llama-3.1-8B sits substantially lower across the entire depth, with mean per-feature cosine well below the Gemma values, so the encoder and decoder vectors of the same Llama Scope feature are even further from being interchangeable. This bimodality between the encoder and decoder of a feature has been documented independently \cite{brzozowski2025BimodalitySparseAutoencoder}. The two sides therefore carry different signal in both architectures, and \Cref{sec:exp1} confirms that each projection surfaces a population the other does not.

\subsection{The do-intervention is decoder-side by construction}
\label{sec:framework-clarification-do}

The intervention $\mathrm{do}(z_i = 0)$ in \cref{eq:do-intervention} zeros the SAE feature and re-decodes, equivalently subtracting $z_i \mW_{\text{dec},L}^{(i)}$ from the residual stream. What it removes is the decoder write, so it is a counterfactual test of claims that route through the decoder row: the decoder--unembedding projection (\cref{eq:dec-unemb}) and the decoder--QK projections (\cref{eq:dec-q,eq:dec-k}). \Cref{sec:exp1,sec:exp2} use it for exactly these claims.

The encoder--embedding projection in \cref{eq:enc-emb} is structurally different: it predicts \emph{which inputs activate feature $i$}, not what feature $i$ does once active. A read-side prediction of this form is supported by the structural projection together with the activation evidence on which it relies, but it is not the kind of claim that $\mathrm{do}(z_i = 0)$ can validate, since zeroing the activation removes the write rather than testing the read. We therefore do not attempt to force the encoder side through the same do-operation, and we treat the encoder--embedding result in \Cref{sec:exp1} as a structural prediction with the LLM-judge agreement check serving as its independent confirmation.

\section{Rationale for Attention Specialization Metrics}
\label{sec:attn-rationale}

The decision to analyze computational features from both a query-based and a key-based perspective is motivated by the inherent asymmetry of the attention mechanism.
The row-wise application of the softmax function imparts fundamentally different properties and, consequently, different computational roles to features acting as queries versus those acting as keys.

\subsection{The Asymmetry of Query and Key}

The attention score from a query feature $i$ to a key feature $j$ is calculated as part of a distribution over all possible keys: $\va_i^{(h)} = \text{softmax}(\vs_i^{(h)})$.
This means that for a given query $i$, all keys $j \in \{1, \ldots, d_{\text{sae}}\}$ are in direct competition to receive its attention.

\paragraph{Query-based (Outgoing) Perspective} A query feature must allocate a fixed budget of attention (summing to 1.0) across all possible keys.
To achieve a high score for a specific key, it only needs to have a pre-softmax alignment with that key that is significantly higher than its alignment with any other key.
This identifies features that are specialized ``lookers'' or ``actors,'' designed to perform a targeted information lookup.

\paragraph{Key-based (Incoming) Perspective} A key feature's incoming attention is the sum of attention it receives from many independent, row-wise softmax operations.
There is no column-wise competition.
For a key to be considered ``dominant,'' it must consistently win the attention competition across many different queries.
This identifies features that represent ``information hubs'':
concepts that are so useful that multiple computational circuits attend to them.

\subsection{Justification of Metric Choices}

This asymmetry necessitates different metrics for each perspective:

\textbf{For query specialists:} We use \textit{sum} of top-k \textgreater~0.95.

This metric is ideal for identifying ``specialist lookers.''
It asks: ``Does this feature concentrate nearly all of its attention budget on a very small set of targets?''
This is a very intuitive and powerful definition of a computationally specialized feature.
It directly identifies features that participate in targeted circuits, capturing the essence of ``doing something specific'' in attention.

\textbf{For key hubs:} We use \textit{mean} of top-k \textgreater~0.25.

This is a much stricter criterion designed to identify ``information hubs.'' A simple sum could be inflated by one query giving 0.95 attention and nine others giving almost none.
The mean, however, requires that a feature be consistently and substantially important to multiple independent queries (e.g., each of its top-10 admirers dedicates, on average, \textgreater25\% of their attention to it).
This filters for features that are important to the model's computations, for which a lower pass rate is expected and informative.

We chose $k = 10$ primarily as a compromise between retaining sufficient information and managing the computational budget required for all features across all layers and models, since each column or row stores $d_{\text{sae}}$ scores for both pre- and post-softmax matrices during exploration.

Given that these thresholds may be somewhat arbitrary, we also provide several alternative values for $\tau_Q$ and $\tau_K$ for reference in \cref{tab:query-pass-rates} and \cref{tab:key-pass-rates}, respectively.

\begin{table}[h]
  \centering
  \caption{Average percentage of features qualifying as query specialists across all layers for a range of thresholds ($\tau_Q$). The criterion requires the sum of a feature's top-10 post-softmax attention scores to exceed the threshold. The highlighted column corresponds to our primary choice.}
  \label{tab:query-pass-rates}
    \begin{tabular}{l>{\columncolor{Aquamarine!25}}ccccc}
    \toprule
    Threshold $\tau_Q$ & 0.95 & 0.90 & 0.75 & 0.50 & 0.25 \\
    \midrule
    Gemma-2-2B & 39.54\% & 43.79\% & 50.94\% & 59.29\% & 68.98\% \\
    Gemma-2-9B & 53.82\% & 58.49\% & 65.60\% & 72.84\% & 79.73\% \\
    Llama-3.1-8B & 9.75\% & 11.71\% & 15.62\% & 21.14\% & 28.83\% \\
    \bottomrule
    \end{tabular}
\end{table}

\begin{table}[ht]
  \centering
  \caption{Average percentage of features qualifying as key hubs across all layers for a range of thresholds ($\tau_K$). The criterion requires the mean of a feature's top-10 incoming post-softmax attention scores to exceed the threshold. The highlighted column corresponds to our primary choice.}
  \label{tab:key-pass-rates}
    \begin{tabular}{lc>{\columncolor{Aquamarine!25}}cccc}
    \toprule
    Threshold $\tau_K$ & 0.50 & 0.25 & 0.10 & 0.05 & 0.01 \\
    \midrule
    Gemma-2-2B & 1.49\% & 2.90\% & 8.88\% & 14.16\% & 25.82\% \\
    Gemma-2-9B & 3.16\% & 6.15\% & 17.52\% & 27.31\% & 43.59\% \\
    Llama-3.1-8B & 0.42\% & 0.85\% & 2.15\% & 3.57\% & 8.03\% \\
    \bottomrule
    \end{tabular}
\end{table}

\section{Validating Computational Specialization: Intervention}
\label{sec:exp2-validation}

This appendix tests, with the same $\mathrm{do}(z_i = 0)$ primitive used in \Cref{sec:exp1-validation-intervention}, whether features flagged as query specialists or key hubs in \Cref{sec:exp2} actually shape the QK computation of the next attention layer. The question is mirrored to each role: for a query specialist at layer $L$ firing at position $p$, removing its decoder write should perturb the post-softmax attention row at $(L\!+\!1, h^\star, p)$; for a key hub at $p$, removal should reduce the attention that later queries ($q > p$) place on $p$ at $(L\!+\!1, h^\star)$. The structurally predicted head $h^\star$ is the head that exceeds the role's threshold; ties broken by the role-specific score.

The two roles have asymmetric prompt requirements that follow directly from where they sit in the attention computation. A query specialist participates in attention as soon as it fires: its decoder write enters the residual stream at position $p$, and at the very next layer that residual is read as a query at the same position $p$, so the attention behaviour we want to inspect already exists at the firing token. We therefore truncate the prompt at $p$ and feed the model the prefix that ends exactly where the feature peaks, which is also the convention used for the previous intervention experiment (\Cref{sec:exp1-validation-intervention}). A key hub, by contrast, only matters for queries that come \emph{after} the firing position: its decoder write puts a key vector at position $p$ that subsequent positions can attend back to. Truncating at $p$ would leave no later queries to read the hub. We therefore keep the prompt un-truncated for key hubs so the firing position $p$ sits mid-sequence and the queries at $q > p$ are present to react to the intervention.

\subsection{Setup}
\label{sec:exp2-validation-setup}

\paragraph{Sampling.} For each (model, role, layer) cell we draw 100 flagged features and pair each with a \emph{matched-norm baseline}: a strict-non-specialist feature whose decoder norm is within $\pm 5\%$ of the flagged feature's and that fires on the same prompt at the same position. \emph{Strict-non-specialist} means failing both criteria $\sum_\text{top-10}\mA^{(h)}_{i,\cdot} > \tau_Q$ and $\frac{1}{10}\!\sum_\text{top-10}\mA^{(h)}_{\cdot,j} > \tau_K$ at \emph{every} head.

\paragraph{Prompt construction.} For each flagged feature we fetch its top max-activating prompt from Neuronpedia~\cite{neuronpedia}. For query specialists the prompt is truncated at the firing position so the last token fed to the model is exactly the firing token. For key hubs the prompt is kept un-truncated so the firing position $p$ sits mid-sequence and later queries can attend back to it. The flagged head $h^\star$ is the argmax over heads of the role-specific specialist score on the flagged feature.

\paragraph{Metric.} For query specialists we report the KL divergence between the clean and intervened post-softmax attention rows at $(L\!+\!1, h^\star, p)$ over keys $j \!\in\! [0, p]$. For key hubs we report the change in mean incoming attention from later queries, $\Delta = \frac{1}{|Q|}\sum_{q \in Q}\!\big(\!\mA^{(h^\star)}_{q,p,\text{clean}} - \mA^{(h^\star)}_{q,p,\text{intervened}}\big)$ with $Q = \{q : q > p\}$. KL is non-negative; $\Delta$ is signed, and a positive value matches the structural prediction that the hub was a target the queries concentrated on.

We sample 8 SAE layers per model, evenly spaced across depth (Gemma-2-9B: $\{5, 10, \ldots, 40\}$; Llama-3.1-8B: $\{16, 18, \ldots, 30\}$, plus $\{10, 14\}$ for query specialists where the flagged pool remains populated). Llama is constrained to $L \!\geq\! 14$ for key hubs because at $L \!<\! 13$ the hub pool is too thin to support $n\!=\!100$ with matched-norm baselines.

\subsection{Query specialists: causal effect on the attention row}
\label{sec:exp2-validation-query}

\Cref{tab:exp2-intervention-query} reports the 18 sweeps (8 layers in Gemma plus 10 in Llama). Removing a flagged query specialist shifts its attention row by a median KL of $0.010$--$0.246$ across all sweeps, against $4\!\times\!10^{-5}$--$1.7\!\times\!10^{-3}$ for the matched-norm baseline firing on the same prompt. The flagged-to-matched ratio (per-pair median, not the ratio of the two median columns) is $43\times$ to $2161\times$ across cells, with a typical value near $100\times$. The win-rate, the fraction of pairs in which the flagged feature produces a strictly larger KL than its matched baseline, is $0.93$--$1.00$ in every cell, and reaches $1.00$ in 6 of the 18.

The KL effect is robust across depth in both models. Gemma's median KL is largest at the shallowest sampled layer and stays in $[0.010, 0.056]$ for $L \!\geq\! 10$. Llama's median KL is consistently in $[0.012, 0.060]$ across $L \!\in\! [14, 30]$ with a single early peak at $L\!=\!10$ ($0.181$).

\begin{table}[ht]
  \centering
  \caption{Intervention validation for Experiment 2, query specialists. For each sampled SAE layer we draw 100 flagged query specialists and pair each with a matched-norm baseline: a strict-non-specialist feature whose decoder norm is within $\pm 5\%$ of the flagged feature's and that fires on the same prompt at the same position. We apply $\mathrm{do}(z\!=\!0)$ to each and read the KL divergence between clean and intervened post-softmax attention rows of the flagged head at the firing position. $n_{m}$ is the number of pairs for which a matched baseline was available.}
  \label{tab:exp2-intervention-query}
  \begin{tabular}{l|cc|cc|c}
  \toprule
  \multirow{2}{*}{Model} & \multirow{2}{*}{Layer} & \multirow{2}{*}{$n_{m}$} & \multicolumn{2}{c|}{Median KL} & Win-rate \\
   &  &  & flagged & matched & flagged $>$ matched \\
  \midrule
  Gemma-2-9B & 5 & 96 & 0.2456 & $8.21\!\!\times\!\!10^{-5}$ & 1.00 \\
   & 10 & 99 & 0.05559 & $5.67\!\!\times\!\!10^{-5}$ & 0.98 \\
   & 15 & 100 & 0.05594 & $8.50\!\!\times\!\!10^{-5}$ & 1.00 \\
   & 20 & 95 & 0.04033 & $2.94\!\!\times\!\!10^{-4}$ & 0.99 \\
   & 25 & 96 & 0.03372 & $1.35\!\!\times\!\!10^{-4}$ & 1.00 \\
   & 30 & 98 & 0.02398 & $6.36\!\!\times\!\!10^{-5}$ & 1.00 \\
   & 35 & 99 & 0.01032 & $4.30\!\!\times\!\!10^{-5}$ & 1.00 \\
   & 40 & 99 & 0.02766 & $4.85\!\!\times\!\!10^{-5}$ & 0.98 \\
  \midrule
  Llama-3.1-8B & 10 & 95 & 0.1811 & $1.68\!\!\times\!\!10^{-3}$ & 0.99 \\
   & 14 & 90 & 0.01206 & $1.72\!\!\times\!\!10^{-4}$ & 1.00 \\
   & 16 & 96 & 0.04495 & $5.81\!\!\times\!\!10^{-4}$ & 0.97 \\
   & 18 & 98 & 0.04454 & $5.44\!\!\times\!\!10^{-4}$ & 0.97 \\
   & 20 & 99 & 0.05926 & $6.59\!\!\times\!\!10^{-4}$ & 0.97 \\
   & 22 & 95 & 0.03351 & $4.84\!\!\times\!\!10^{-4}$ & 0.97 \\
   & 24 & 95 & 0.01196 & $2.53\!\!\times\!\!10^{-4}$ & 0.97 \\
   & 26 & 96 & 0.01617 & $5.20\!\!\times\!\!10^{-4}$ & 0.99 \\
   & 28 & 89 & 0.02982 & $5.23\!\!\times\!\!10^{-4}$ & 0.93 \\
   & 30 & 90 & 0.01197 & $2.27\!\!\times\!\!10^{-4}$ & 0.98 \\
  \bottomrule
  \end{tabular}
\end{table}

\subsection{Key hubs: causal effect on incoming attention}
\label{sec:exp2-validation-key}

\Cref{tab:exp2-intervention-key} reports the results for key hubs. The headline magnitude is $|\Delta|$, the absolute change in mean incoming attention from later queries. Median $|\Delta|$ for flagged hubs is $\sim\!10^{-3}$ across cells (Gemma: $8.4\!\times\!10^{-4}$--$2.5\!\times\!10^{-3}$; Llama: $1.1\!\times\!10^{-4}$--$2.9\!\times\!10^{-3}$), against $\sim\!10^{-5}$ for most matched baselines. These magnitudes are intrinsically small in absolute terms because the metric averages over many later queries (median $|Q|\!\in\![50, 80]$). The cross-row scale is therefore not directly comparable to the query KL; the relevant comparison is the ratio against the matched-norm baseline. That ratio (again, per-pair median) is $9\times$ to $37\times$ across all 17 sweeps, and the win-rate against matched is $0.88$--$0.98$ in every cell.

\begin{table}[ht]
  \centering
  \caption{Intervention validation for Experiment 2, key hubs. For each sampled SAE layer we draw 100 flagged key hubs and pair each with a matched-norm baseline. We apply $\mathrm{do}(z\!=\!0)$ at the hub's firing position and measure the change in mean incoming attention from later queries ($q > p$) at the flagged head. Magnitude columns report $|\Delta|$ of the drop.}
  \label{tab:exp2-intervention-key}
  \begin{tabular}{l|cc|cc|c}
  \toprule
  \multirow{2}{*}{Model} & \multirow{2}{*}{Layer} & \multirow{2}{*}{$n_{m}$} & \multicolumn{2}{c|}{Median $|$drop$|$} & Win-rate \\
   &  &  & flagged & matched & flagged $>$ matched \\
  \midrule
  Gemma-2-9B & 5 & 98 & $1.705\!\!\times\!\!10^{-3}$ & $1.02\!\!\times\!\!10^{-4}$ & 0.94 \\
   & 10 & 93 & $2.504\!\!\times\!\!10^{-3}$ & $6.22\!\!\times\!\!10^{-5}$ & 0.98 \\
   & 15 & 99 & $1.058\!\!\times\!\!10^{-3}$ & $4.21\!\!\times\!\!10^{-5}$ & 0.97 \\
   & 20 & 93 & $2.026\!\!\times\!\!10^{-3}$ & $2.33\!\!\times\!\!10^{-4}$ & 0.94 \\
   & 25 & 92 & $1.570\!\!\times\!\!10^{-3}$ & $1.29\!\!\times\!\!10^{-4}$ & 0.92 \\
   & 30 & 93 & $1.048\!\!\times\!\!10^{-3}$ & $4.61\!\!\times\!\!10^{-5}$ & 0.90 \\
   & 35 & 89 & $8.436\!\!\times\!\!10^{-4}$ & $2.77\!\!\times\!\!10^{-5}$ & 0.98 \\
   & 40 & 98 & $1.097\!\!\times\!\!10^{-3}$ & $4.08\!\!\times\!\!10^{-5}$ & 0.98 \\
  \midrule
  Llama-3.1-8B & 14 & 92 & $1.101\!\!\times\!\!10^{-4}$ & $8.61\!\!\times\!\!10^{-6}$ & 0.92 \\
   & 16 & 98 & $2.447\!\!\times\!\!10^{-3}$ & $1.76\!\!\times\!\!10^{-4}$ & 0.92 \\
   & 18 & 96 & $7.716\!\!\times\!\!10^{-4}$ & $6.07\!\!\times\!\!10^{-5}$ & 0.95 \\
   & 20 & 94 & $2.654\!\!\times\!\!10^{-3}$ & $1.95\!\!\times\!\!10^{-4}$ & 0.88 \\
   & 22 & 93 & $2.352\!\!\times\!\!10^{-3}$ & $1.08\!\!\times\!\!10^{-4}$ & 0.97 \\
   & 24 & 96 & $1.437\!\!\times\!\!10^{-3}$ & $1.44\!\!\times\!\!10^{-4}$ & 0.95 \\
   & 26 & 96 & $2.852\!\!\times\!\!10^{-3}$ & $9.16\!\!\times\!\!10^{-5}$ & 0.92 \\
   & 28 & 89 & $1.118\!\!\times\!\!10^{-3}$ & $6.96\!\!\times\!\!10^{-5}$ & 0.91 \\
   & 30 & 93 & $1.146\!\!\times\!\!10^{-3}$ & $7.81\!\!\times\!\!10^{-5}$ & 0.94 \\
  \bottomrule
  \end{tabular}
\end{table}

Taken together, query and key interventions deliver a causal counterpart to the projection-based criteria of \Cref{sec:exp2}. A feature flagged by the post-softmax thresholds is, with very high consistency, the feature that actually steers the next layer's attention at the position where it fires; a non-specialist of equal decoder norm firing on the same prompt produces an effect one to three orders of magnitude smaller.

\section{Analysis of Lower Computational Feature Pass Rates in Llama-3.1-8B}

A notable finding from our computational analysis is the lower pass rate for both query-specialists and key-hubs in Llama-3.1-8B compared to the Gemma-2 models (e.g., 9.75\% vs. 53.82\% for query-specialists with $\tau_Q=0.95$). We hypothesize this discrepancy arises from several interacting factors related to model and SAE architecture, as well as the nature of our out-of-context methodology.

\subsection{SAE dictionary size}
The most direct architectural difference is the SAE dictionary size: 32,768 for Llama-3.1-8B versus 16,384 for Gemma-2-9B. This has two potential consequences:
A larger dictionary size naturally leads to a lower percentage pass rate, even if the absolute number of specialized features for a given function were similar across models.

\subsection{Feature Splitting}
More importantly, a larger dictionary may encourage the SAE to learn a more distributed or compositional code \citep{chanin2025AbsorptionStudyingFeature}.
Instead of representing a concept with a single feature, it might reconstruct the same activation using a linear combination of several more granular features.
Our methodology, which evaluates each feature's decoder vector in isolation, would register a diluted signal in such cases, as no single feature vector carries the full computational weight of the concept.
Consequently, fewer individual features would meet the dominance thresholds.

We can formalize how feature splitting would lead to lower measured scores for individual features.
For simplicity, we ignore normalizations and reconstruction coefficients.

Let $\vx \in \mathbb{R}^{d_{\text{model}}}$ be a target vector in the residual stream that represents a specific computational role.
An ideal SAE might learn a single feature, say feature $k$, to represent this vector. In this scenario, the feature's decoder vector is proportional to the target vector:
\begin{equation}
  \mW_{\text{dec}}^{(k)} = \vx
\end{equation}
Our methodology measures this feature's participation in a QK circuit by projecting it. For example, its query vector for head $h$ would be:
\begin{equation}
  \vq_k^{(h)} = \mW_{\text{dec}}^{(k)} \mW_Q^{(h)} = \vx \mW_Q^{(h)}
\end{equation}
The resulting pre-softmax attention scores are directly proportional to the full magnitude of $\vx$.

Now, consider an SAE with a larger dictionary. It might learn to represent the same vector $\vx$ using a linear combination of two or more features. Let's say features $m$ and $n$ combine to form $\vx$:
\begin{equation}
  \vx = \mW_{\text{dec}}^{(m)} + \mW_{\text{dec}}^{(n)}
\end{equation}
A simple solution for the SAE optimization is to distribute the representation, for instance:
\begin{equation}
  \mW_{\text{dec}}^{(m)} = \alpha\vx \quad \text{and} \quad \mW_{\text{dec}}^{(n)} = (1-\alpha)\vx
\end{equation}
for some fraction $\alpha \in (0,1)$.

Our methodology analyzes each feature in isolation. When we evaluate feature $m$, its query vector is:
\begin{equation}
  \vq_m^{(h)} = \mW_{\text{dec}}^{(m)} \mW_Q^{(h)} = (\alpha\vx) \mW_Q^{(h)} = \alpha (\vx \mW_Q^{(h)}) = \alpha \vq_k^{(h)}
\end{equation}
The query vector for the split feature $m$ is a scaled-down version of the query vector for the ideal single feature $k$. Consequently, its pre-softmax attention score with any key feature $j$ is also scaled down:
\begin{equation}
  s_{m,j}^{(h)} = \frac{\vq_m^{(h)} \cdot \vk_j^{(h)}}{\sqrt{d_{\text{head}}}} = \alpha \left( \frac{\vq_k^{(h)} \cdot \vk_j^{(h)}}{\sqrt{d_{\text{head}}}} \right) = \alpha s_{k,j}^{(h)}
\end{equation}
The measured score for the individual split feature is only a fraction of the score it would have if it represented the concept alone. This makes it significantly less likely for any single feature to pass a fixed, high dominance threshold like $\tau_q$ or $\tau_k$.

\subsection{Strictness of out-of-context competition}
Our out-of-context framework is inherently strict, as it simulates attention competition across the entire SAE dictionary without the benefit of contextual sparsity from a real forward pass. In an actual inference step, only a few hundred features might be active, and the softmax competition would occur among this small set. Our method pits one feature against all others in the dictionary. This effect is likely magnified for Llama-3.1's larger dictionary; the bar for a single feature to dominate a competition against 32,767 others is substantially higher than against 16,383, potentially contributing to its lower pass rate.

\subsection{Focus on intra-layer interactions}
Our analysis focuses exclusively on intra-layer attention, where features from layer $L$ interact with the QK circuits of the immediately subsequent layer, $L+1$. It is plausible that some features are specialized for longer-range, inter-layer interactions, for example, by being preserved in the residual stream to be primarily used by attention heads in layers $L+2$ or beyond. Such long-range specialists would not be captured by our current methodology. While this is a limitation common to our analysis of all models, architectural differences could lead to Llama relying more heavily on such mechanisms, which would not be reflected in our reported scores.

\section{Additional Results}
\label{sec:more-results}

This appendix presents additional results that were omitted from the main text for brevity.

\begin{table}[ht]
  \centering
  \caption{Thresholds for semantic feature classification, set at the 50th percentile of scores from a representative layer. The arrows indicate the desired direction for each metric.}
  \label{tab:exp1-thresholds}
  \resizebox{\linewidth}{!}{
  \begin{tabular}{l|lc|ccc}
  \toprule
  Analysis Type & Model & Layer & Levenshtein Sim. $\uparrow$ & Cosine Sim. $\uparrow$ & Top-100 Entropy $\downarrow$ \\
  \midrule
  \multirow{3}{*}{\begin{tabular}[c]{@{}l@{}}Decoder-\\Unembedding\end{tabular}}
    & Gemma-2-2B & 16 & 0.157385 & 0.292363 & 0.612010 \\
    & Gemma-2-9B & 27 & 0.194815 & 0.332252 & 0.550316 \\
    & Llama-3.1-8B & 20 & 0.154048 & 0.092486 & 2.911260 \\
  \midrule
  \multirow{3}{*}{\begin{tabular}[c]{@{}l@{}}Encoder-\\Embedding\end{tabular}}
    & Gemma-2-2B & 3 & 0.149182 & 0.282485 & 4.604819 \\
    & Gemma-2-9B & 6 & 0.153151 & 0.348118 & 4.604956 \\
    & Llama-3.1-8B & 3 & 0.143704 & 0.028160 & 4.597589 \\
  \bottomrule
  \end{tabular}
  }
\end{table}

\begin{figure}[ht]
  \centering
  \begin{subfigure}{.48\linewidth}
    \centering
    \includegraphics[width=\linewidth,trim={.0cm .0cm .0cm .0cm}]{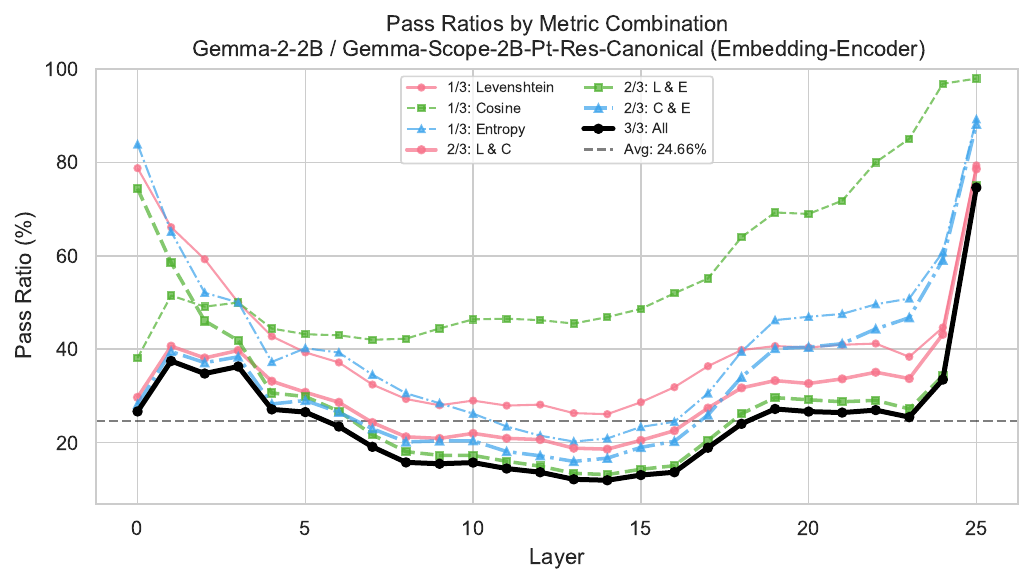}
  \end{subfigure}
  \begin{subfigure}{.48\linewidth}
    \centering
    \includegraphics[width=\linewidth,trim={.0cm .0cm .0cm .0cm}]{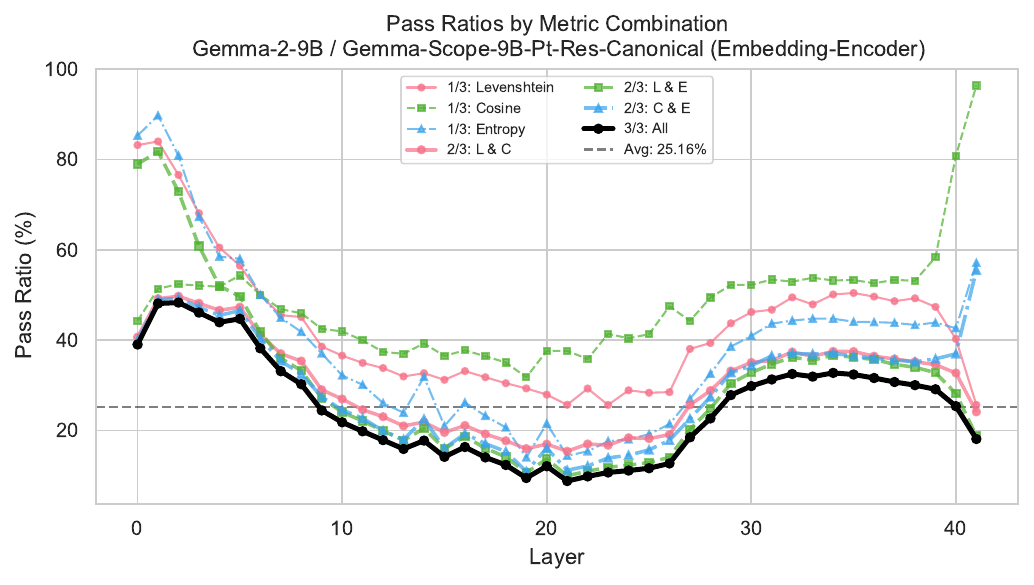}
  \end{subfigure}
  \caption{Supplementary results for Experiment 1 on Gemma-2-2B and 9B, using encoder--embedding alignment.
  Semantic features display a U-shaped distribution, with an average joint pass rate of 24.66\% and 25.16\%, respectively.}
  \label{fig:pass-rate-subsets-gemma-enc}
\end{figure}

\begin{figure}[ht]
  \centering
  \includegraphics[width=.5\linewidth,trim={.0cm .0cm .0cm .0cm}]{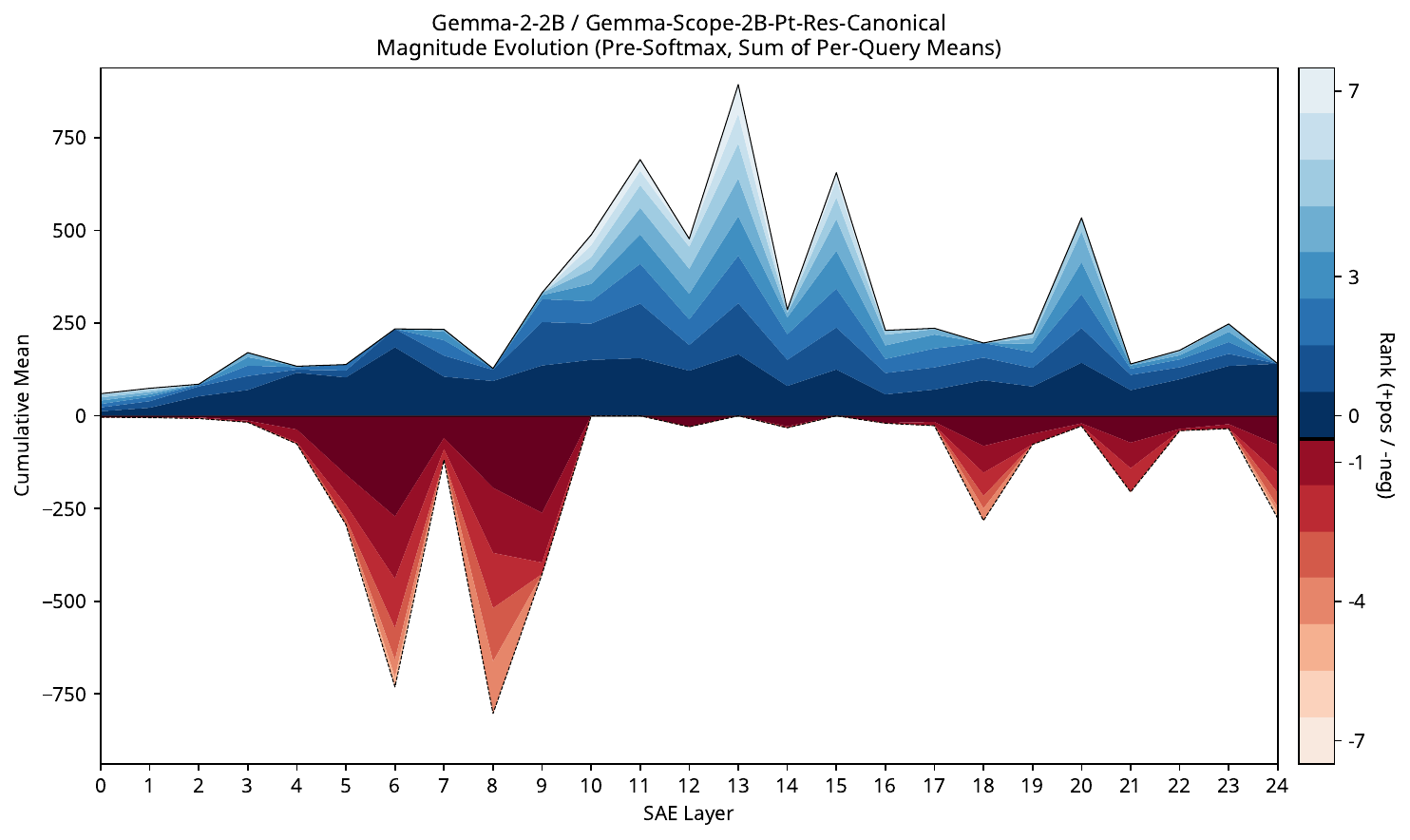}
  \caption{Main results for Experiment 2 on Gemma-2-2B: Magnitude of attention participation across layers,
with the total height at each layer representing the aggregate pre-softmax score. These scores are calculated as the mean across all keys for each query feature, then summed for each head. The stacking direction of the head values is determined based on whether they are positive or negative. Each color represents a rank of magnitude, rather than a specific head index, across all layers.}
  \label{fig:mag-evo-2b}
\end{figure}

\begin{figure}[ht]
  \centering
  \begin{subfigure}{.48\linewidth}
    \centering
    \includegraphics[width=\linewidth,trim={.0cm .0cm .0cm .0cm}]{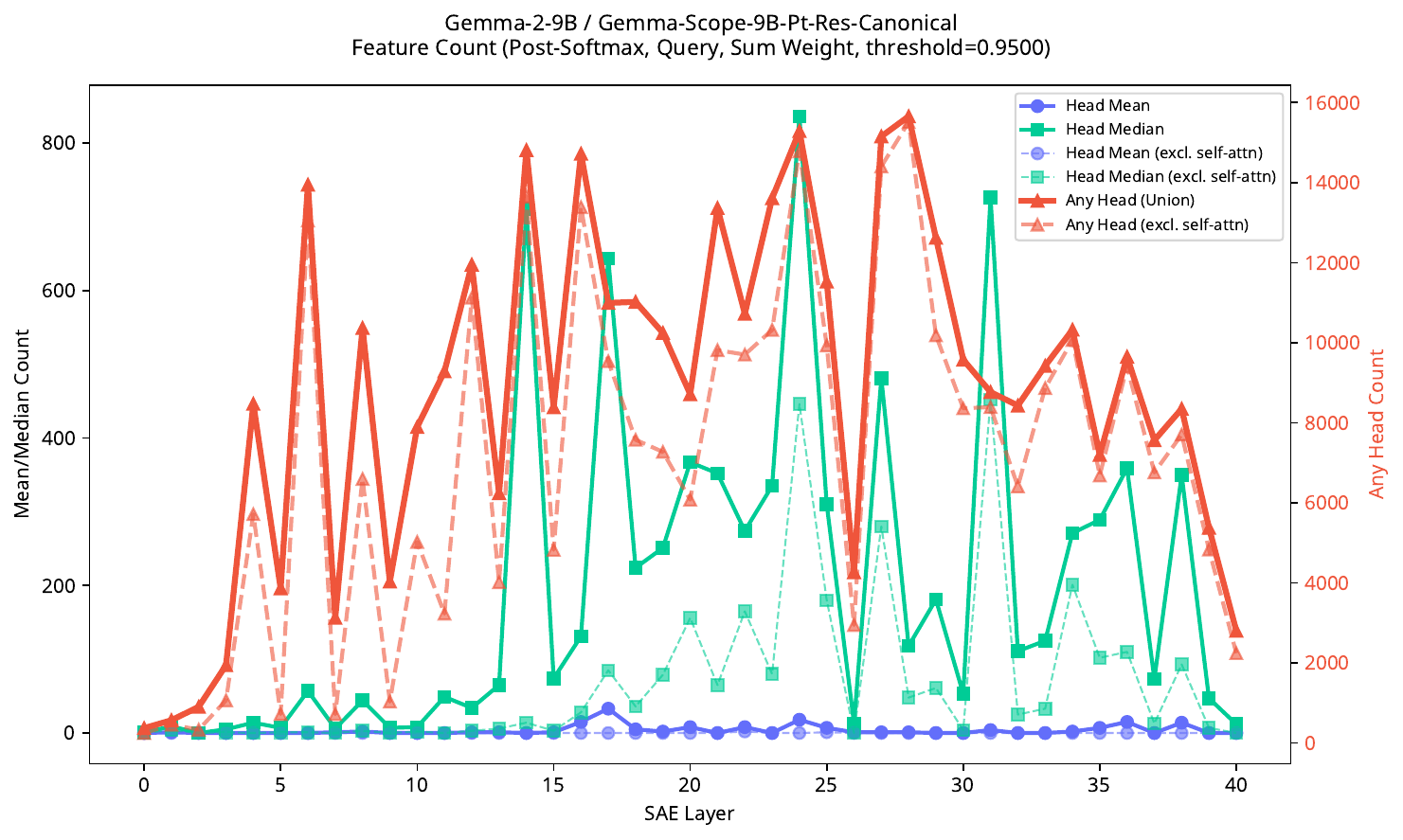}
  \end{subfigure}
  \begin{subfigure}{.48\linewidth}
    \centering
    \includegraphics[width=\linewidth,trim={.0cm .5cm .0cm .0cm}]{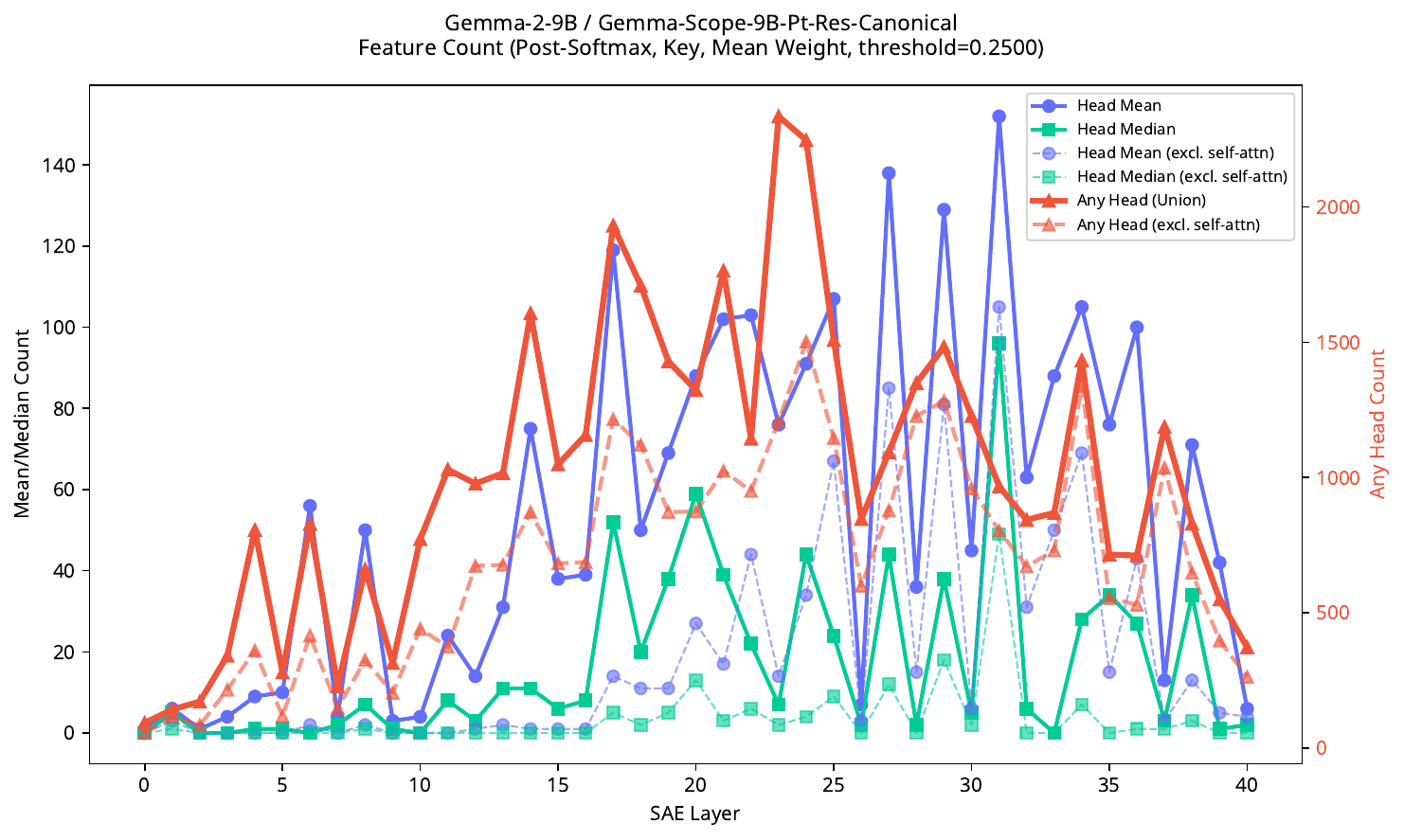}
  \end{subfigure}
  \caption{Feature count trends across Gemma-2-9B layers showing features with top-10 query-based post-softmax attention weight sum above 0.95 (left) and key-based attention weight mean above 0.25 (right).
  Note that there are two y-axes of different magnitudes: the left y-axis is for the mean and median counts across heads, while the right y-axis is for the union across all heads (``any head'').
  Solid lines include weights of self-attention; dashed lines exclude self-attention.}
  \label{fig:feature-count-9b}
\end{figure}

\begin{figure}[ht]
  \centering
  \begin{subfigure}{.48\linewidth}
    \centering
    \includegraphics[width=\linewidth,trim={.0cm .0cm .0cm .0cm}]{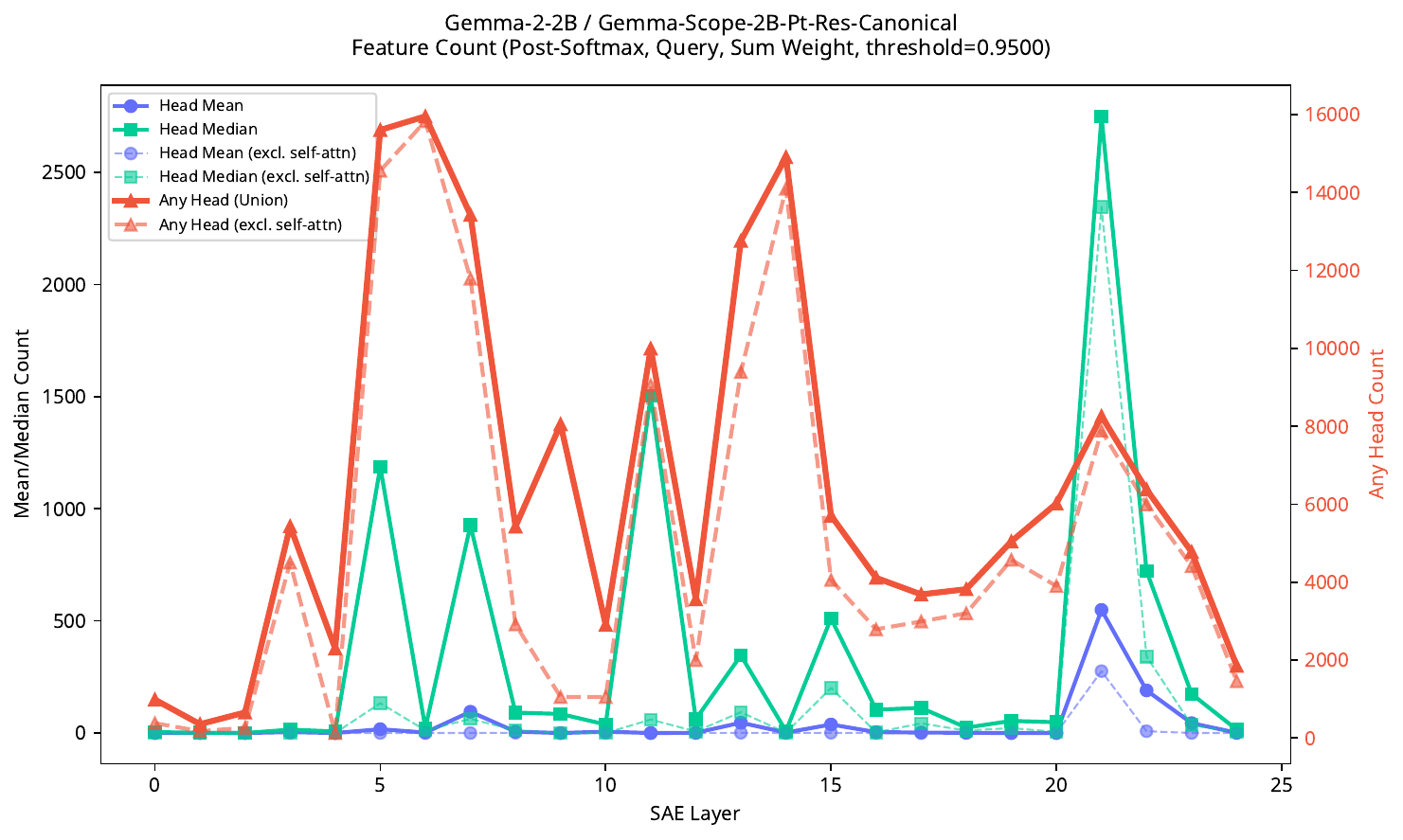}
  \end{subfigure}
  \begin{subfigure}{.48\linewidth}
    \centering
    \includegraphics[width=\linewidth,trim={.0cm .0cm .0cm .0cm}]{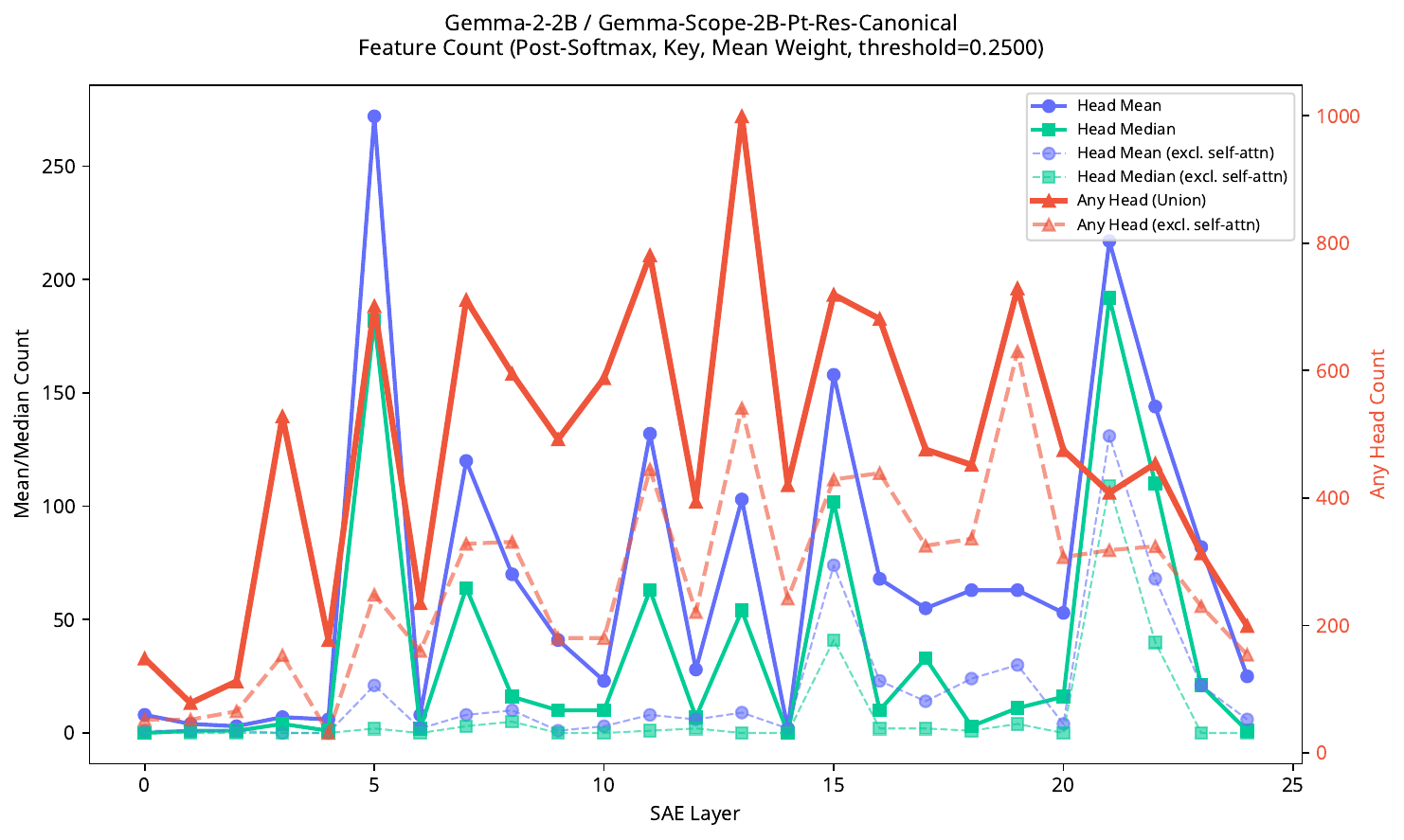}
  \end{subfigure}
  \caption{Feature count trends across Gemma-2-2B layers showing
features with top-10 query-based post-softmax attention weight
sum above 0.95 (left) and key-based attention weight mean above
0.25 (right). Note that there are two y-axes of different magnitudes:
the left y-axis is for the mean and median counts across
heads, while the right y-axis is for the union across all heads (``any
head''). Solid lines include weights of self-attention; dashed lines
exclude self-attention.}
  \label{fig:feature-count-2b}
\end{figure}

\begin{figure}[ht]
  \centering
  \begin{subfigure}{.5\linewidth}
    \centering
    \includegraphics[width=\linewidth,trim={.0cm .0cm .0cm .0cm}]{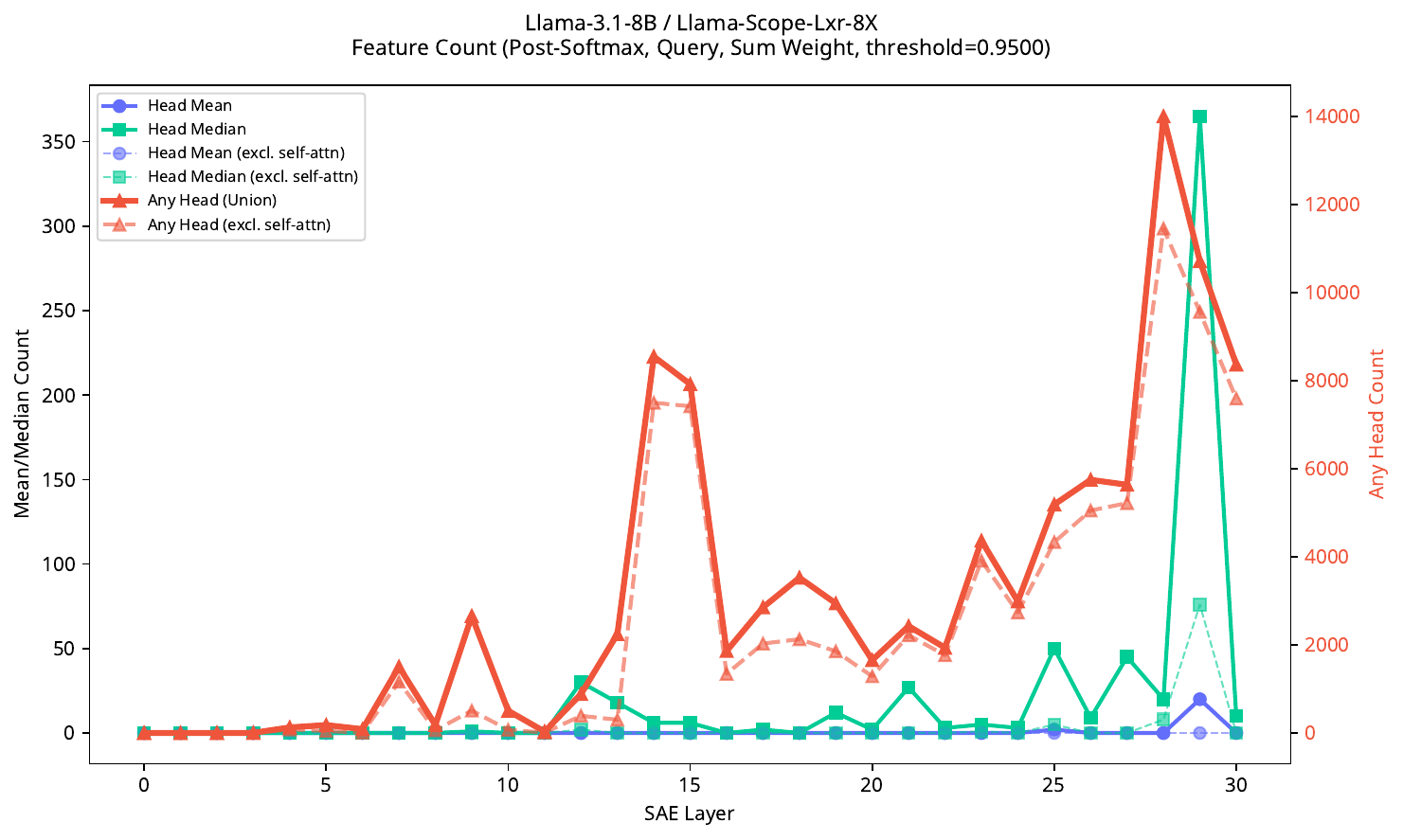}
  \end{subfigure}
  \begin{subfigure}{.48\linewidth}
    \centering
    \includegraphics[width=\linewidth,trim={.0cm .0cm .0cm .0cm}]{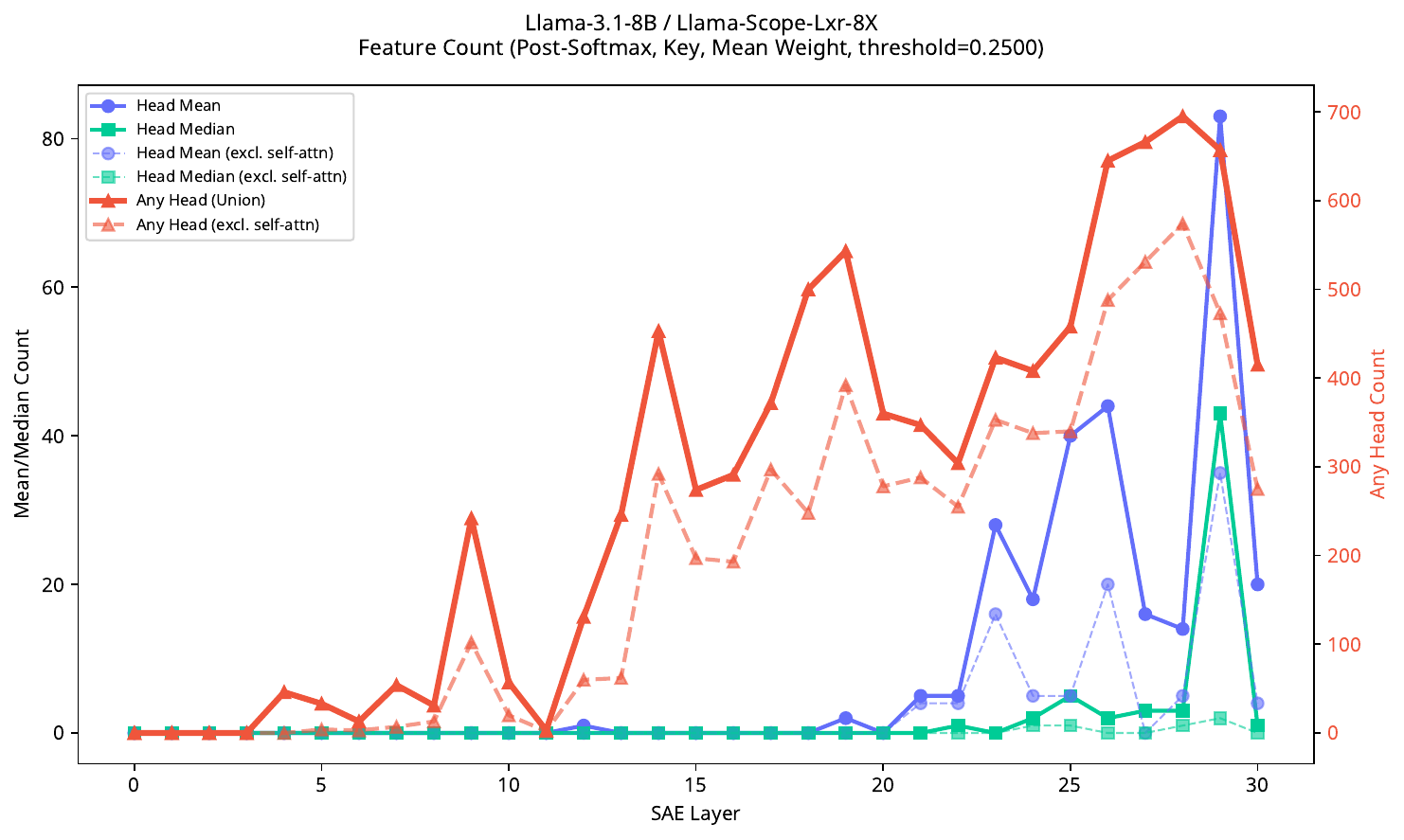}
  \end{subfigure}
  \caption{Feature count trends across Llama-3.1-8B layers showing
features with top-10 query-based post-softmax attention weight
sum above 0.95 (left) and key-based attention weight mean above
0.25 (right). Note that there are two y-axes of different magnitudes:
the left y-axis is for the mean and median counts across
heads, while the right y-axis is for the union across all heads (``any
head''). Solid lines include weights of self-attention; dashed lines
exclude self-attention.}
  \label{fig:feature-count-8b}
\end{figure}

\begin{figure}[ht]
  \centering
  \begin{subfigure}{.48\linewidth}
    \centering
    \includegraphics[width=\linewidth,trim={.0cm .0cm .0cm .0cm}]{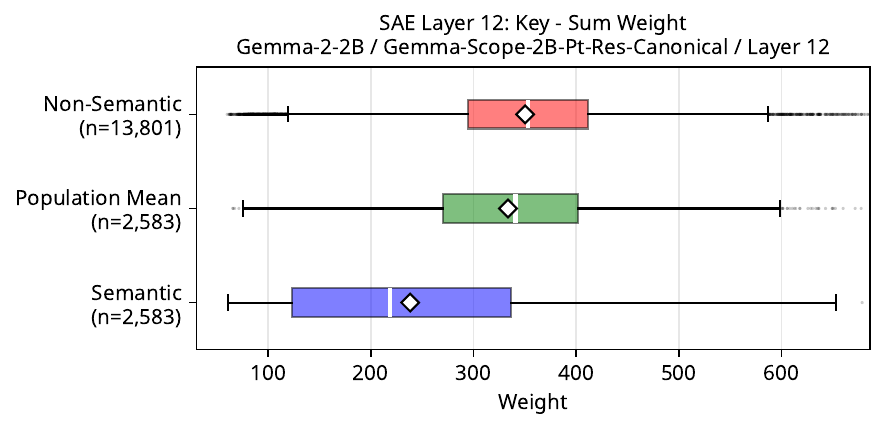}
    \caption{Key-based.}
    \label{fig:per-layer-boxplot-2b-k}
  \end{subfigure}
  \begin{subfigure}{.48\linewidth}
    \centering
    \includegraphics[width=\linewidth,trim={.0cm .0cm .0cm .0cm}]{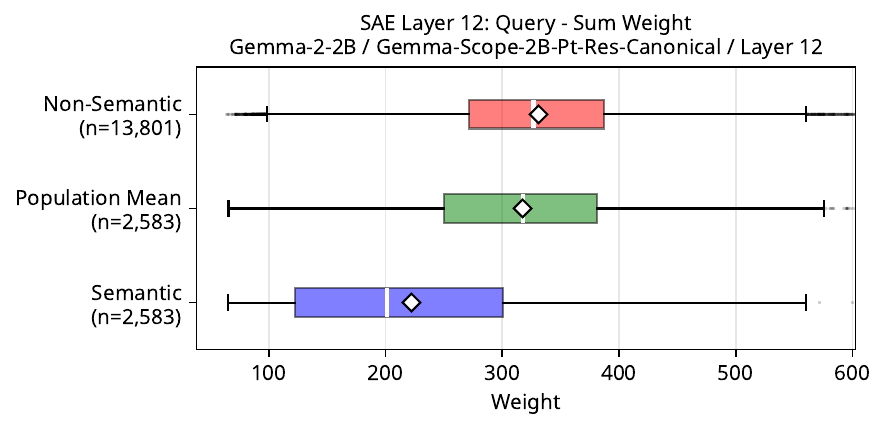}
    \caption{Query-based.}
    \label{fig:per-layer-boxplot-2b-q}
  \end{subfigure}
  \caption{Attention weight distributions in Gemma-2-2B layer 12. White diamonds mark means and white lines mark medians.}
  \label{fig:per-layer-boxplot-2b}
\end{figure}

\begin{figure}[ht]
  \centering
  \begin{subfigure}{.48\linewidth}
    \centering
    \includegraphics[width=\linewidth,trim={.0cm .0cm .0cm .0cm}]{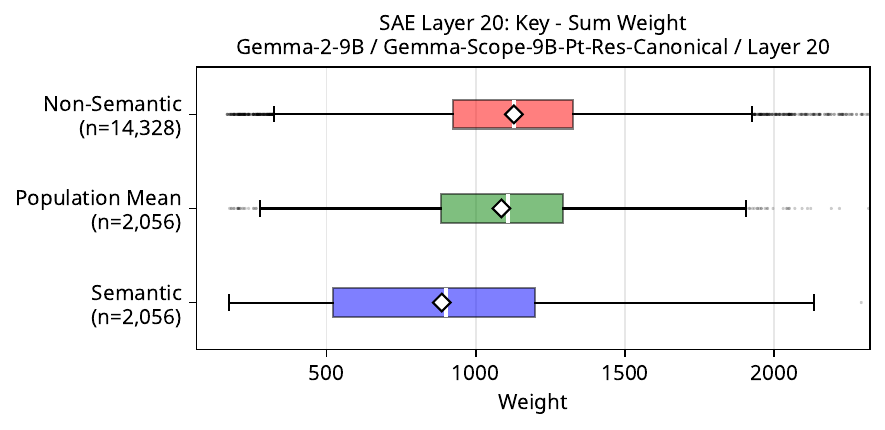}
    \caption{Key-based.}
    \label{fig:per-layer-boxplot-9b-k}
  \end{subfigure}
  \begin{subfigure}{.48\linewidth}
    \centering
    \includegraphics[width=\linewidth,trim={.0cm .0cm .0cm .0cm}]{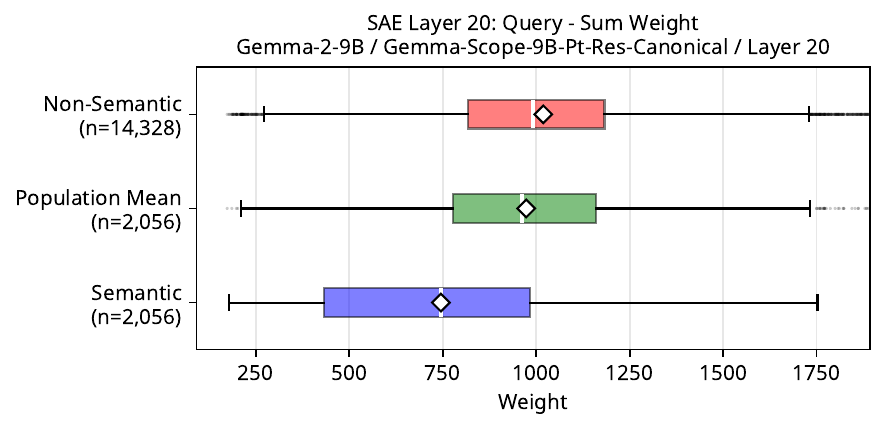}
    \caption{Query-based.}
    \label{fig:per-layer-boxplot-9b-q}
  \end{subfigure}
  \caption{Attention weight distributions in Gemma-2-9B layer 20. White diamonds mark means and white lines mark medians.}
  \label{fig:per-layer-boxplot-9b}
\end{figure}

\begin{figure}[ht]
  \centering
  \begin{subfigure}{.48\linewidth}
    \centering
    \includegraphics[width=\linewidth,trim={.0cm .0cm .0cm .0cm}]{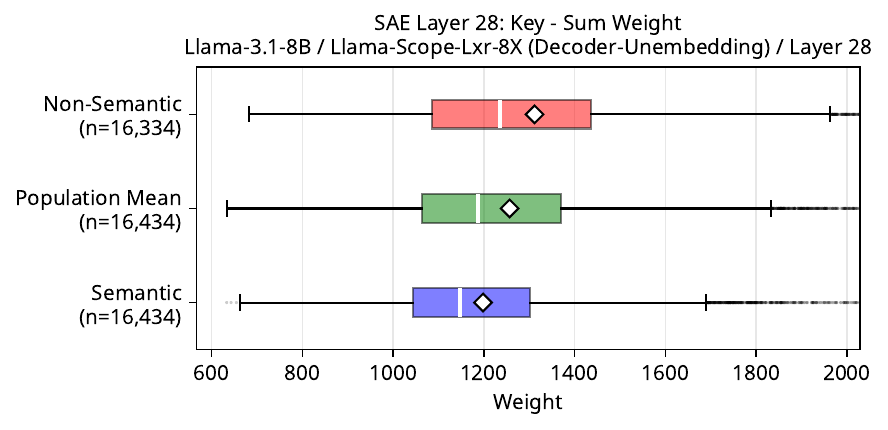}
    \caption{Key-based.}
    \label{fig:per-layer-boxplot-8b-dec-k}
  \end{subfigure}
  \begin{subfigure}{.48\linewidth}
    \centering
    \includegraphics[width=\linewidth,trim={.0cm .0cm .0cm .0cm}]{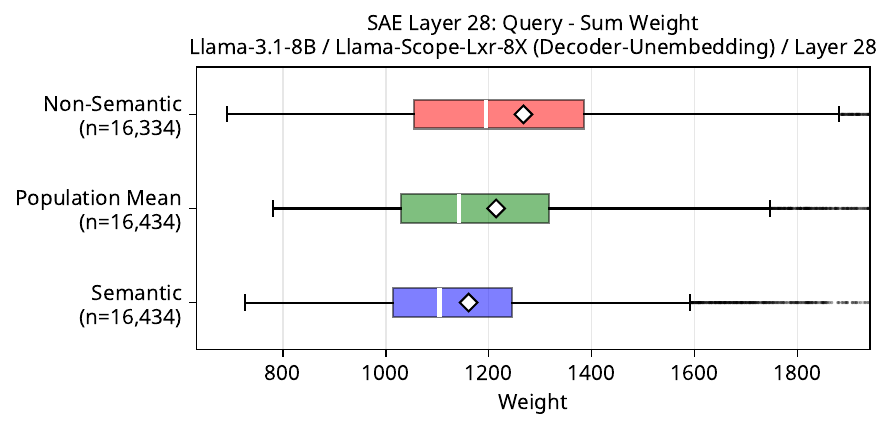}
    \caption{Query-based.}
    \label{fig:per-layer-boxplot-8b-dec-q}
  \end{subfigure}
  \caption{Attention weight distributions in Llama-3.1-8B layer 28. White diamonds mark means and white lines mark medians.}
  \label{fig:per-layer-boxplot-8b-dec}
\end{figure}

\begin{figure}[ht]
  \centering
  \begin{subfigure}{.48\linewidth}
    \centering
    \includegraphics[width=\linewidth,trim={.0cm .0cm .0cm .0cm}]{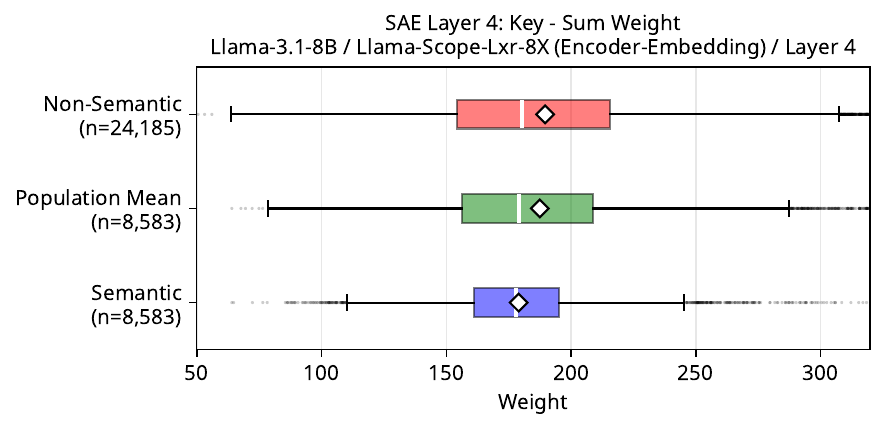}
    \caption{Key-based.}
    \label{fig:per-layer-boxplot-8b-enc-k}
  \end{subfigure}
  \begin{subfigure}{.48\linewidth}
    \centering
    \includegraphics[width=\linewidth,trim={.0cm .0cm .0cm .0cm}]{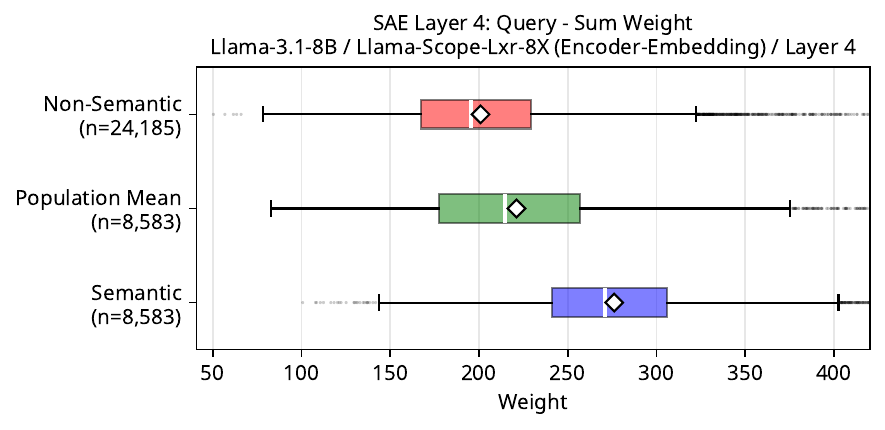}
    \caption{Query-based.}
    \label{fig:per-layer-boxplot-8b-enc-q}
  \end{subfigure}
  \caption{Attention weight distributions in Llama-3.1-8B layer 4. White diamonds mark means and white lines mark medians.}
  \label{fig:per-layer-boxplot-8b-enc}
\end{figure}

\begin{figure}[ht]
  \centering
  \begin{subfigure}{.7\linewidth}
    \centering
    \includegraphics[width=\linewidth,trim={.0cm .0cm .0cm .0cm}]{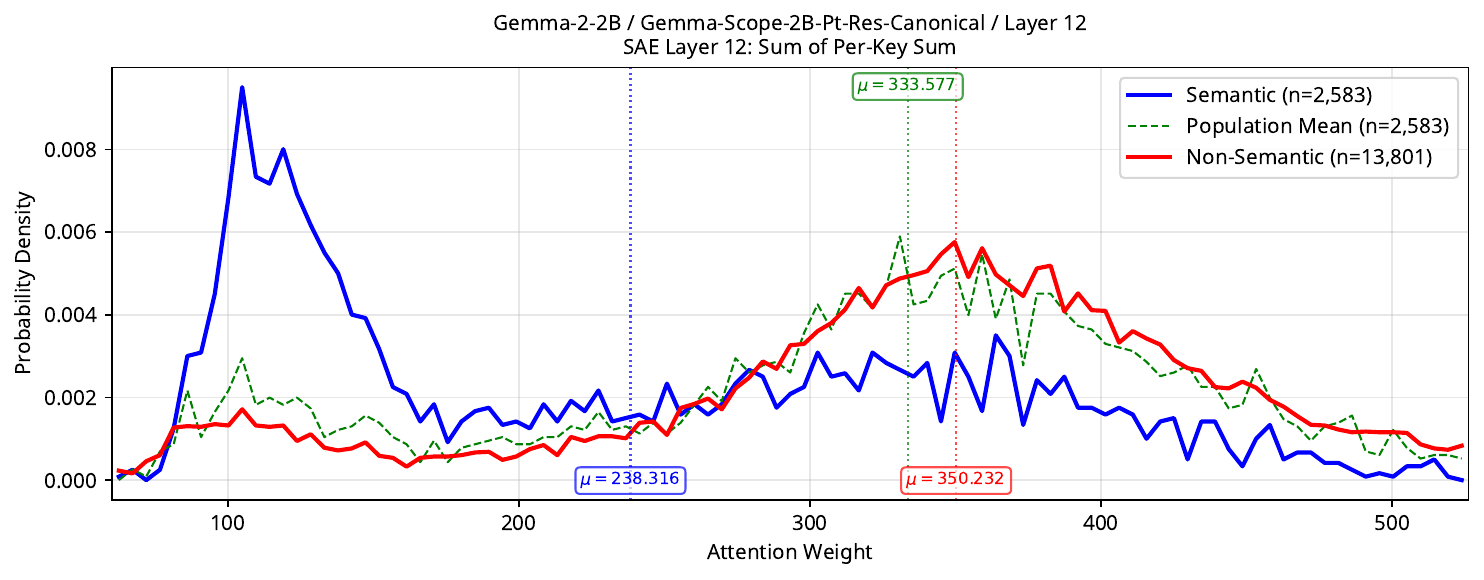}
    \caption{Key-based.}
    \label{fig:per-layer-density-2b-k}
  \end{subfigure}
  \begin{subfigure}{.7\linewidth}
    \centering
    \includegraphics[width=\linewidth,trim={.0cm .0cm .0cm .0cm}]{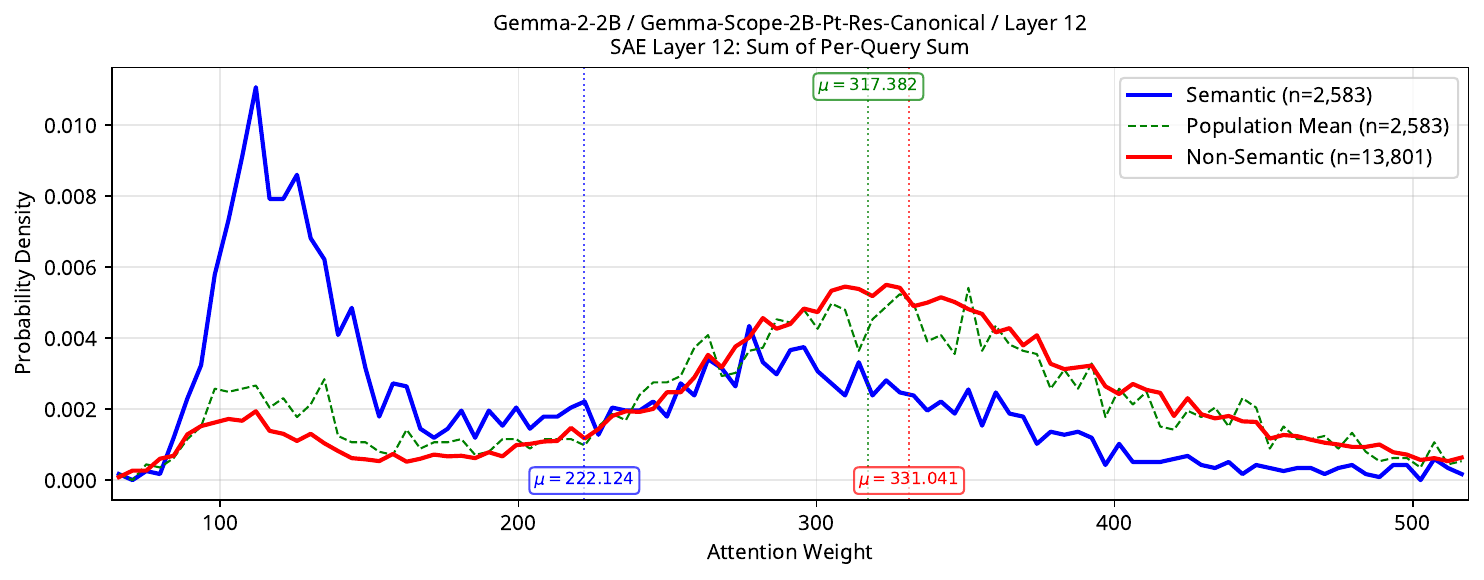}
    \caption{Query-based.}
    \label{fig:per-layer-density-2b-q}
  \end{subfigure}
  \caption{Probability densities of attention weights in Gemma-2-2B layer 12. Dotted lines mark means.}
  \label{fig:per-layer-density-2b}
\end{figure}

\begin{figure}[ht]
  \centering
  \begin{subfigure}{.7\linewidth}
    \centering
    \includegraphics[width=\linewidth,trim={.0cm .0cm .0cm .0cm}]{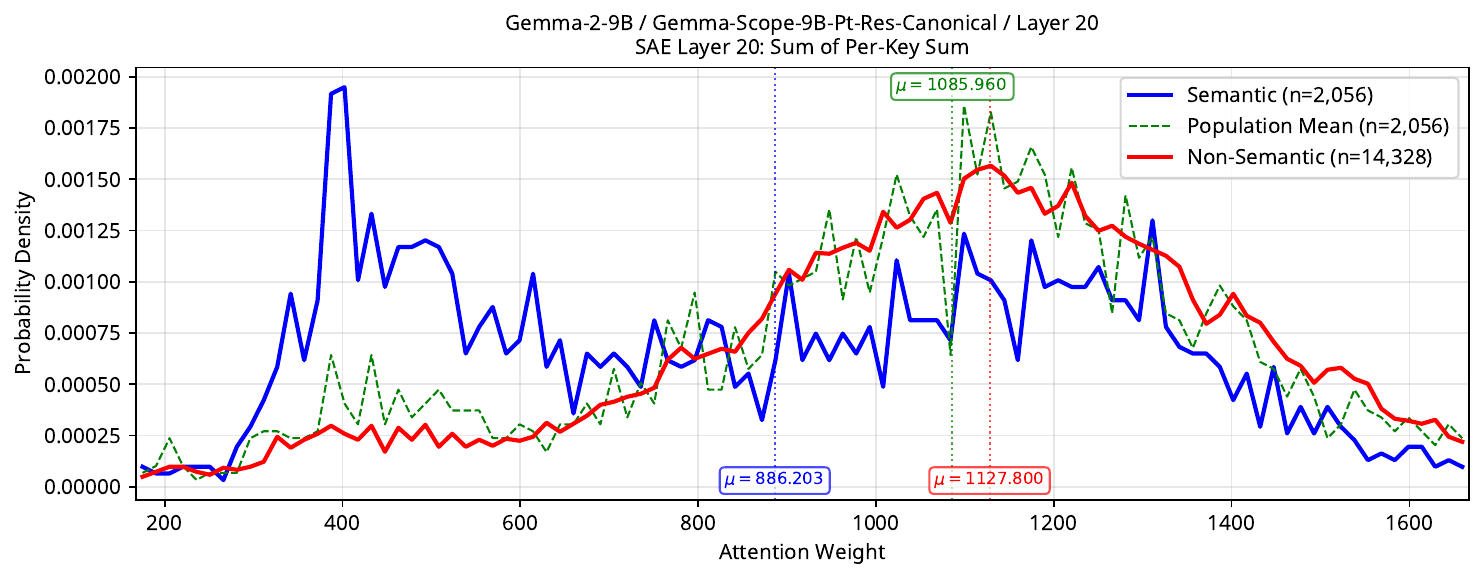}
    \caption{Key-based.}
    \label{fig:per-layer-density-9b-k}
  \end{subfigure}
  \begin{subfigure}{.7\linewidth}
    \centering
    \includegraphics[width=\linewidth,trim={.0cm .0cm .0cm .0cm}]{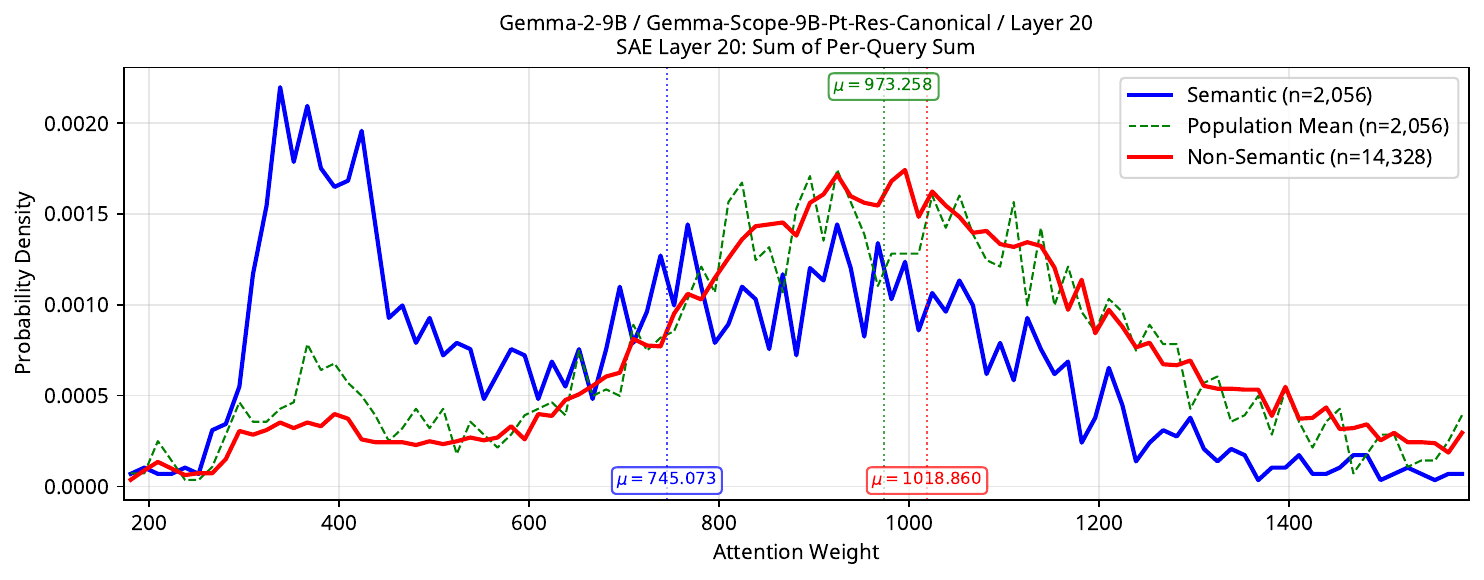}
    \caption{Query-based.}
    \label{fig:per-layer-density-9b-q}
  \end{subfigure}
  \caption{Probability densities of attention weights in Gemma-2-9B layer 20. Dotted lines mark means.}
  \label{fig:per-layer-density-9b}
\end{figure}

\begin{figure}[ht]
  \centering
  \begin{subfigure}{.7\linewidth}
    \centering
    \includegraphics[width=\linewidth,trim={.0cm .0cm .0cm .0cm}]{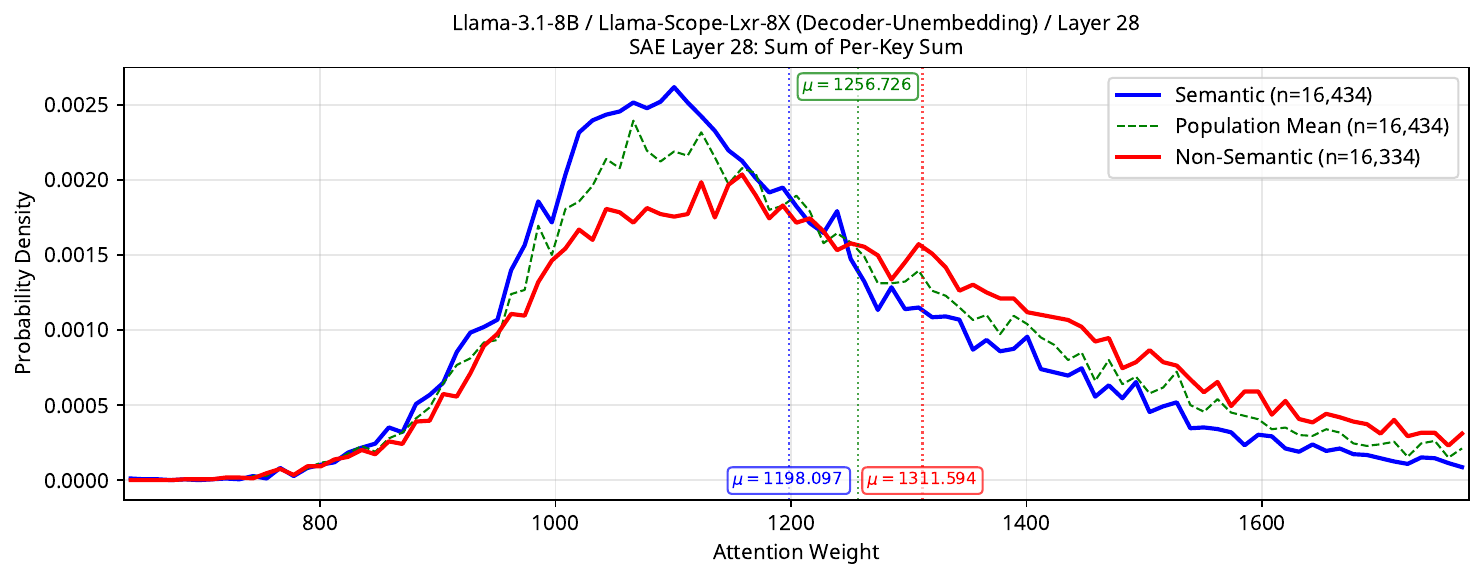}
    \caption{Key-based.}
    \label{fig:per-layer-density-8b-dec-k}
  \end{subfigure}
  \begin{subfigure}{.7\linewidth}
    \centering
    \includegraphics[width=\linewidth,trim={.0cm .0cm .0cm .0cm}]{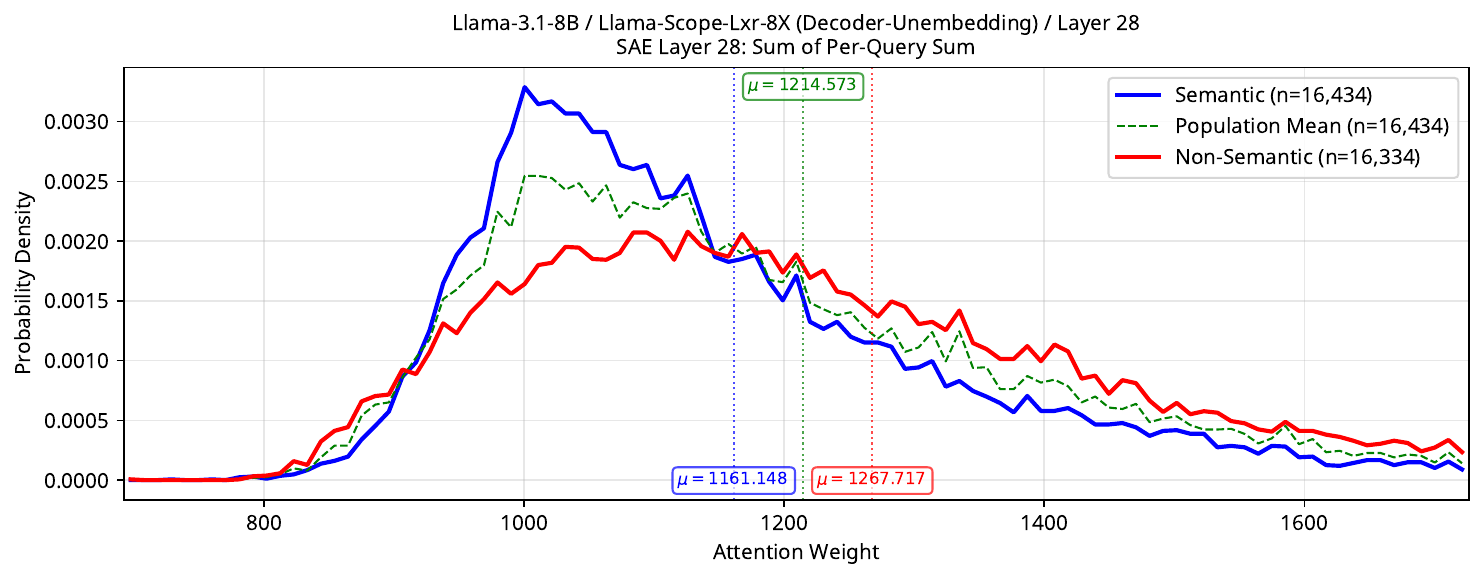}
    \caption{Query-based.}
    \label{fig:per-layer-density-8b-dec-q}
  \end{subfigure}
  \caption{Probability densities of attention weights in Llama-3.1-8B layer 28. Dotted lines mark means.}
  \label{fig:per-layer-density-8b-dec}
\end{figure}

\begin{figure}[ht]
  \centering
  \begin{subfigure}{.7\linewidth}
    \centering
    \includegraphics[width=\linewidth,trim={.0cm .0cm .0cm .0cm}]{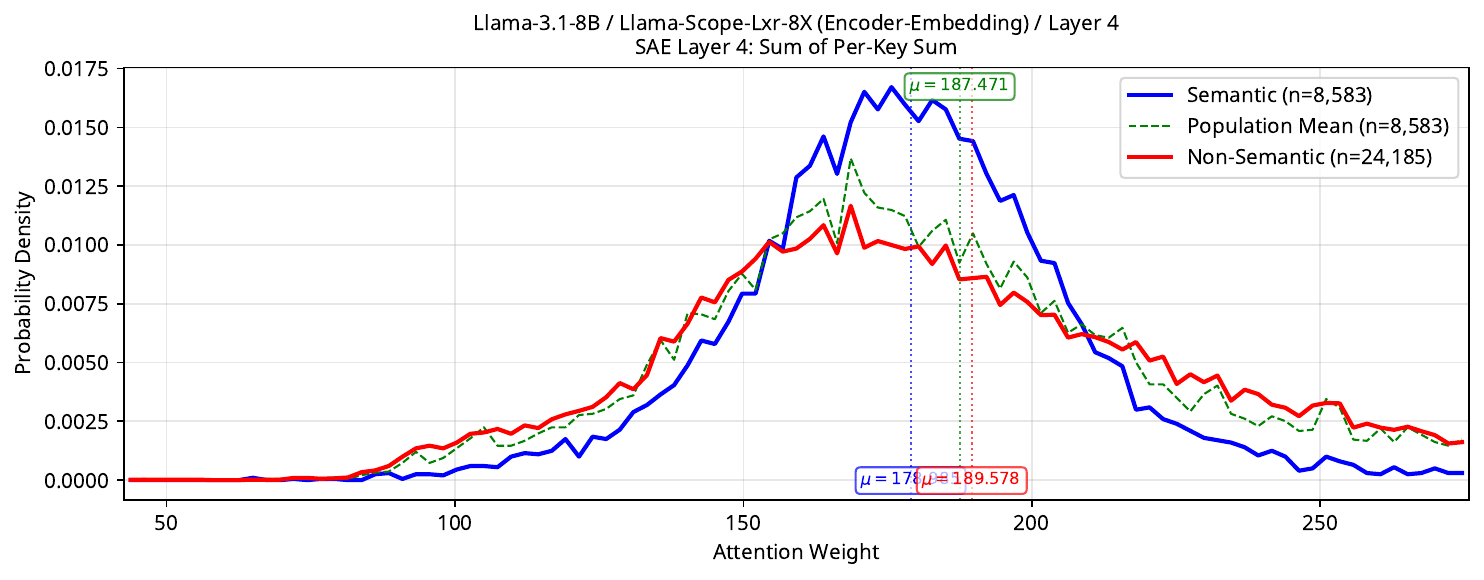}
    \caption{Key-based.}
    \label{fig:per-layer-density-8b-enc-k}
  \end{subfigure}
  \begin{subfigure}{.7\linewidth}
    \centering
    \includegraphics[width=\linewidth,trim={.0cm .0cm .0cm .0cm}]{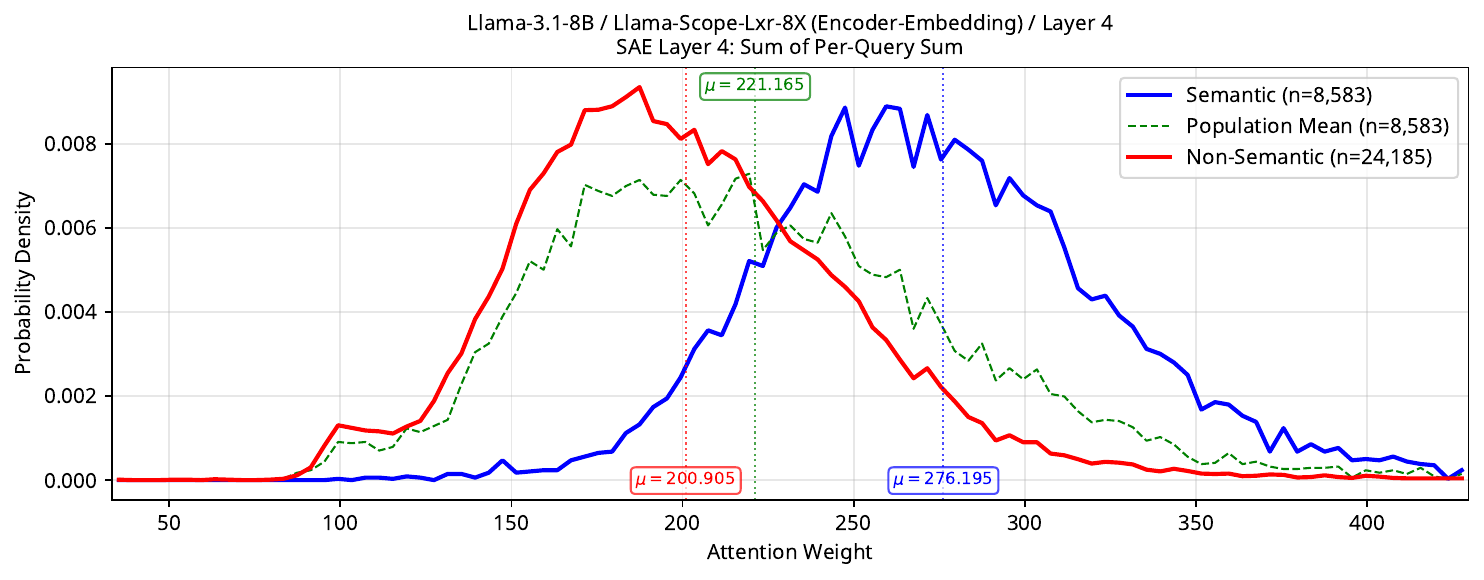}
    \caption{Query-based.}
    \label{fig:per-layer-density-8b-enc-q}
  \end{subfigure}
  \caption{Probability densities of attention weights in Llama-3.1-8B layer 4. Dotted lines mark means.}
  \label{fig:per-layer-density-8b-enc}
\end{figure}

\begin{figure}[ht]
  \centering
  \begin{subfigure}{.48\linewidth}
    \centering
    \includegraphics[width=\linewidth,trim={.0cm .0cm .0cm .0cm}]{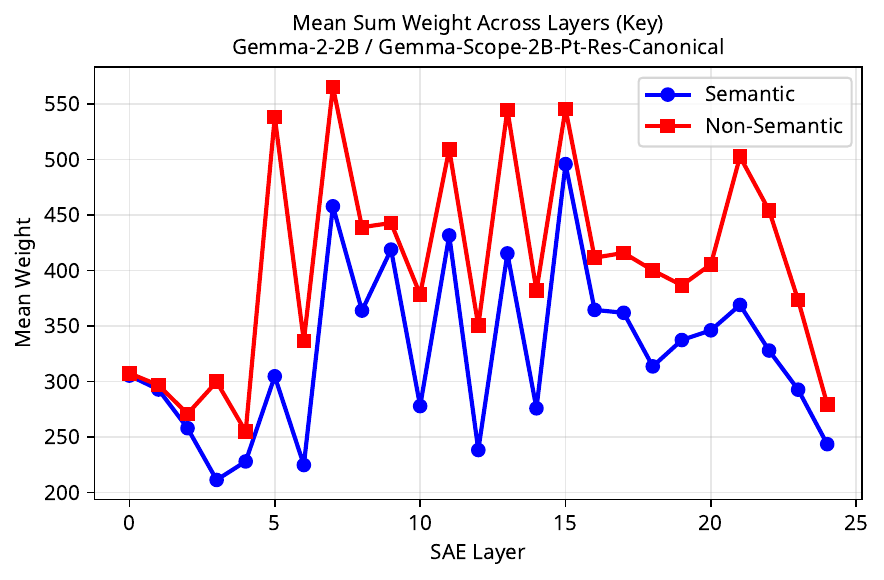}
    \caption{Key-based.}
    \label{fig:cross-layer-mean-2b-k}
  \end{subfigure}
  \begin{subfigure}{.48\linewidth}
    \centering
    \includegraphics[width=\linewidth,trim={.0cm .0cm .0cm .0cm}]{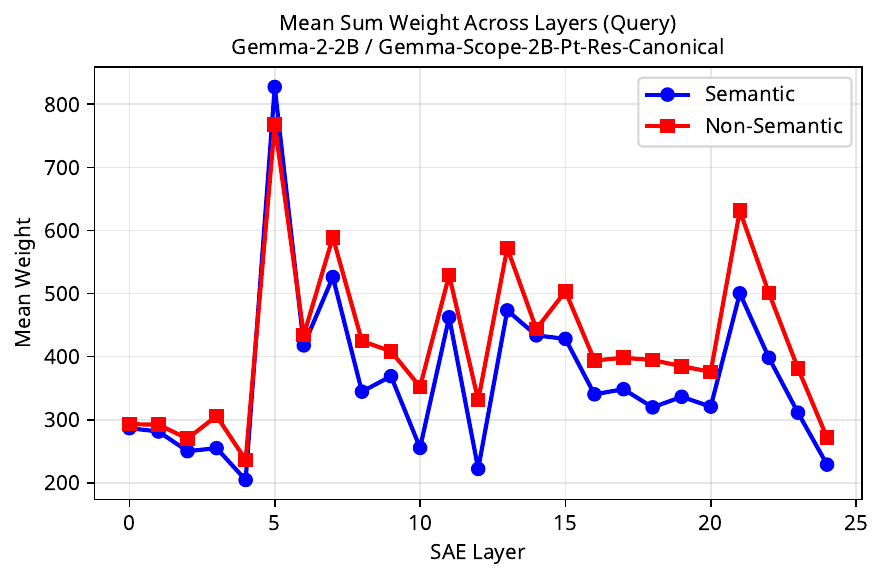}
    \caption{Query-based.}
    \label{fig:cross-layer-mean-2b-q}
  \end{subfigure}
  \caption{Mean attention weights across layers in Gemma-2-2B.}
  \label{fig:cross-layer-mean-2b}
\end{figure}

\begin{figure}[ht]
  \centering
  \begin{subfigure}{.48\linewidth}
    \centering
    \includegraphics[width=\linewidth,trim={.0cm .0cm .0cm .0cm}]{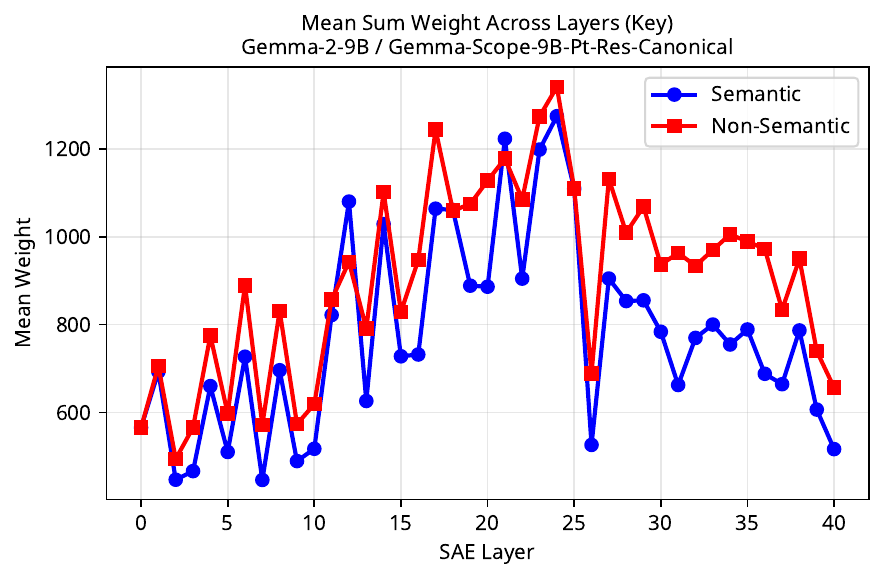}
    \caption{Key-based.}
    \label{fig:cross-layer-mean-9b-k}
  \end{subfigure}
  \begin{subfigure}{.48\linewidth}
    \centering
    \includegraphics[width=\linewidth,trim={.0cm .0cm .0cm .0cm}]{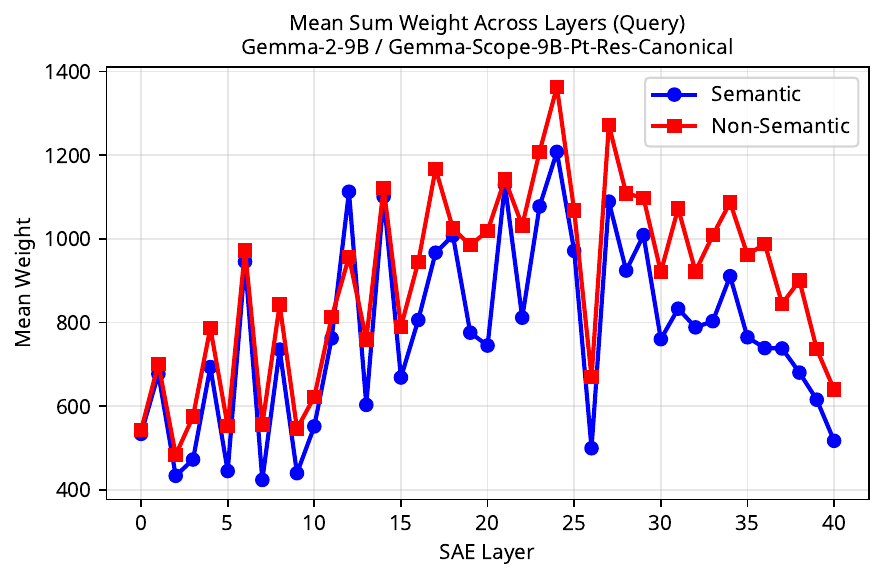}
    \caption{Query-based.}
    \label{fig:cross-layer-mean-9b-q}
  \end{subfigure}
  \caption{Mean attention weights across layers in Gemma-2-9B.}
  \label{fig:cross-layer-mean-9b}
\end{figure}

\begin{figure}[ht]
  \centering
  \begin{subfigure}{.48\linewidth}
    \centering
    \includegraphics[width=\linewidth,trim={.0cm .0cm .0cm .0cm}]{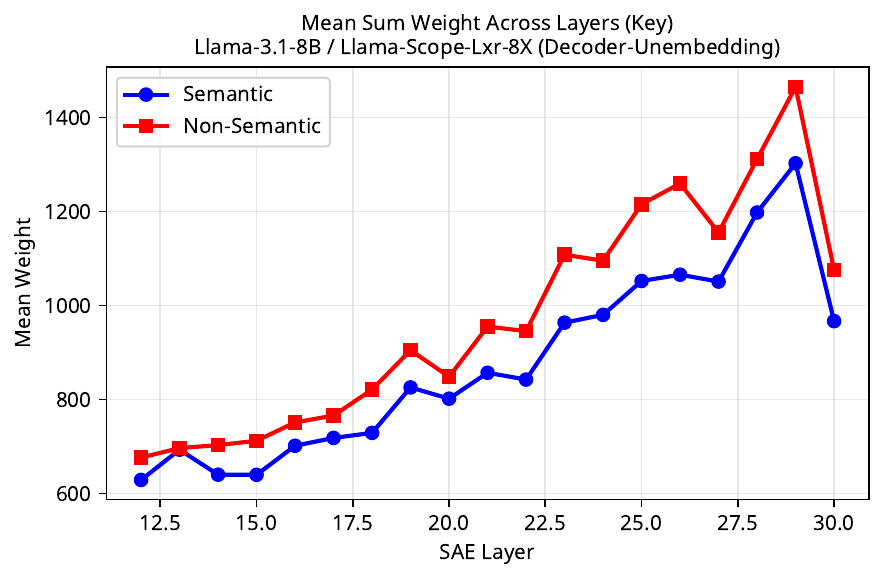}
    \caption{Key-based.}
    \label{fig:cross-layer-mean-8b-dec-k}
  \end{subfigure}
  \begin{subfigure}{.48\linewidth}
    \centering
    \includegraphics[width=\linewidth,trim={.0cm .0cm .0cm .0cm}]{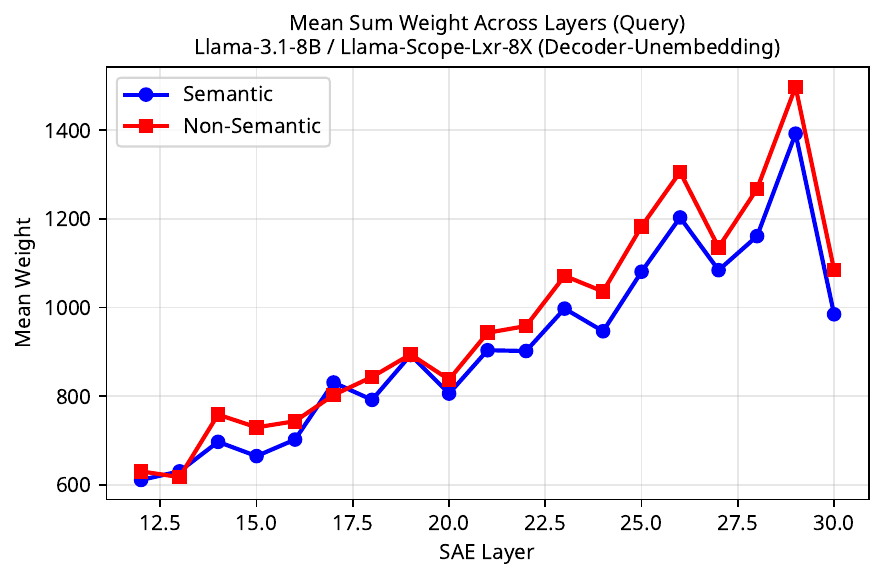}
    \caption{Query-based.}
    \label{fig:cross-layer-mean-8b-dec-q}
  \end{subfigure}
  \caption{Mean attention weights across layers in Llama-3.1-8B, decoder--unembedding phase (layers 12--31).}
  \label{fig:cross-layer-mean-8b-dec}
\end{figure}

\begin{figure}[ht]
  \centering
  \begin{subfigure}{.48\linewidth}
    \centering
    \includegraphics[width=\linewidth,trim={.0cm .0cm .0cm .0cm}]{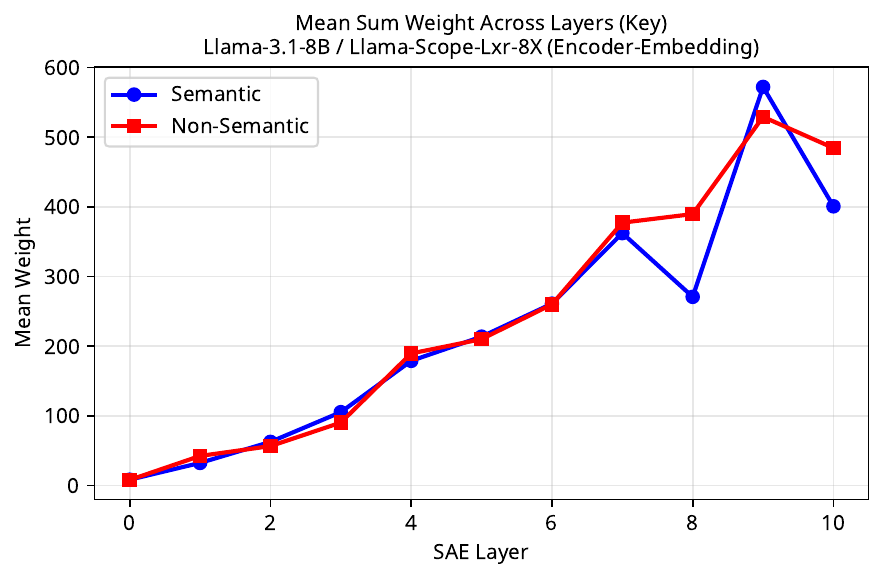}
    \caption{Key-based.}
    \label{fig:cross-layer-mean-8b-enc-k}
  \end{subfigure}
  \begin{subfigure}{.48\linewidth}
    \centering
    \includegraphics[width=\linewidth,trim={.0cm .0cm .0cm .0cm}]{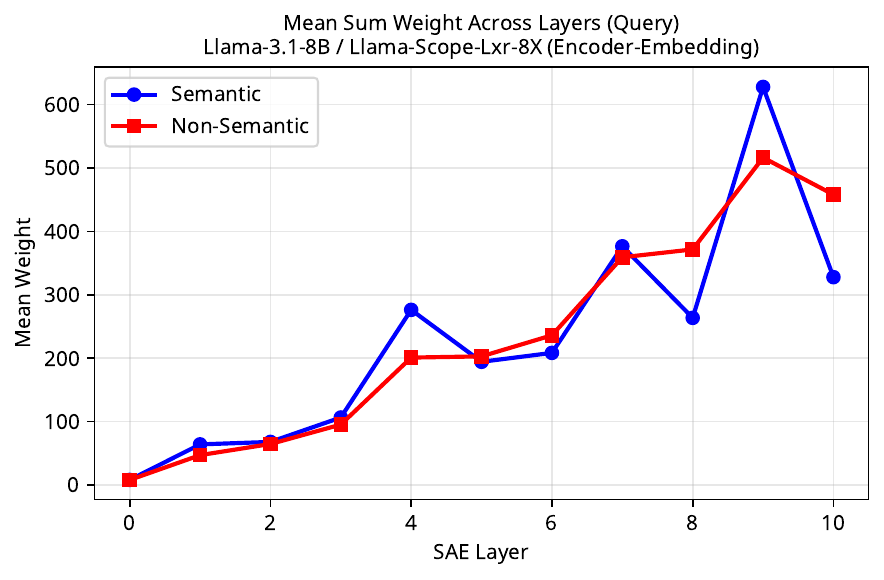}
    \caption{Query-based.}
    \label{fig:cross-layer-mean-8b-enc-q}
  \end{subfigure}
  \caption{Mean attention weights across layers in Llama-3.1-8B, encoder--embedding phase (layers 0--10).}
  \label{fig:cross-layer-mean-8b-enc}
\end{figure}

\begin{figure}[ht]
  \centering
  \begin{subfigure}{.48\linewidth}
    \centering
    \includegraphics[width=\linewidth,trim={.0cm .0cm .0cm .0cm}]{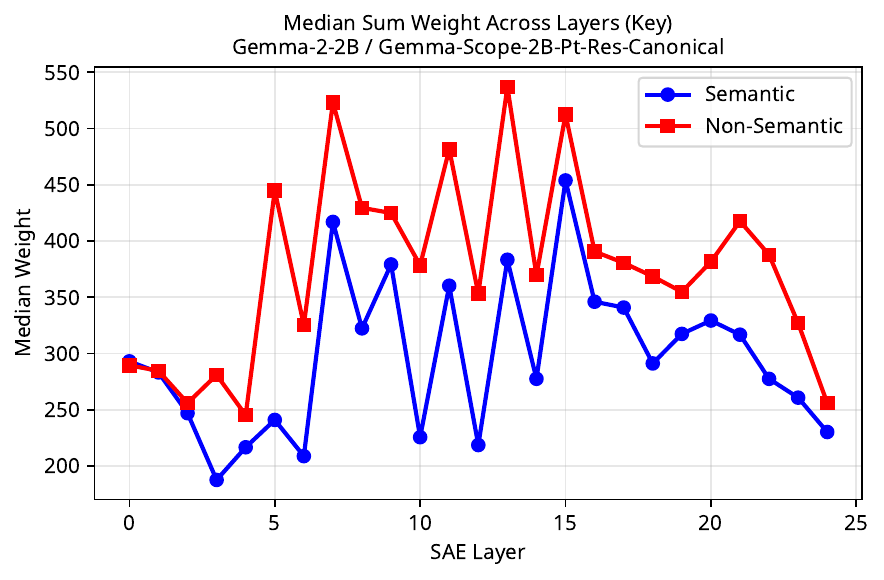}
    \caption{Key-based.}
    \label{fig:cross-layer-median-2b-k}
  \end{subfigure}
  \begin{subfigure}{.48\linewidth}
    \centering
    \includegraphics[width=\linewidth,trim={.0cm .0cm .0cm .0cm}]{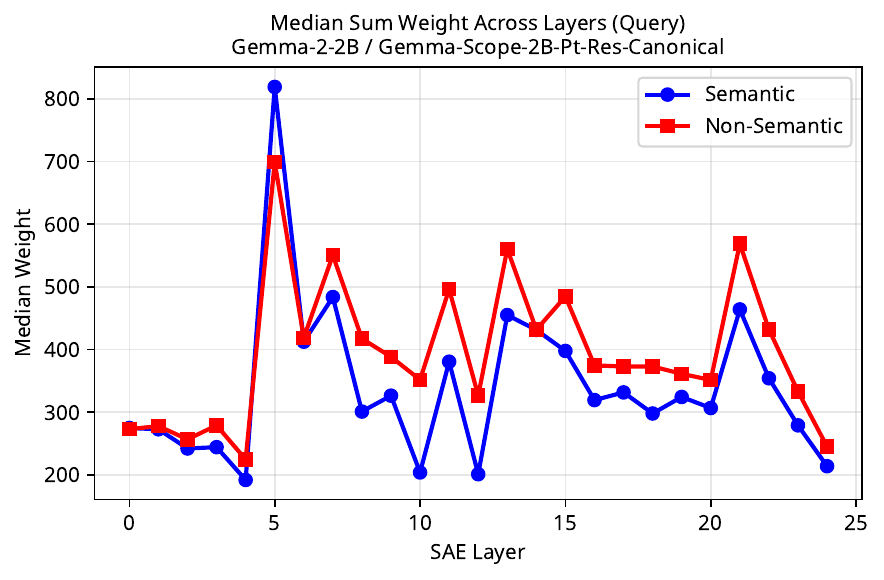}
    \caption{Query-based.}
    \label{fig:cross-layer-median-2b-q}
  \end{subfigure}
  \caption{Median attention weights across layers in Gemma-2-2B.}
  \label{fig:cross-layer-median-2b}
\end{figure}

\begin{figure}[ht]
  \centering
  \begin{subfigure}{.48\linewidth}
    \centering
    \includegraphics[width=\linewidth,trim={.0cm .0cm .0cm .0cm}]{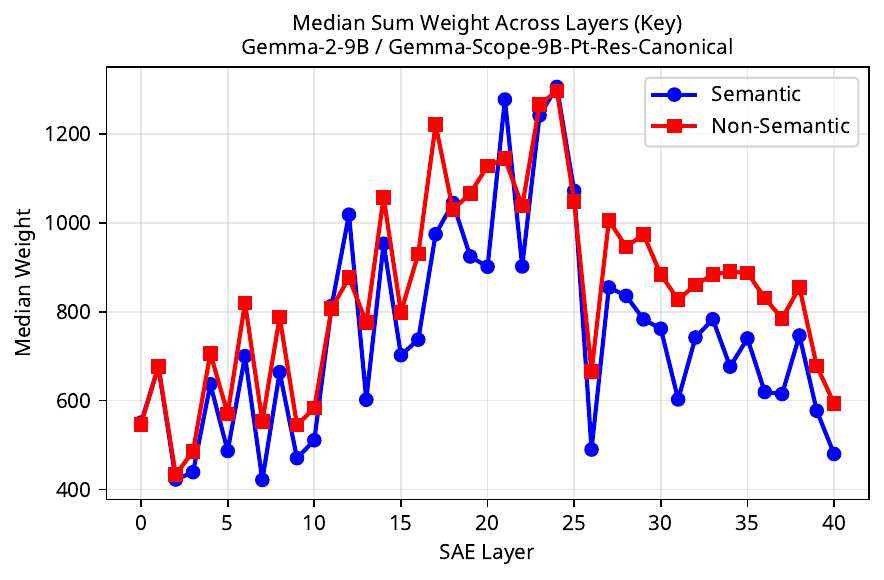}
    \caption{Key-based.}
    \label{fig:cross-layer-median-9b-k}
  \end{subfigure}
  \begin{subfigure}{.48\linewidth}
    \centering
    \includegraphics[width=\linewidth,trim={.0cm .0cm .0cm .0cm}]{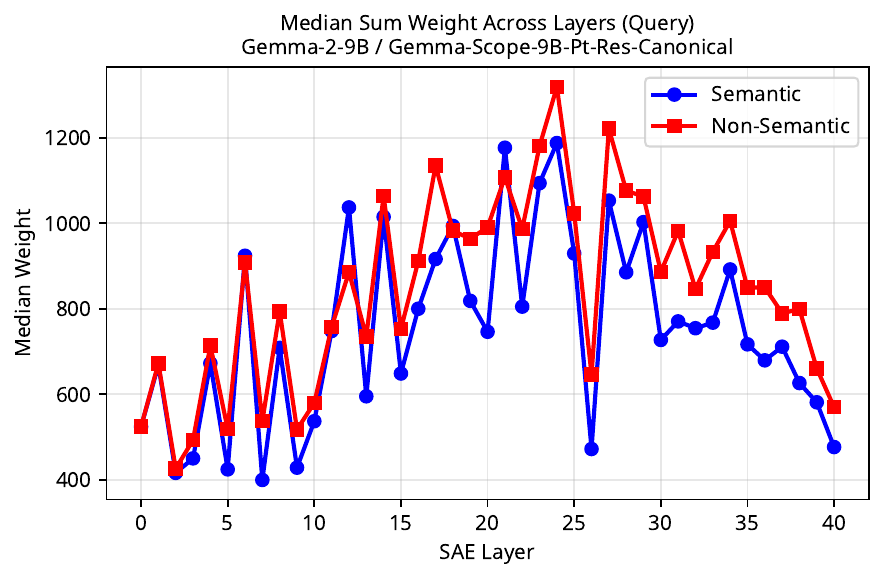}
    \caption{Query-based.}
    \label{fig:cross-layer-median-9b-q}
  \end{subfigure}
  \caption{Median attention weights across layers in Gemma-2-9B.}
  \label{fig:cross-layer-median-9b}
\end{figure}

\begin{figure}[ht]
  \centering
  \begin{subfigure}{.48\linewidth}
    \centering
    \includegraphics[width=\linewidth,trim={.0cm .0cm .0cm .0cm}]{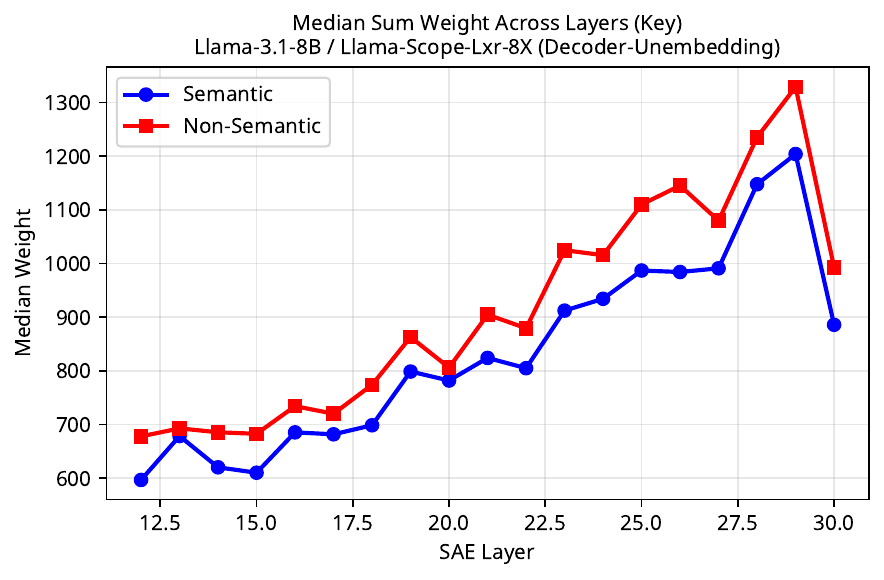}
    \caption{Key-based.}
    \label{fig:cross-layer-median-8b-dec-k}
  \end{subfigure}
  \begin{subfigure}{.48\linewidth}
    \centering
    \includegraphics[width=\linewidth,trim={.0cm .0cm .0cm .0cm}]{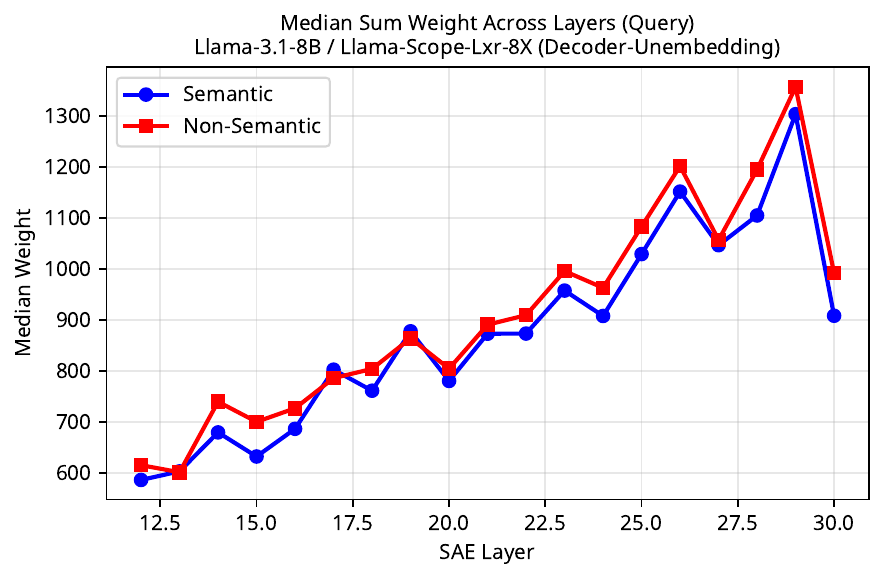}
    \caption{Query-based.}
    \label{fig:cross-layer-median-8b-dec-q}
  \end{subfigure}
  \caption{Median attention weights across layers in Llama-3.1-8B, decoder--unembedding phase (layers 12--31).}
  \label{fig:cross-layer-median-8b-dec}
\end{figure}

\begin{figure}[ht]
  \centering
  \begin{subfigure}{.48\linewidth}
    \centering
    \includegraphics[width=\linewidth,trim={.0cm .0cm .0cm .0cm}]{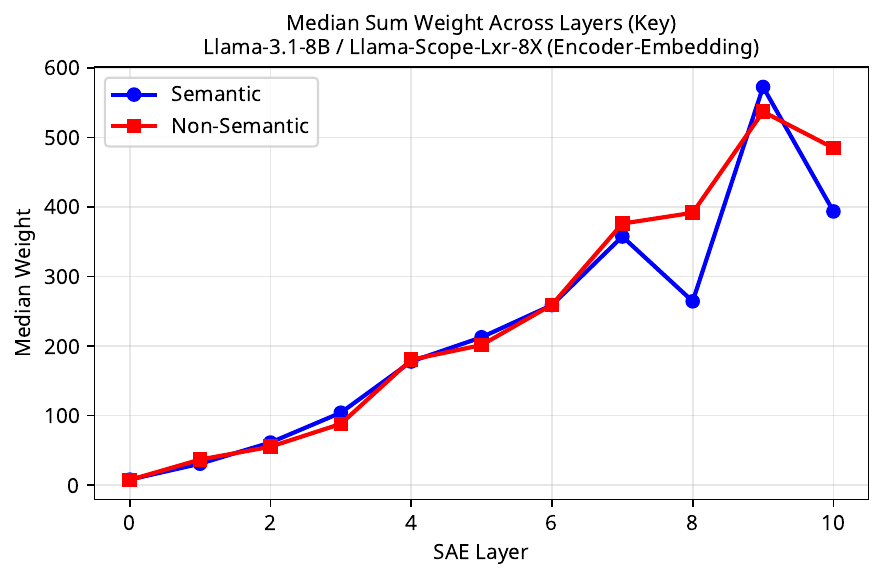}
    \caption{Key-based.}
    \label{fig:cross-layer-median-8b-enc-k}
  \end{subfigure}
  \begin{subfigure}{.48\linewidth}
    \centering
    \includegraphics[width=\linewidth,trim={.0cm .0cm .0cm .0cm}]{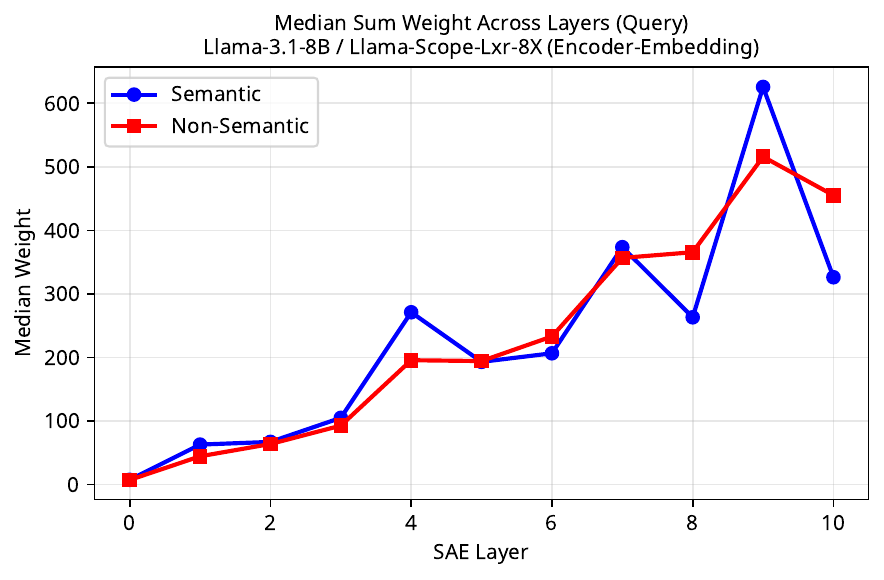}
    \caption{Query-based.}
    \label{fig:cross-layer-median-8b-enc-q}
  \end{subfigure}
  \caption{Median attention weights across layers in Llama-3.1-8B, encoder--embedding phase (layers 0--10).}
  \label{fig:cross-layer-median-8b-enc}
\end{figure}

\begin{figure}[ht]
  \centering
  \begin{subfigure}{.48\linewidth}
    \centering
    \includegraphics[width=\linewidth,trim={.0cm .0cm .0cm .0cm}]{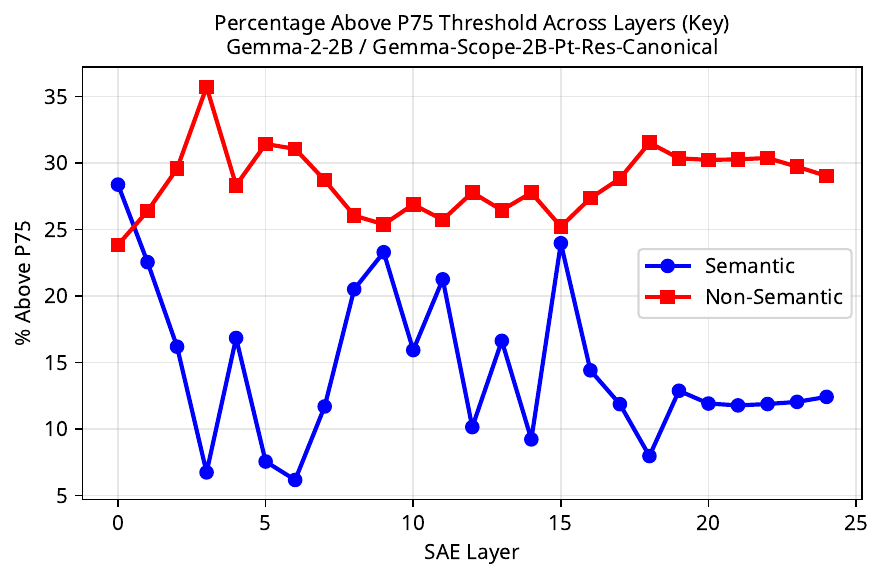}
    \caption{Key-based.}
    \label{fig:cross-layer-p75-2b-k}
  \end{subfigure}
  \begin{subfigure}{.48\linewidth}
    \centering
    \includegraphics[width=\linewidth,trim={.0cm .0cm .0cm .0cm}]{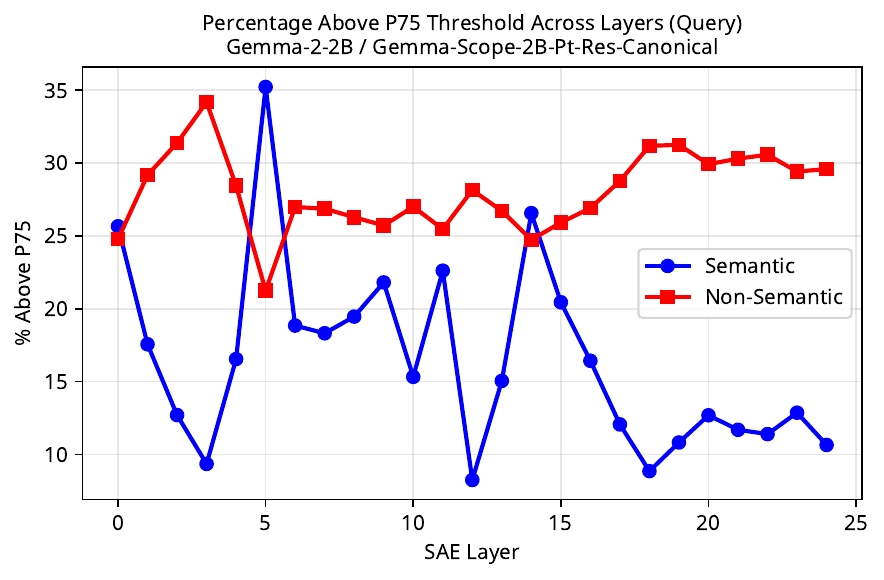}
    \caption{Query-based.}
    \label{fig:cross-layer-p75-2b-q}
  \end{subfigure}
  \caption{Percentage of features exceeding global $P_{75}$ threshold across layers in Gemma-2-2B.}
  \label{fig:cross-layer-p75-2b}
\end{figure}

\begin{figure}[ht]
  \centering
  \begin{subfigure}{.48\linewidth}
    \centering
    \includegraphics[width=\linewidth,trim={.0cm .0cm .0cm .0cm}]{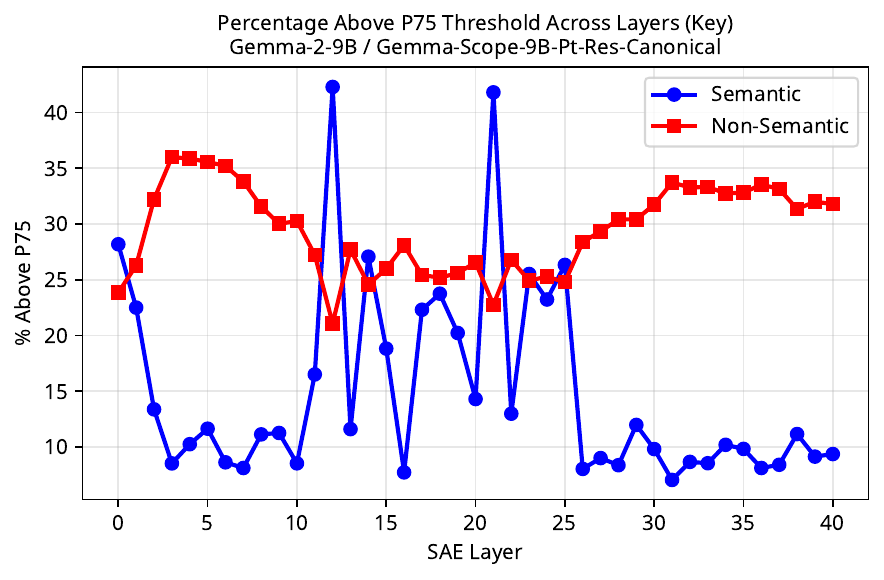}
    \caption{Key-based.}
    \label{fig:cross-layer-p75-9b-k}
  \end{subfigure}
  \begin{subfigure}{.48\linewidth}
    \centering
    \includegraphics[width=\linewidth,trim={.0cm .0cm .0cm .0cm}]{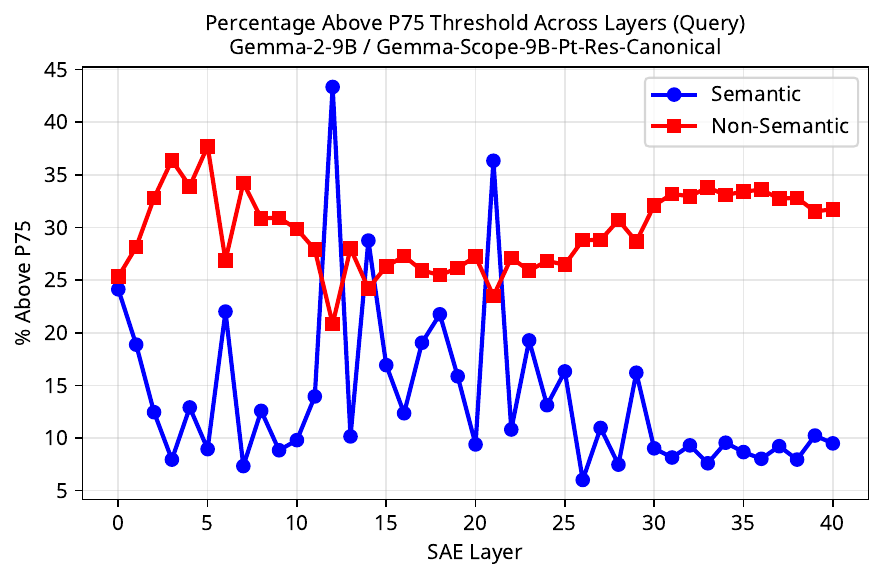}
    \caption{Query-based.}
    \label{fig:cross-layer-p75-9b-q}
  \end{subfigure}
  \caption{Percentage of features exceeding global $P_{75}$ threshold across layers in Gemma-2-9B.}
  \label{fig:cross-layer-p75-9b}
\end{figure}

\begin{figure}[ht]
  \centering
  \begin{subfigure}{.48\linewidth}
    \centering
    \includegraphics[width=\linewidth,trim={.0cm .0cm .0cm .0cm}]{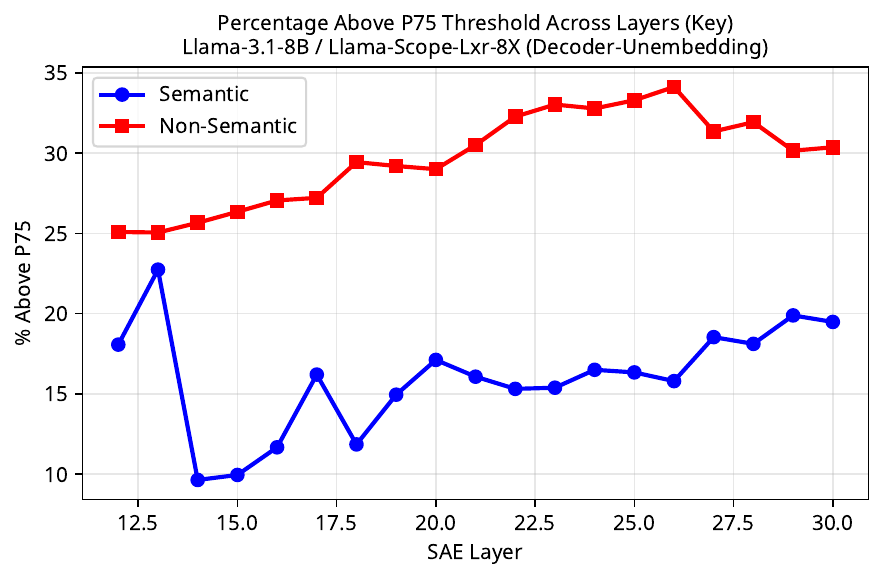}
    \caption{Key-based.}
    \label{fig:cross-layer-p75-8b-dec-k}
  \end{subfigure}
  \begin{subfigure}{.48\linewidth}
    \centering
    \includegraphics[width=\linewidth,trim={.0cm .0cm .0cm .0cm}]{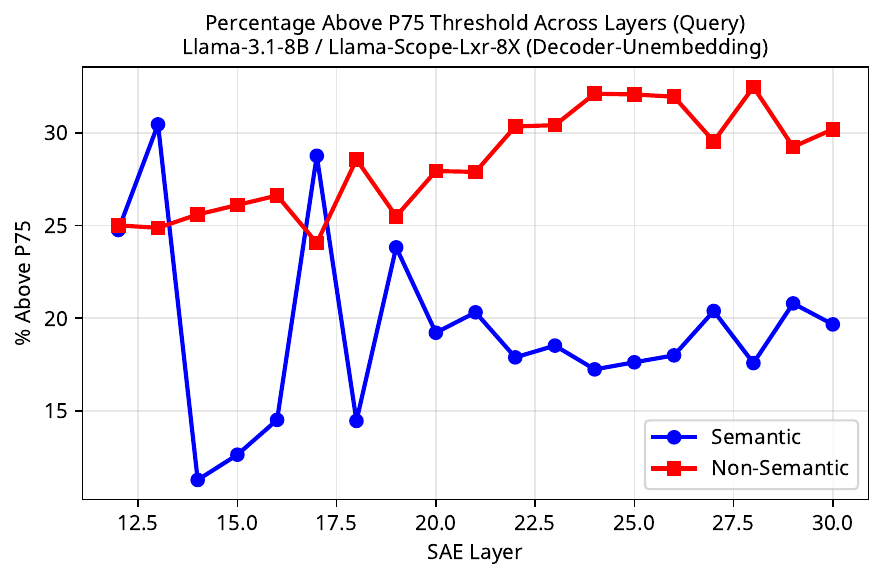}
    \caption{Query-based.}
    \label{fig:cross-layer-p75-8b-dec-q}
  \end{subfigure}
  \caption{Percentage of features exceeding global $P_{75}$ threshold across layers in Llama-3.1-8B, decoder--unembedding phase (layers 12--31).}
  \label{fig:cross-layer-p75-8b-dec}
\end{figure}

\begin{figure}[ht]
  \centering
  \begin{subfigure}{.48\linewidth}
    \centering
    \includegraphics[width=\linewidth,trim={.0cm .0cm .0cm .0cm}]{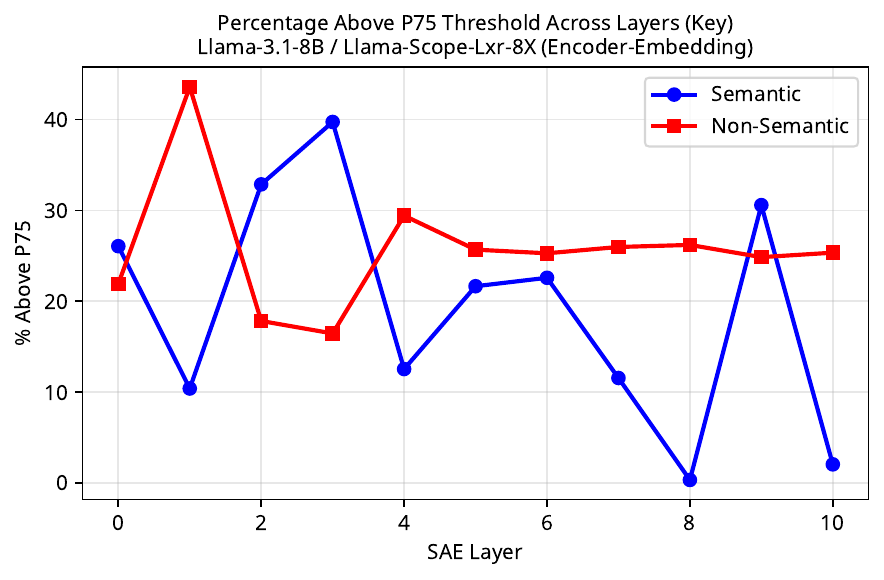}
    \caption{Key-based.}
    \label{fig:cross-layer-p75-8b-enc-k}
  \end{subfigure}
  \begin{subfigure}{.48\linewidth}
    \centering
    \includegraphics[width=\linewidth,trim={.0cm .0cm .0cm .0cm}]{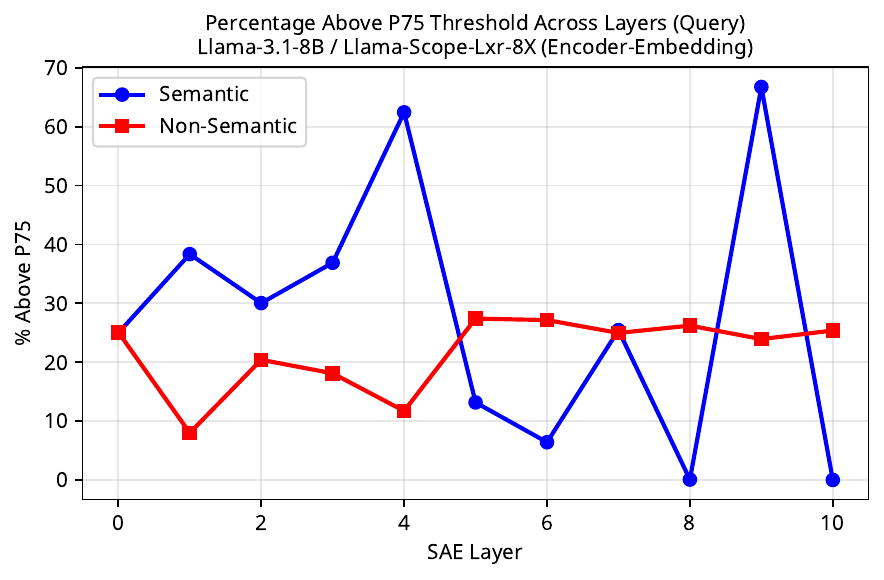}
    \caption{Query-based.}
    \label{fig:cross-layer-p75-8b-enc-q}
  \end{subfigure}
  \caption{Percentage of features exceeding global $P_{75}$ threshold across layers in Llama-3.1-8B, encoder--embedding phase (layers 0--10).}
  \label{fig:cross-layer-p75-8b-enc}
\end{figure}

\end{document}